\documentclass[runningheads]{llncs}

\usepackage{eccv}

\usepackage{eccvabbrv}
\usepackage{graphicx}
\usepackage{amsmath}
\usepackage{amssymb}
\usepackage{booktabs}
\usepackage{tikz}
\usepackage{bbm}
\usepackage{bm}
\usepackage{pifont}
\usepackage{cuted}
\usepackage{colortbl}
\usepackage{algorithm}
\usepackage{algpseudocode}
\usepackage{xifthen}
\usepackage{comment}
\usepackage{wrapfig}
\usepackage{tcolorbox}

\usepackage{mathtools, nccmath}
\DeclarePairedDelimiter{\nint}\lfloor\rceil

\usepackage[labelfont=bf,font=small,skip=3pt]{caption}

\usepackage[accsupp]{axessibility}  %

\usepackage[breaklinks,colorlinks,citecolor=eccvblue]{hyperref}

\usepackage[capitalize]{cleveref}
\crefname{section}{Sec.}{Secs.}
\Crefname{section}{Section}{Sections}
\Crefname{table}{Table}{Tables}
\crefname{table}{Tab.}{Tabs.}

\usepackage{orcidlink}

\begin{document}

\newcommand{\tree}{\bm{\Gamma}}
\newcommand{\shenlong}[1]{\textcolor{magenta}{#1}}
\newcommand{\todocite}[1]{\textcolor{red}{\textit{Citation needed []}}}
\newcommand{\shenlongsay}[1]{\textcolor{blue}{[Shenlong: #1]}}
\newcommand{\sg}[1]{\textcolor{blue}{[Saurabh: #1]}}
\newcommand{\todo}[1]{\textcolor{red}{\textit{TODO: #1}}}
\newcommand{\shaowei}[1]{\textcolor{magenta}{[Shaowei: #1]}}

\newcommand{\imgtile}[2]{
    {\tikz{
    \node[draw=black, draw opacity=1.0, line width=.3mm, fill opacity=0.7,fill=white, inner sep=0pt](gt) at (0, 0) {\includegraphics[width=#2\linewidth]{#1}};
    \node[draw=black, draw opacity=1.0, line width=.3mm, fill opacity=0.7,fill=white, inner sep=0pt](gt) at (-0.1, 0.1) {\includegraphics[width=#2\linewidth]{#1}};
    \node[draw=black, draw opacity=1.0, line width=.3mm, fill opacity=0.7,fill=white, inner sep=0pt](gt) at (-0.2, 0.2) {\includegraphics[width=#2\linewidth]{#1}};
    \node[draw=black, draw opacity=1.0, line width=.3mm, fill opacity=0.7,fill=white, inner sep=0pt](gt) at (-0.3, 0.3) {\includegraphics[width=#2\linewidth]{#1}}; }}
}

\newcommand{\robotD}[0]{RoboArt\xspace}
\newcommand{\sapiensD}[0]{Sapiens\xspace}
\newcommand{\wim}[0]{WatchItMove\xspace}
\newcommand{\mbs}[0]{MultiBodySync\xspace}
\newcommand{\xpar}[1]{\noindent\textbf{#1}\ \ }
\newcommand{\vpar}[1]{\vspace{3mm}\noindent\textbf{#1}\ \ }

\newcommand{\sect}[1]{Section~\ref{#1}}
\newcommand{\sects}[1]{Sections~\ref{#1}}
\newcommand{\eqn}[1]{Equation~\ref{#1}}
\newcommand{\eqns}[1]{Equations~\ref{#1}}
\newcommand{\fig}[1]{Figure~\ref{#1}}
\newcommand{\figs}[1]{Figures~\ref{#1}}
\newcommand{\tab}[1]{Table~\ref{#1}}
\newcommand{\tabs}[1]{Tables~\ref{#1}}

\newcommand{\ignorethis}[1]{}
\newcommand{\norm}[1]{\lVert#1\rVert}
\newcommand{\fcseven}{$\mbox{fc}_7$}

\renewcommand*{\thefootnote}{\fnsymbol{footnote}}

\def\naive{na\"{\i}ve\xspace}
\def\Naive{Na\"{\i}ve\xspace}

\makeatletter
\DeclareRobustCommand\onedot{\futurelet\@let@token\@onedot}
\def\@onedot{\ifx\@let@token.\else.\null\fi\xspace}

\def\iid{\emph{i.i.d}\onedot}
\def\eg{\emph{e.g}\onedot} \def\Eg{\emph{E.g}\onedot}
\def\ie{\emph{i.e}\onedot} \def\Ie{\emph{I.e}\onedot}
\def\cf{\emph{c.f}\onedot} \def\Cf{\emph{C.f}\onedot}
\def\etc{\emph{etc}\onedot} \def\vs{\emph{vs}\onedot}
\def\wrt{w.r.t\onedot} \def\dof{d.o.f\onedot}
\def\etal{\emph{et al}\onedot}
\makeatother

\definecolor{citecolor}{RGB}{34,139,34}
\definecolor{mydarkblue}{rgb}{0,0.08,1}
\definecolor{mydarkgreen}{rgb}{0.02,0.6,0.02}
\definecolor{mydarkred}{rgb}{0.8,0.02,0.02}
\definecolor{mydarkorange}{rgb}{0.40,0.2,0.02}
\definecolor{mypurple}{RGB}{111,0,255}
\definecolor{myred}{rgb}{1.0,0.0,0.0}
\definecolor{mygold}{rgb}{0.75,0.6,0.12}
\definecolor{myblue}{rgb}{0,0.2,0.8}
\definecolor{mydarkgray}{rgb}{0.66,0.66,0.66}

\newcommand{\myparagraph}[1]{\vspace{-6pt}\paragraph{#1}}

\newcommand{\bbR}{{\mathbb{R}}}
\newcommand{\bK}{\mathbf{K}}
\newcommand{\bX}{\mathbf{X}}
\newcommand{\bY}{\mathbf{Y}}
\newcommand{\bk}{\mathbf{k}}
\newcommand{\bx}{\mathbf{x}}
\newcommand{\by}{\mathbf{y}}
\newcommand{\bhy}{\hat{\mathbf{y}}}
\newcommand{\bty}{\tilde{\mathbf{y}}}
\newcommand{\bG}{\mathbf{G}}
\newcommand{\bI}{\mathbf{I}}
\newcommand{\bg}{\mathbf{g}}
\newcommand{\bS}{\mathbf{S}}
\newcommand{\bs}{\mathbf{s}}
\newcommand{\bM}{\mathbf{M}}
\newcommand{\bw}{\mathbf{w}}
\newcommand{\eye}{\mathbf{I}}
\newcommand{\bU}{\mathbf{U}}
\newcommand{\bV}{\mathbf{V}}
\newcommand{\bW}{\mathbf{W}}
\newcommand{\bn}{\mathbf{n}}
\newcommand{\bv}{\mathbf{v}}
\newcommand{\bq}{\mathbf{q}}
\newcommand{\bR}{\mathbf{R}}
\newcommand{\bi}{\mathbf{i}}
\newcommand{\bj}{\mathbf{j}}
\newcommand{\bp}{\mathbf{p}}
\newcommand{\bt}{\mathbf{t}}
\newcommand{\bJ}{\mathbf{J}}
\newcommand{\bu}{\mathbf{u}}
\newcommand{\bB}{\mathbf{B}}
\newcommand{\bD}{\mathbf{D}}
\newcommand{\bz}{\mathbf{z}}
\newcommand{\bP}{\mathbf{P}}
\newcommand{\bC}{\mathbf{C}}
\newcommand{\bA}{\mathbf{A}}
\newcommand{\bZ}{\mathbf{Z}}
\newcommand{\bff}{\mathbf{f}}
\newcommand{\bF}{\mathbf{F}}
\newcommand{\bo}{\mathbf{o}}
\newcommand{\bc}{\mathbf{c}}
\newcommand{\bT}{\mathbf{T}}
\newcommand{\bQ}{\mathbf{Q}}
\newcommand{\bL}{\mathbf{L}}
\newcommand{\bl}{\mathbf{l}}
\newcommand{\ba}{\mathbf{a}}
\newcommand{\bE}{\mathbf{E}}
\newcommand{\bH}{\mathbf{H}}
\newcommand{\bd}{\mathbf{d}}
\newcommand{\br}{\mathbf{r}}
\newcommand{\bb}{\mathbf{b}}
\newcommand{\bh}{\mathbf{h}}

\newcommand{\btheta}{\bm{\theta}}

\newcommand{\bhh}{\hat{\mathbf{h}}}
\newcommand{\ci}{{\cal I}}
\newcommand{\ct}{{\cal T}}
\newcommand{\co}{{\cal O}}
\newcommand{\ck}{{\cal K}}
\newcommand{\cu}{{\cal U}}
\newcommand{\cS}{{\cal S}}
\newcommand{\cQ}{{\cal Q}}
\newcommand{\cT}{{\cal S}}
\newcommand{\cC}{{\cal C}}
\newcommand{\cE}{{\cal E}}
\newcommand{\cF}{{\cal F}}
\newcommand{\cL}{{\cal L}}
\newcommand{\X}{{\cal{X}}}
\newcommand{\Y}{{\cal Y}}
\newcommand{\cH}{{\cal H}}
\newcommand{\cP}{{\cal P}}
\newcommand{\cN}{{\cal N}}
\newcommand{\cU}{{\cal U}}
\newcommand{\cV}{{\cal V}}
\newcommand{\cX}{{\cal X}}
\newcommand{\cY}{{\cal Y}}
\newcommand{\graph}{{\cal H}}
\newcommand{\bayes}{{\cal B}}
\newcommand{\cx}{{\cal X}}
\newcommand{\cg}{{\cal G}}
\newcommand{\cm}{{\cal M}}
\newcommand{\cM}{{\cal M}}
\newcommand{\cG}{{\cal G}}
\newcommand{\cR}{\cal{R}}
\newcommand{\R}{\cal{R}}
\newcommand{\eig}{\mathrm{eig}}

\newcommand{\D}{{\cal D}}
\newcommand{\bfp}{{\bf p}}
\newcommand{\bfd}{{\bf d}}

\newcommand{\cv}{{\cal V}}
\newcommand{\ce}{{\cal E}}
\newcommand{\cy}{{\cal Y}}
\newcommand{\cz}{{\cal Z}}
\newcommand{\cb}{{\cal B}}
\newcommand{\cq}{{\cal Q}}
\newcommand{\cd}{{\cal D}}
\newcommand{\bcf}{{\cal F}}
\newcommand{\cI}{\mathcal{I}}

\newcommand{\ut}{^{(t)}}
\newcommand{\up}{^{(t-1)}}

\newcommand{\bpi}{\boldsymbol{\pi}}
\newcommand{\bphi}{\boldsymbol{\phi}}
\newcommand{\bPhi}{\boldsymbol{\Phi}}
\newcommand{\bmu}{\boldsymbol{\mu}}
\newcommand{\bSigma}{\boldsymbol{\Sigma}}
\newcommand{\bGamma}{\boldsymbol{\Gamma}}
\newcommand{\bbeta}{\boldsymbol{\beta}}
\newcommand{\bomega}{\boldsymbol{\omega}}
\newcommand{\blambda}{\boldsymbol{\lambda}}
\newcommand{\bkappa}{\boldsymbol{\kappa}}
\newcommand{\btau}{\boldsymbol{\tau}}
\newcommand{\balpha}{\boldsymbol{\alpha}}
\def\bgamma{\boldsymbol\gamma}

\newcommand{\prox}{{\mathrm{prox}}}

\newcommand{\pardev}[2]{\frac{\partial #1}{\partial #2}}
\newcommand{\dev}[2]{\frac{d #1}{d #2}}
\newcommand{\dw}{\delta\bw}
\newcommand{\lab}{\mathcal{L}}
\newcommand{\unlab}{\mathcal{U}}
\newcommand{\ind}{1{\hskip -2.5 pt}\hbox{I}}
\newcommand{\ff}[2]{   \cf_{\prec (#1 \rightarrow #2)}}
\newcommand{\vv}[2]{   \cv_{\prec (#1 \rightarrow #2)}}
\newcommand{\dd}[2]{   \delta_{#1 \rightarrow #2}}
\newcommand{\ld}[2]{   \lambda_{#1 \rightarrow #2}}
\newcommand{\en}[2]{  \bD(#1|| #2)}
\newcommand{\ex}[3]{  \bE_{#1 \sim #2}\left[ #3\right]} 
\newcommand{\exd}[2]{  \bE_{#1 }\left[ #2\right]}

\newcommand{\se}[1]{\mathfrak{se}(#1)}
\newcommand{\SE}[1]{\mathbb{SE}(#1)}
\newcommand{\so}[1]{\mathfrak{so}(#1)}
\newcommand{\SO}[1]{\mathbb{SO}(#1)}

\newcommand{\poselow}{\xi}
\newcommand{\pose}{\bm{\poselow}}
\newcommand{\linpose}{\pose^\ell}
\newcommand{\cbpose}{\pose^c}
\newcommand{\rateparam}{v_i}
\newcommand{\bapose}{\bm{\poselow}_i}
\newcommand{\trackingpose}{\bm{\poselow}}
\newcommand{\rotlow}{\omega}
\newcommand{\rot}{\bm{\rotlow}}
\newcommand{\translow}{v}
\newcommand{\trans}{\bm{\translow}}
\newcommand{\hnorm}[1]{\left\lVert#1\right\rVert_{\gamma}}
\newcommand{\lnorm}[1]{\left\lVert#1\right\rVert}
\newcommand{\barate}{v_i}
\newcommand{\trackingrate}{v}
\newcommand{\imgpt}{\mathbf{u}_{i,k,j}}
\newcommand{\mappt}{\mathbf{X}_{j}}
\newcommand{\timet}[1]{\bar{t}_{#1}}
\newcommand{\mf}[1]{\text{MF}_{#1}}
\newcommand{\kmf}[1]{\text{KMF}_{#1}}
\newcommand{\Exp}{\text{Exp}}
\newcommand{\Log}{\text{Log}}

\title{PhysGen: Rigid-Body Physics-Grounded Image-to-Video Generation} 

\titlerunning{PhysGen: Rigid Physics-Grounded Img2Vid}

\author{Shaowei Liu \and Zhongzheng Ren \and Saurabh Gupta* \and Shenlong Wang*}

\authorrunning{S.~Liu et al.}

\institute{University of Illinois Urbana-Champaign \\
\normalsize \url{https://stevenlsw.github.io/physgen/}
}

\maketitle

\footnotetext[0]{*\ Equal advising}

\newcommand{\visvideo}[8]{
    \includegraphics[width=0.16\linewidth]{src_figs/#1/#2/frame_#3.pdf} &
    \includegraphics[width=0.16\linewidth]{src_figs/#1/#2/frame_#4.pdf} &
    \includegraphics[width=0.16\linewidth]{src_figs/#1/#2/frame_#5.pdf} &
    \includegraphics[width=0.16\linewidth]{src_figs/#1/#2/frame_#6.pdf} & 
    \includegraphics[width=0.16\linewidth]{src_figs/#1/#2/frame_#7.pdf} 
    \ifthenelse{\isempty{#8}}{}{ & \includegraphics[width=0.16\linewidth]{src_figs/#1/#2/frame_#8.pdf} \\}
}

\begin{table}
\vspace{-6mm}
    \centering
    \footnotesize
    \resizebox{\linewidth}{!}{
    \setlength{\tabcolsep}{0.1em} %
        \begin{tabular}{c|ccccc}
        {Input} & 
            \multicolumn{5}{c}{{Generation (left$\rightarrow$right: time steps)}} \\
            \visvideo{ours}{car2}{0}{1}{3}{5}{7}{9}
            \visvideo{ours}{balls}{0}{2}{5}{8}{11}{14}
            \visvideo{ours}{domino}{0}{4}{6}{8}{10}{12}
          
    \end{tabular}  
    }

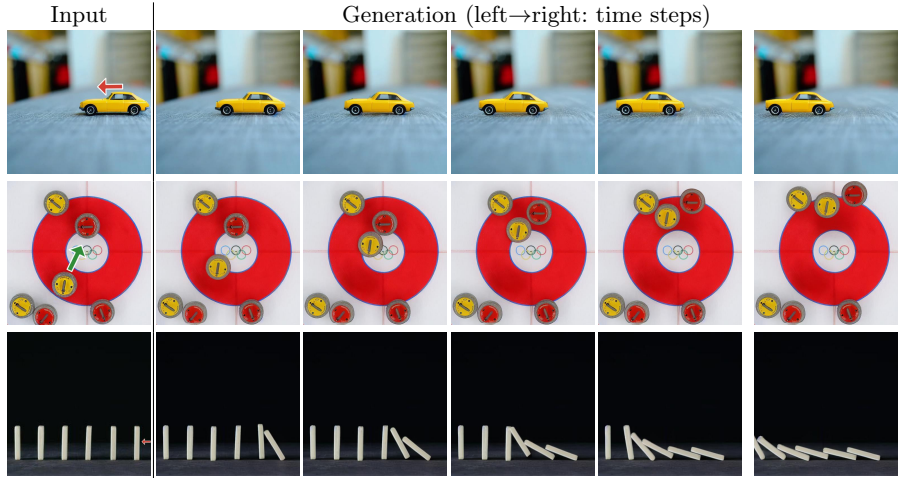
\captionof{figure}{Given a single image, training-free PhysGen generates future frames controlled by physics and initial state. Semantics, geometry and dynamics are reasoned during generation and the result videos are both physics-grounded and photo-realistic.}
\label{fig:teaser}
\vspace{-14mm}
\end{table}

\begin{abstract}

We present PhysGen, a novel image-to-video generation method that converts a single image and an input condition (\eg, force and torque applied to an object in the image) to produce a realistic, physically plausible, and temporally consistent video.
Our key insight is to integrate model-based physical simulation with a data-driven video generation process, enabling plausible image-space dynamics. At the heart of our system are three core components: (i) an image understanding module that effectively captures the geometry, materials, and physical parameters of the image; (ii) an image-space dynamics simulation model that utilizes rigid-body physics and inferred parameters to simulate realistic behaviors; and (iii) an image-based rendering and refinement module that leverages generative video diffusion to produce realistic video footage featuring the simulated motion. The resulting videos are realistic in both physics and appearance and are even precisely controllable, showcasing superior results over existing data-driven image-to-video generation works through quantitative comparison and comprehensive user study. PhysGen's resulting videos can be used for various downstream applications, such as turning an image into a realistic animation or allowing users to interact with the image and create various dynamics. 

\end{abstract}
\section{Introduction}
\label{sec:intro}

Looking at the images in Fig.~\ref{fig:teaser}, we can effortlessly predict or visualize the potential outcomes of various physical effects applied to the car, the curling stone, and the stack of dominos. 
This understanding of physics empowers our imagination of the counterfactual: we can mentally imagine various consequences of an action from an image without experiencing them. We aim to provide computers with similar capabilities -- understanding and simulating physics from a single image and creating realistic animations. We expect the resulting video to produce realistic animations with physically plausible dynamics and interactions.

Tremendous progress has been made in generating realistic and physically plausible video footage. Conventional graphics leverage model-based dynamics to simulate plausible motions and utilize a graphics renderer to simulate images. Such methods provide realistic and controllable dynamics. However, both the dynamics of physics and lighting physics are predefined, making it hard to achieve our goal of animating an image captured in the real world. On the other hand, data-driven image-to-video (I2V) generation techniques~\cite{sora,bar2024lumiere,blattmann2023stable} learn to create realistic videos from a single image through training diffusion models over internet-scale data. However, despite advancements in video generative models~\cite{sora,bar2024lumiere,blattmann2023stable}, the incorporation of real-world physics principles into the video generation process remains largely unexplored and unsolved. Consequently, the synthesized videos often lack temporal coherence and fail to replicate authentic object motions observed in reality. Furthermore, such text-driven methods also lack fine-grained controllability. For instance, they cannot simulate the consequences of different forces and torques applied to an object.

In light of this gap, we propose a paradigm shift in video generation through the introduction of \emph{model-based video generative models}. In contrast to existing purely generative methods, which are trained in a data-driven manner and rely on diffusion models to learn image space dynamics via scaling laws, we propose grounding object dynamics explicitly using rigid body physics, thereby integrating fundamental physical principles into the generation process. As classical mechanics theory reveals, the motion of an object is determined by its physical properties (\emph{e.g.}, mass, elasticity, surface roughness) and external factors (\emph{e.g.}, external forces, environmental conditions, boundary conditions).

We tackle our image-to-video generation problem via a computational framework, PhysGen. PhysGen consists of three stages: 1) image-based physics understanding; 2) model-based dynamics simulation; and 3) generative video rendering with dynamics guidance. 
Our method first infers object compositions and physical parameters from the input image through large visual foundation model-based reasoning (Sec.~\ref{sec:perception}). 
Given an input force, we then utilize the inferred physical parameters to perform realistic dynamic simulations for each object, accounting for their rigid body dynamics, collisions, frictions, and elasticity (Sec.~\ref{sec:phy-world-sim}). Finally, guided by motion and the input image, we present a novel generative video diffusion-based renderer that outputs the final realistic video footage (Sec.~\ref{sec:render}). 
Several illustrative examples of our framework are shown in Fig.~\ref{fig:teaser}. As shown in this figure, our approach combines the best of the worlds of data-driven generative modeling's appearance realism and model-based dynamics's verified physical plausibility. 
Notably, our generation process is highly controllable, allowing users to specify underlying physics parameters and initial conditions. This capability facilitates a range of interactive applications. Moreover, our proposed generation pipeline operates solely during inference time, eliminating the need for any training.

We evaluate PhysGen generation ability on multiple data source including the web and self-captured images. We compare against both state-of-the-art image-to-video models~\cite{chen2023seine, 2023I2VGen-XL, xing2023dynamicrafter} as well as image editing method~\cite{geng2024motion} quantitatively and qualitatively. We also perform user-study to evaluate the physical-realism and photo-realism of the generated video.
Our method demonstrates physics-informed results for controllable image-to-video generation, producing realistic video sequences with grounded rigid-body dynamics and interaction. PhysGen combines learning-based generative approaches with traditional model-based physics to deliver visually appealing results without any training. Our approach harnesses a plethora of ingredients from previous works, particularly recent progress in large foundational visual models for segmentation~\cite{kirillov2023segany, liu2023grounding, ren2024grounded, yu2023inpaint}, physical understanding~\cite{achiam2023gpt4, gpt4v}, normal estimation~\cite{eftekhar2021omnidata, fu2024geowizard}, relighting, and generative rendering~\cite{blattmann2023stable, chen2023seine}. While these individual ingredients exist, our system combines them in a novel manner to enable an exciting new capacity, featuring physical plausibility for video generation at an unprecedented level. 

Our contributions can be summarized as follows:
1) We propose a novel image-space dynamics model that can understand physical parameters from a single image and simulate realistic, controllable, and interactive dynamics.
2) We present a novel image-to-video system by integrating our presented dynamics model with generative diffusion-based video priors. The model is training-free, and the resulting videos are highly realistic, physically plausible, and controllable, consistently surpassing existing state-of-the-art video generative models.

\section{Related work}
\label{sec:related}

\paragraph{\textbf{Image-based physics reasoning and simulation.}}

Our image-based dynamics model is built upon physical simulation. Symbolic physics simulators or physics engines have been extensively studied for various physical processes, including rigid body dynamics, fluid simulation, and deformable objects~\cite{todorov2012mujoco, physx, Koenig-2004-394, hu2018moving, hu2019taichi, hu2019difftaichi, liu2013fast}. Such methods have been widely used across graphics, scientific discovery, robotics, and video effects. However, these rule-based or solver-based simulators face limitations in terms of expressiveness, efficiency, generalizability, and parameter tuning. To overcome these challenges, neural physics, which combines traditional physics simulators/engines with deep neural networks, has been developed to model dynamic processes such as planar pushing and bouncing~\cite{physplus}, as well as object interactions~\cite{sain}. Domain-specific network architectures~\cite{mrowca2018flexible,li2018learning,battaglia2016interaction,sanchez2018graph} have been developed to address various physical problems. Additionally, the integration of neural scene representations or large language models with physics simulators~\cite{li2023pac,xie2023physgaussian, lv2024gpt4motion} has also been investigated recently. 
Nevertheless, most forward physics models, whether neural or not, have not resolved the dependency on pre-defined physical parameters, which are often specified by users. 

Reasoning physical parameters from visual data allows us to simulate without manually defined physics, further enhancing convenience and realism for image-based physics simulation. To enable this, subsequent research efforts~\cite{li2020visual, vda, phys101, galileo, DensePhysNet, davis2015visual, li2023pac, Video2Game} have introduced pipelines that first extract scene representations of the physical world from images or videos using neural networks, and then utilize these representations for either physics reasoning or simulation. 
Despite the promise of these approaches, they primarily rely on synthetic data for training, which might suffer from generalization issues, or on inverse physics, which requires inference from observed physical phenomena, often requiring video observations instead of a single image. 
In our work, we aim to generate physics-grounded, high-fidelity predictive videos from static input images of real-world objects with complex backgrounds. Contrary to existing work, we investigate leveraging the power of large pretrained visual foundational models~\cite{kirillov2023segany, achiam2023gpt4, gpt4v} for physics reasoning in a zero-shot manner.

\paragraph{\textbf{Video generative model.}}
Video generative models have experienced remarkable advancements in recent years~\cite{bar2024lumiere,sora,blattmann2023stable,blattmann2023videoldm,ge2023preserve,girdhar2023emu,ho2022imagen,kondratyuk2023videopoet,gupta2023photorealistic,yu2023language,villegas2022phenaki,singer2022make,wang2023lavie, chen2023seine, xing2023dynamicrafter, 2023I2VGen-XL}, with the state-of-the-art framework~\cite{sora} achieving the capability to generate photo-realistic and coherent imaginative videos from text instructions using diffusion models~\cite{sohl2015deep,song2019generative,ho2020denoising,rombach2022high,Peebles2022DiT}. 
Despite progress, challenges remain, such as the need for extensive training data, object permanence, realistic lighting, and reducing unrealistic motion and physics violations. Overcoming these is key for advancing video generative modeling and its applications, especially in physics-accurate areas like scientific discovery and robotics. Our work addresses some issues by proposing a novel image-to-video framework that merges a model-based approach with a pretrained video prior, grounding the generated content with real-world physics.

\paragraph{\textbf{Image animation.}}

Our method is heavily inspired by image-based animation~\cite{chuang2005animating, davis2015image, xue2018visual, holynski2021animating, schodl2000video,Szummer96a,wei2000fast} and cinemagraphs. 
One common way to improve quality is to incorporate class-specific heuristics~\cite{jhou2015animating,chuang2005animating}, which can be physically simulated and integrated into the generation process. 
Recent advancements in deep learning have led to data-driven solutions in this domain, where temporal neural networks are trained on video datasets to directly predict consecutive video frames from an input image~\cite{holynski2021animating,Blattmann_2021_CVPR,endo2019animating,chen2023livephoto,2023VideoComposer,guo2023animatediff}. 
Moreover, there is growing interest in interactive~\cite{blattmann2021ipoke,li2023generative,blattmann2021understanding} and controllable~\cite{davis2015image,aberman2020unpaired} image-to-video synthesis. To further improve temporal coherence, more priors such as motion fields~\cite{endo2019animating,holynski2021animating,mahapatra2022controllable,mallya2022implicit,sugimoto2022water,xue2018visual} or flows~\cite{geng2024motion,zhao2022thin,bowen2022dimensions}, 3D geometry information~\cite{weng2019photo,qiu2023relitalk}, and user annotations~\cite{li2023generative} have been subsequently introduced. 
In this work, we propose a model-based solution and tackle the image animation task from a new perspective, where our method takes a single image and user-specific initial force or torque as input, and models object dynamics explicitly via rigid physics simulation.  Our key insight is that physical parameters can be inferred from a single image automatically through large foundation models such as GPT-4V~\cite{achiam2023gpt4, gpt4v}.

\section{Approach}
\label{sec:approach}

\begin{figure}[t]
\centering
\includegraphics[width=0.9\textwidth]{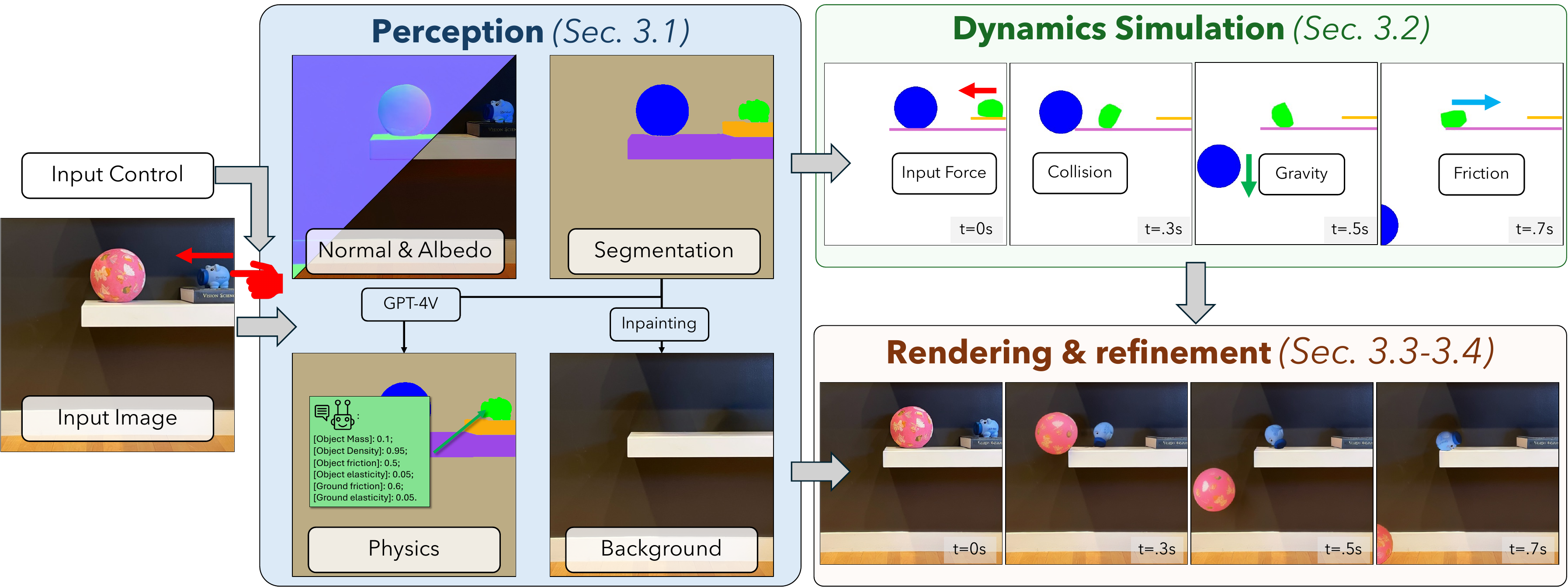}
\caption{\textbf{Method overview.} Our framework consists of three interleaved components: the perception module, the dynamics simulation module, and the rendering module. The perception module interprets the semantics, geometry, and physical parameters in the given image. The dynamics simulation module simulates the rigid-body motion and interactions of each instance in the scene, governed by Newton's Laws and physical constraints. The rendering module renders the final outcome, leveraging an off-the-shelf relighting model and diffusion-model-based video priors.}
\label{fig:archi}
\end{figure}

Given a single input image, our goal is to generate a realistic $T$-frame video %
featuring rigid body dynamics and interactions. Our generation can be conditioned on user-specific inputs in the form of an initial force $\mathbf{F}_0$ and/or torque $\boldsymbol{\tau}_0$. Our key insight is to incorporate physics-informed simulation into the generation process, ensuring the dynamics are realistic and interactive, thus producing large, plausible object motion. To achieve this, we present a novel framework consisting of three stages: physical understanding, dynamics generation, and generative rendering. The perception module aims to reason about the semantics, collision geometry, and physics of the objects presented in the image (Sec.~\ref{sec:perception}). The dynamics module then leverages the input force/torque, as well as the inferred shape and geometry, to simulate the rigid-body motions of the objects, as well as object-object and object-scene interactions, such as friction and collisions (Sec.~\ref{sec:phy-world-sim}). Finally, we convert the simulated dynamics into pixel-level motion and combine image-based warping and generative video to render the final video (Sec.~\ref{sec:render}). Fig.~\ref{fig:archi} depicts the overall framework.

\subsection{Perception}
\label{sec:perception}

Simulating dynamics on an image requires a holistic understanding of object compositions, materials, geometry, and physics. Toward this goal, we designed our perception module to infer such properties from a single image. Our key insight is to harness readily available, large pretrained models~\cite{ren2024grounded, fu2024geowizard, 
careaga2023intrinsic, achiam2023gpt4, gpt4v} to achieve this goal, as shown in \cref{fig:archi}.

\paragraph{\textbf{Segmentation.}}
\label{para:seg}
Identifying and segmenting each individual physical entity lays the foundation for image-based dynamics. Inspired by the recent success of large pretrained segmentation models, we incorporate GPT-4V~\cite{achiam2023gpt4, gpt4v} to recognize all image categories and send to Grounded-SAM~\cite{ren2024grounded} to detect and segment each individual instance-level objects $\{\mathbf{o}^i \in \mathbb{R}^{W\times H}\}_i^N$, where $\mathbf{o}^i$ is the binary mask for $i$-th object. We also query the GPT-4V to classify the objects into foreground and background based on their movability. The foreground objects are send to the physical simulation once user inputs are applied. We extract collision boundaries and supporting edges (e.g., ground, walls, etc.) from the background objects. The details are presented in the \cref{sec:implement}.

\paragraph{\textbf{Physical properties reasoning.}}
Realistic physics dependents on accurate physical parameters, such as surface friction, mass, and elasticity. Unlike prior physics-based image dynamics work~\cite{chuang2005animating}, our work leverages visual foundational models to reason physical properties directly. Our solution is simple yet effective. Inspired by the success of mask-based prompting~\cite{achiam2023gpt4, gpt4v, yang2023set}, we directly ask GPT-4V~\cite{gpt4v} for certain physical properties, providing an object mask overlaid on the input image. Following~\cite{albert2024phynerf, hong2024video2game}, we send the crafted prompt to GPT-4V, which contains GPT-4V instructions to return a quantitative measure of each queried property in a metric unit. For each object, we query its mass, elasticity, and friction coefficient $(M_i, E_i, \mu_i)$. 
We use the Coulomb friction model~\cite{popova2015research} for friction modeling. The elasticity coefficient is a scalar, where 0.0 results in no bounce, and 1.0 results in a perfect bounce. Given each object mass $M_i$ in grams and its shape primitive, we compute the rotational inertia $\mathbf{I}_i$ accordingly.

\paragraph{\textbf{Geometry primitives.}}
Rasterized instance masks are not suitable for physical simulation; hence, we need to convert each object into vectorized shape primitives. We fit two types of primitive shapes: a circle and generic polygons. For each object, we choose the primitive type with the maximum coverage of its segmentation. For circles, we fit the center and radius to cover the segmentation mask. Circles allow us to realistically simulate rolling motions. 
For non-circle objects, we instead fit the generic polygons. Specifically, we perform contour extraction on the segmentation masks to obtain the polygon vertices.

\paragraph{\textbf{Intrinsic decomposition.}}
\label{para:intrinsic}
The dynamic movement of the objects will also result in comprehensive changes in shading, as their position wrt the lighting environment changes. To compensate for this effect during rendering, we must perform image decomposition to infer albedo, normal for each object $(\mathbf{A}_0^i, \mathbf{N}_0^i)$ and background scene lighting $\mathbf{L}$
from the single image. To achieve this we leverage off-the-shelf intrinsic decomposition model~\cite{careaga2023intrinsic} to compute $\mathbf{A}_0^i$ and surface normal estimator~\cite{fu2024geowizard} to compute $\mathbf{N}_0^i$. We model $\mathbf{L}$ as directional light source and estimate $\mathbf{L}$ using the optimization proposed in~\cite{careagaCompositing}.

\paragraph{\textbf{Background inpainting.}}
\label{para:inpaint}
Once the foreground objects move, they will leave holes in the background. To produce realistic and complete video footage, we leverage an off-the-shelf generative image inpainting model~\cite{yu2023inpaint} to recover the complete background scene without foreground objects. The inpainted image $\mathbf{B} \in \mathbb{R}^{W\times H\times 3}$ will be used for composition in the rendering stage.

\subsection{Image space dynamics simulation}
\label{sec:phy-world-sim}
Given the foreground objects with physical properties, we use rigid-body physics for dynamics simulation in image spaces. We choose to perform simulations in image spaces for three reasons: 1) image-space dynamics are better coupled with our output video; 2) a full 3D simulation requires complete 3D scene reconstruction and understanding, which remains an unsolved problem from a single image; 3) image-space dynamics can already cover various object dynamics and have been widely used in prior work~\cite{physplus, phys101, chuang2005animating, holynski2021animating, li2023generative}. 
Specifically, at a given time $t$, each rigid object $i$ is characterized by its 2D pose and velocity at its center of mass. The position includes a translation $\mathbf{t}^i(t) \in \bbR^2$ and a rotation $\mathbf{R}^i(t) \in \SO2$ specified in a world coordinate. The velocity includes linear velocity $\boldsymbol{\nu}^i(t) \in \bbR^2$ and angular velocity $\boldsymbol{\omega}^i(t) \in \bbR^2$. Hence, the state of the each object $i$ at time $t$ can be represented as $\mathbf{q}^i(t) = \left[\mathbf{t}^i(t), \mathbf{R}^i(t), \boldsymbol{\nu}^i(t), \boldsymbol{\omega}^i(t) \right]$. 

Following~\cite{friedland2012control}, the rigid body motion dynamics is given by ~\cref{eq:newton} for each object. For simplicity, we omit the object index $i$ in the following: 
\begin{equation}
\label{eq:newton}
    \frac{d}{dt}\mathbf{q}(t) =
    \frac{d}{dt}\begin{bmatrix}
        \mathbf{t}(t) \\
        \mathbf{R}(t) \\
        \boldsymbol{\nu}(t) \\
        \boldsymbol{\omega}(t)
      \end{bmatrix} = 
      \begin{bmatrix}
        \mathbf{v}(t) \\
        \boldsymbol{\omega}(t) \times \mathbf{R}(t) \\
        \frac{\mathbf{F}(t)}{M} \\
        {\mathbf{I}}(t)^{-1}(\boldsymbol{\tau} - \boldsymbol{\omega}(t) \times \mathbf{I}(t)\boldsymbol{\omega}(t))
      \end{bmatrix}
\end{equation}
where $\mathbf{F}$ is the force, $\boldsymbol{\tau}$ is the torque, $M$ is the mass of the object, $\mathbf{I}(t) = \mathbf{R}(t)\mathbf{I}{\mathbf{R}}(t)^{T}$ is the rotation inertia in world coordinates, $\mathbf{I} \in \bbR^{2 \times 2}$ is the inertia matrix in its body coordinates. This inertia indicates its resistance against rotational motion. The state $\mathbf{q}(t)$ is given by integrating \cref{eq:newton}: 
\begin{equation}
\label{eq:integral}
    \mathbf{q}(t) = \mathbf{q}(0) + \int_{0}^{t}\frac{d}{dt}\mathbf{q}(t)dt = \mathbf{q}(0) + \sum_{i=1}^{T} \frac{d}{dt}\mathbf{q}(t)|_{t=t_i}\mathit{\Delta}_t
\end{equation}
where $\mathbf{q}(0)$ is the initial condition specified in the input image. We compute the integral using numerical ODE integrations, \eg Euler method~\cite{ciarlet1990handbook}. Given the initial state of the objects, the physical motion simulation module synthesis each foreground object future motion. The location of each object is updated by the affine transformation $\mathbf{T}(t)=\left[ \mathbf{R}(t), \mathbf{t}(t)\right]$ from the initial location. 

\paragraph{\textbf{External force and torque.}}
For each object, the external forces \(\mathbf{F}(t)\) and torques $\boldsymbol{\tau}$ include gravity, friction, and elasticity between the object's surface and the environment. Specifically, we consider the initial external force and torque from user input, as well as gravity, rolling friction, and sliding friction.

\paragraph{\textbf{Collision.}}
When objects move, they might collide with each other or with the scene boundary (e.g., hitting a wall). Therefore, collision checking is necessary at every time step. A collision will result in offset forces being applied from its center of mass, producing \(\boldsymbol{\tau}\) and causing the object to rotate. In collision reactions, changes in energy and momentum are minimized to adhere to Newton's conservation laws. In the case of an object falling, it could bounce back or start rolling on the ground due to the elasticity of both the object and the ground.
\begin{table}[t]
    \centering
    \scriptsize
    \resizebox{.85\linewidth}{!}{
    \setlength{\tabcolsep}{0.1em} %
    \begin{tabular}{cccc}
         \includegraphics[width=0.25\linewidth]{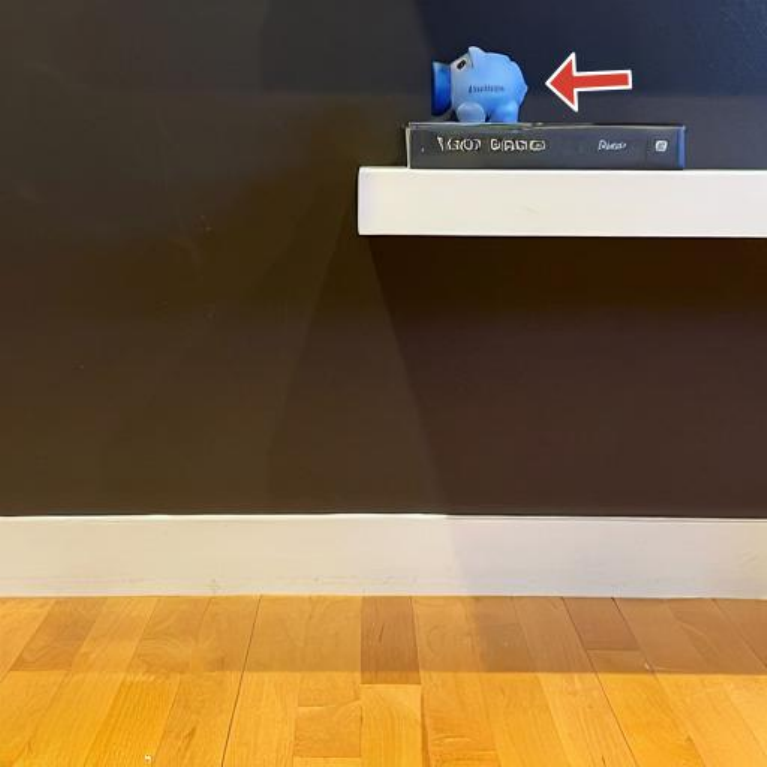} &
         \includegraphics[width=0.25\linewidth]{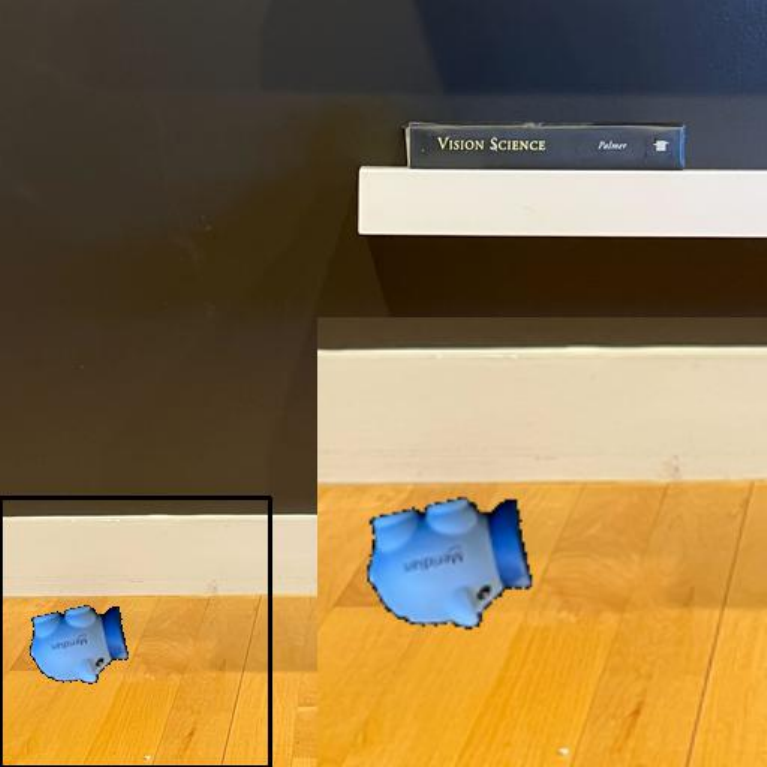} &
        \includegraphics[width=0.25\linewidth]{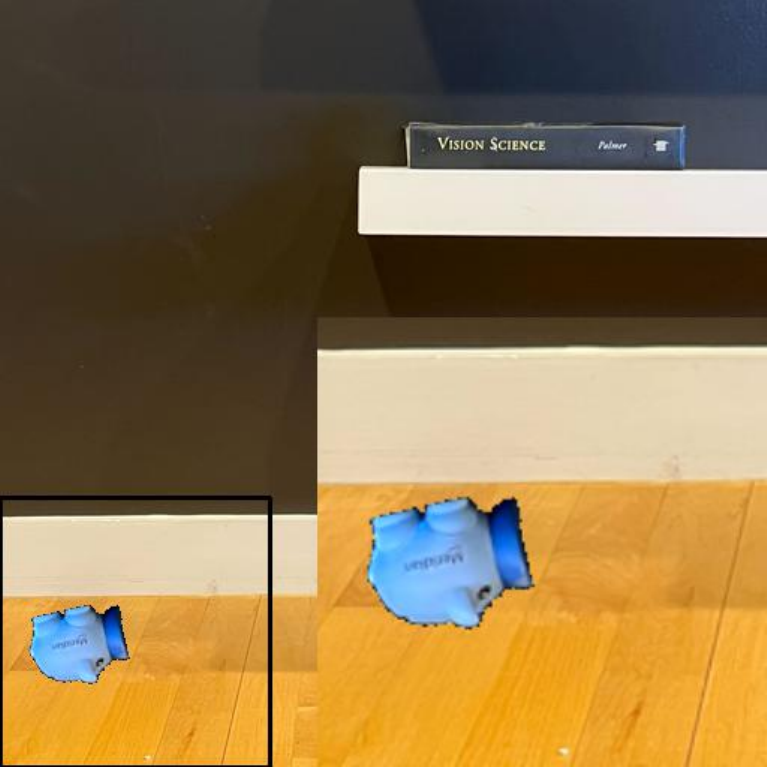} &
         \includegraphics[width=0.25\linewidth]{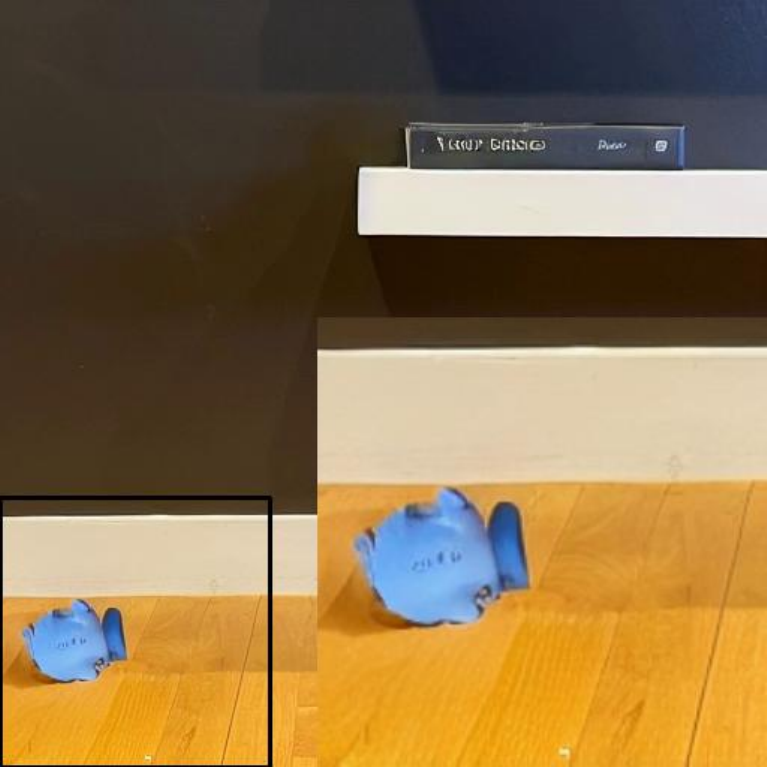} \\
         Input & Composited~(\cref{para:composite})  & Relight~(\cref{para:relight}) & Render~(\cref{para:video_prior})
    \end{tabular}
    }
\captionof{figure}{\textbf{Rendered video comparison.} We show a toy example of piggy bank. The left shows the input frame, and the rest 3 are future frame generations. The composited piggy bank hasn't been aware of the light change. The relighted output synthesized no shadows beneath. The rendered output from diffusion model is most photo realistic.}
\label{fig:render_ex}
\end{table}

\subsection{Rendering}
\label{sec:render}
Given the simulated motion sequence \(\{\mathbf{q}^i(0), \dots, \mathbf{q}^i(t), \dots\}_i^N\) and the input image \(\mathbf{C}\), the rendering module outputs the rendered video \(\{\dots, \mathbf{C}(t), \dots\}\). There are a few desiderata for generating realistic video: it needs to be temporally consistent, complete, reflect the lighting changes as the object moves, and accurately represent the simulated image space dynamics. To achieve this goal, we design a novel motion-guided rendering pipeline consisting of an image-based transformation and composition module to ensure motion awareness, a relighting module to ensure the plausibility of lighting physics and a generative video editing module that retouches the composed and relit video to ensure realism.

\paragraph{\textbf{Composited video.}}
\label{para:composite}
Given the object state from the physical motion module, we compose an initial video by alpha-blending the foreground scene with the static inpainted background from~\cref{para:inpaint}. The foreground scene is rendered by performing forward-warping from the input image to future frames using the affine transformation \(\mathbf{T}(t)\). The alpha channel is computed using the same procedure, with the input being the segmentation mask in~\cref{para:seg}: 
$$
\hat{\mathbf{X}}(t) = \mathtt{composite}(\mathbf{B}, 
\mathtt{warp}(\mathbf{X}^0, \mathbf{T}^0(t))\dots
\mathtt{warp}(\mathbf{X}^i, \mathbf{T}^i(t))\dots..)
$$
where $\mathbf{X}^i$ are segmented input image of $i$-th object. Apart from the RGB sequence $\hat{\mathbf{V}} = \{\dots\hat{\mathbf{X}}(t)\dots \} $, we also use the same affine warping and composition to compute the blended albedo map $\hat{\mathbf{A}}(t)$ and surface normal $\hat{\mathbf{N}}(t)$ from $\hat{\mathbf{A}}(0)$ and $\hat{\mathbf{N}}(0)$ respectively, which is later used for relighting. 

\paragraph{\textbf{Relighting.}}
\label{para:relight}
Our relighting module simulates the changes in shading due to object movement. Specifically, it takes the RGB image $\hat{\mathbf{X}}(t)$, the transformed albedo $\hat{\mathbf{A}}(t)$ and surface normal $\hat{\mathbf{N}}(t)$ at each time $t$, as well as the estimated directional light $\mathbf{L}$ as input. We then perform reshading using the the Lambertian shading model $f$ for foreground objects: $\tilde{\mathbf{X}}(t) = f(\hat{\mathbf{X}}(t), \hat{\mathbf{A}}(t), \hat{\mathbf{N}}(t), \mathbf{L})$, which returns the relit image $\tilde{\mathbf{X}}(t)$ for each frame $t$.

\subsection{Generative refinement with latent diffusion models}
\label{para:video_prior}
One limitation of the single-image based relighting model is that it neglects the background and cannot handle complicated lighting effects due to complex object-object interactions, such as ambient occlusion and cast shadows. Moreover, the composition results in unnatural boundaries. To address these issues and enhance realism, we incorporated a diffusion-based video to refine the relit video, denoted as \(\tilde{\mathbf{V}} = \{\dots, \tilde{\mathbf{X}}(t), \dots\}\), and obtain the final video output.

Specifically, we incorporate a pretrained video diffusion model to refine our video \(\tilde{\mathbf{V}}\). We first use the pretrained latent diffusion encoder~\cite{rombach2022high} to encode the guidance video \(\tilde{\mathbf{V}}\) into a latent code \(\mathbf{z}\). Inspired by SDEdit~\cite{meng2022sdedit}, we add noise to the guided latent code and gradually denoise it using the pretrained video diffusion. Given that the content in the guided latent code is already satisfactory, we do not perform the denoising process from scratch. Instead, we define an initial noise strength \(s\) where \(s \in [0, 1]\) controls the amount of noise added to \(\mathbf{z}\). In practice, we find that a certain approach finds a good trade-off between fidelity and realism. At each denoising step, we ensure the foreground objects are as consistent as possible with the guidance (copied from the guided latent code) while synthesizing new content (e.g., shadows) in the background (using the denoised latent code). To this end, we use a fusion weight \(w\) to control how much of the generated latent to inject into the foreground; the fusion weight is gradually increased during the denoising as the perturbation of noise decreases. We also define a fusion timestamp \(\delta\) as the signal to stop the fusion. The final denoised latent code \(\mathbf{z}_\ast\) is then decoded to our final video output \(\mathbf{V}\). Please see ~\cref{sec:supp_algorithm} for a detailed algorithm.

\cref{fig:render_ex} depicts the composed video \(\hat{\mathbf{V}}\), the relit video \(\tilde{\mathbf{V}}\), and our final output \(\mathbf{V}\), illustrating the effectiveness of each rendering component.

\section{Experiments}
\label{exp}

\paragraph{\textbf{Implementation details.}}
For physical body simulation, we use a 2D rigid body physics simulator Pymunk~\cite{pymunk}. For each experiment, we run 120 simulation steps and uniformly sample 16 frames to form the videos. We set the generation resolution at $512 \times 512$ so that frames can easily fit into the diffusion models. For video diffusion prior model, we use SEINE~\cite{chen2023seine}. For inference, DDIM sampling~\cite{song2020denoising} is used and the noise strength i set to $s=0.5$, \ie, executing 25/50 steps for denoising. The latents fusion stops at timestamp $\delta=5$.

\paragraph{\textbf{Speed.}}
PhysGen's run-time speed is 3 minutes for one image-to-video gen on a Nvidia A40 GPU. Image understanding and relighting takes the most time. 

\paragraph{\textbf{Data.}}
To show the generalizability and robustness of our method, we use both internet data and self-captured indoor images from a cell phone. The dataset is diverse, with variations in lighting, object count, geometries, physical attributes, and environmental boundaries. We compare different approaches on 15 photos that require physical reasoning. Additionally, we collect 50 recorded videos of a given scene as ground truth by varying input conditions for quantitative comparison. Randomly selected sequences are shown in \cref{sec:supp_quantitative}.

\begin{table}[t]
    \centering
    \footnotesize
    \resizebox{\linewidth}{!}{
    \setlength{\tabcolsep}{0.2em} %
        \begin{tabular}{c|ccccc}
        {Input} & \multicolumn{5}{c}{{Generation (left$\rightarrow$right: time steps)}} \\
           \visvideo{ours}{house}{0}{3}{6}{9}{11}{13}
           \visvideo{ours}{wall_toy}{0}{2}{3}{4}{6}{8}
           \visvideo{ours}{wall_bowl}{0}{1}{2}{3}{4}{5}
           \visvideo{ours}{boxes}{0}{1}{4}{8}{11}{14}
           \visvideo{ours}{book_bottle}{0}{1}{3}{5}{6}{8}
           \visvideo{ours}{pig_ball}{0}{1}{3}{5}{7}{9}
    \end{tabular}  
    }
    
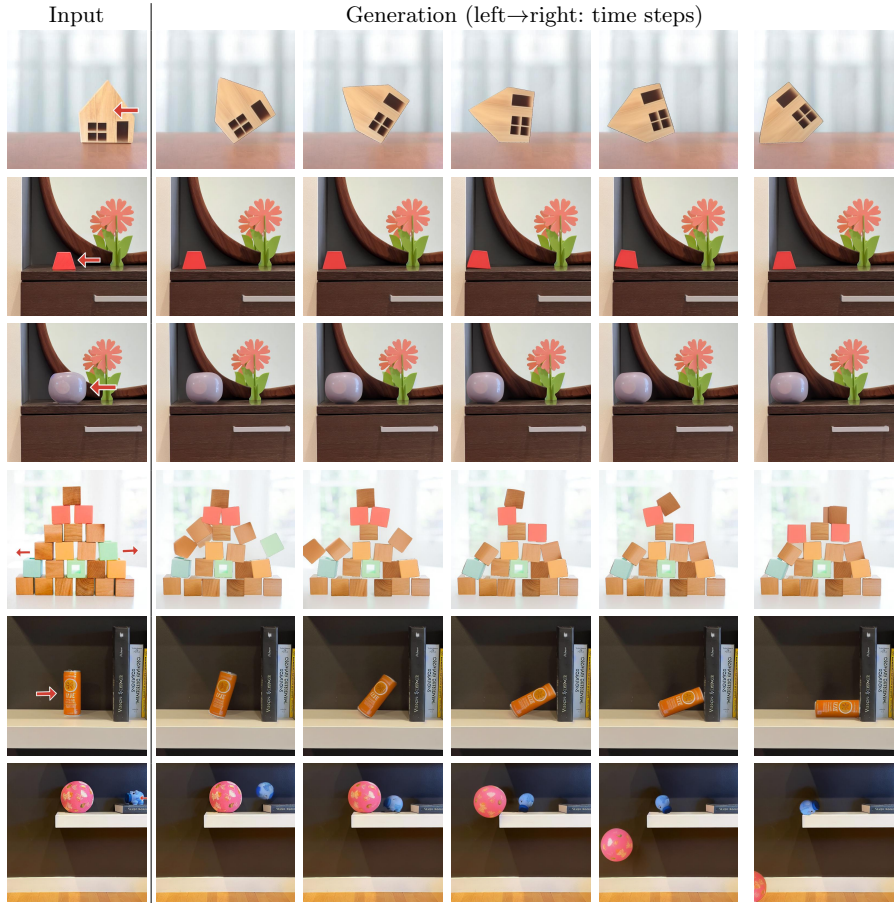
\captionof{figure}{\textbf{Video generation results.} Left: Input initial frame with \textcolor{red}{red arrows} representing the initial force or speed on the object. Right: Generated future frames.}
    \label{fig:exp}
    \end{table}
\paragraph{\textbf{Baselines.}}
We compare against two streams of approaches. The first stream is current state-of-the-art I2V models: SEINE~\cite{chen2023seine}, I2VGen-XL~\cite{2023I2VGen-XL} and DynamiCrafter~\cite{xing2023dynamicrafter}. These methods take both an image and a text prompt as input for generation. The text prompt is generated by GPT-4V~\cite{gpt4v} to describe the future events. Another stream we compare is the image-based manipulation approach Motion Guidance~\cite{geng2024motion}. 
This method uses an image and optical flow as inputs to predict future movement and generate corresponding images. We employ computed motion flow from our physical motion module as the target input for Motion Guidance to optimize each video frame.

\subsection{Results}
\label{sec:results}
We show our generated results on 8 different physical procedures from multiple sources in ~\cref{fig:exp}. 
Our method could simulate reasonable complex physical procedures which involves object-object interaction, object-scene interaction with different physical properties and initial conditions.

\paragraph{\textbf{Visual comparisons.}}
The comparison results are shown in \cref{fig:exp_compare}. As can be seen, image-video generation methods could not produce any physically plausible futures. For each method, even we carefully tune the input prompt, the models couldn't understand the direction, physical meanings and interactions. All 3 models fail to synthesize realistic videos. For motion guidance with accurate motion flow provided, the results is better but brings a lot inconsistency. When there are multiple objects interaction, the method suffer from introducing new contents. For the domino case, the generation is not smooth and contain artifacts as it could not accurately model the physical procedure.

\begin{table}[t]
    \centering    
     \small
    {
    \setlength{\tabcolsep}{0.4em} %
    \begin{tabular}{l|cc}
         Methods & \bf  Physical-realism$\uparrow$ &  \bf Photo-realism$\uparrow$ \\
         \midrule
        SEINE~\cite{chen2023seine} & 1.39 & 1.86\\ DynamiCrafter~\cite{xing2023dynamicrafter} & 1.68 & 1.81 \\
        
        I2VGen-XL~\cite{2023I2VGen-XL} & 2.11 & 2.25\\
        
        Ours & \bf 4.14 & \bf 3.86\\
    \end{tabular}}
\caption{\textbf{Human evaluation.} We evaluate both physical-realism and photo-realism of the three competing image-to-video models. The user is asked to evaluate the generated video using a 5-point scale from strongly disagree (1) to strongly agree(5) that the video is physical-realistic and photo-realistic. Our method ranks the first in both terms.}
\label{table:user}
\end{table}
\begin{table}[t]
    \centering
    \small
    {
    \setlength{\tabcolsep}{0.8em} %
    \begin{tabular}{l|cc}
         Methods & \bf  Image-FID$\downarrow$ &  \bf Motion-FID$\downarrow$ \\
         \midrule
        SEINE~\cite{chen2023seine} & 138.89  & 38.76 \\
    DynamiCrafter~\cite{xing2023dynamicrafter} & \bf 99.64 & 109.61 \\
        I2VGen-XL~\cite{2023I2VGen-XL} & 104.62 & 82.04 \\
        Ours &  105.70 &  \bf 30.20 \\
    \end{tabular}}
\caption{\textbf{Quantitative evaluation.} We measure the Image-FID and Motion-FID of videos generated by four methods against the collected GT videos for the given scene. 
Our method achieves both low Image-FID and Motion-FID.}
\label{table:fid}
\end{table}

\begin{table}
    \centering
    \footnotesize
    \resizebox{\linewidth}{!}{
    \setlength{\tabcolsep}{0.2em} %
    \begin{tabular}{c|cccccc}
        {Input} & 
        \multicolumn{5}{c}{{Generation (left$\rightarrow$right: time steps)}} \\
    & \visvideo{dynamcrafter}{pool}{2}{5}{8}{11}{14}{} & \rotatebox{270}{\hspace{-40pt}DC~\cite{xing2023dynamicrafter}} \\
    
     & \visvideo{i2vgen-xl}{pool}{1}{4}{7}{11}{15}{} & \rotatebox{270}{\hspace{-40pt}VGen~\cite{2023I2VGen-XL}} \\

    \includegraphics[width=0.16\linewidth]{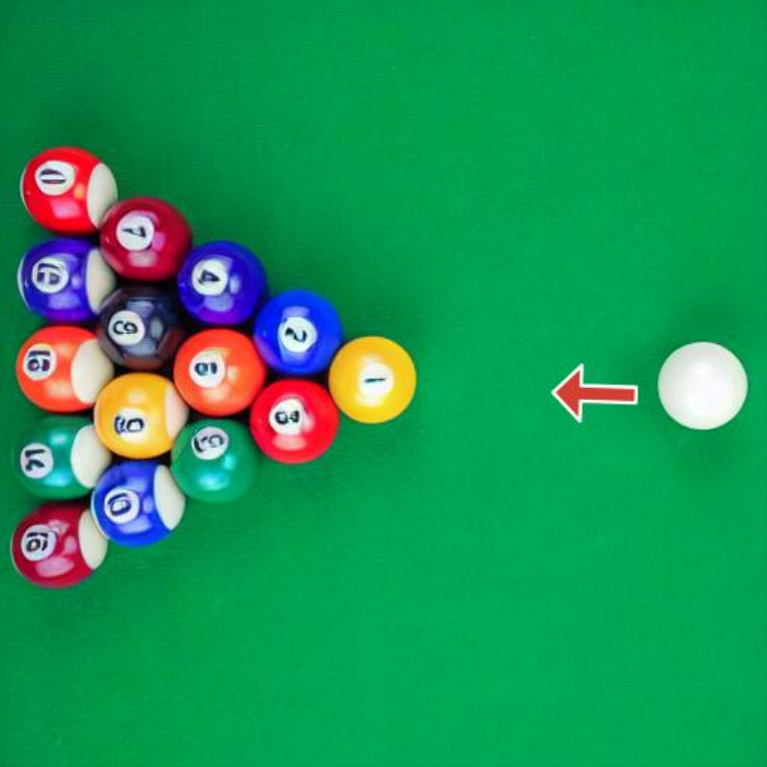} & \visvideo{seine}{pool}{1}{4}{8}{12}{15}{}
    & \rotatebox{270}{\hspace{-45pt}SEINE~\cite{chen2023seine}} \\

    & \visvideo{motion_guidance}{pool}{1}{4}{7}{10}{13}{} & \rotatebox{270}{\hspace{-40pt}MG~\cite{geng2024motion}} \\

     & \visvideo{ours}{pool}{1}{4}{7}{10}{13}{} & \rotatebox{270}{\hspace{-30pt}Ours} \\ %
    
      & \visvideo{dynamcrafter}{domino}{1}{3}{7}{10}{13}{} & \rotatebox{270}{\hspace{-40pt}DC~\cite{xing2023dynamicrafter}} \\
    
     & \visvideo{i2vgen-xl}{domino}{1}{4}{7}{10}{14}{} & \rotatebox{270}{\hspace{-40pt}VGen~\cite{2023I2VGen-XL}} \\

    \includegraphics[width=0.16\linewidth]{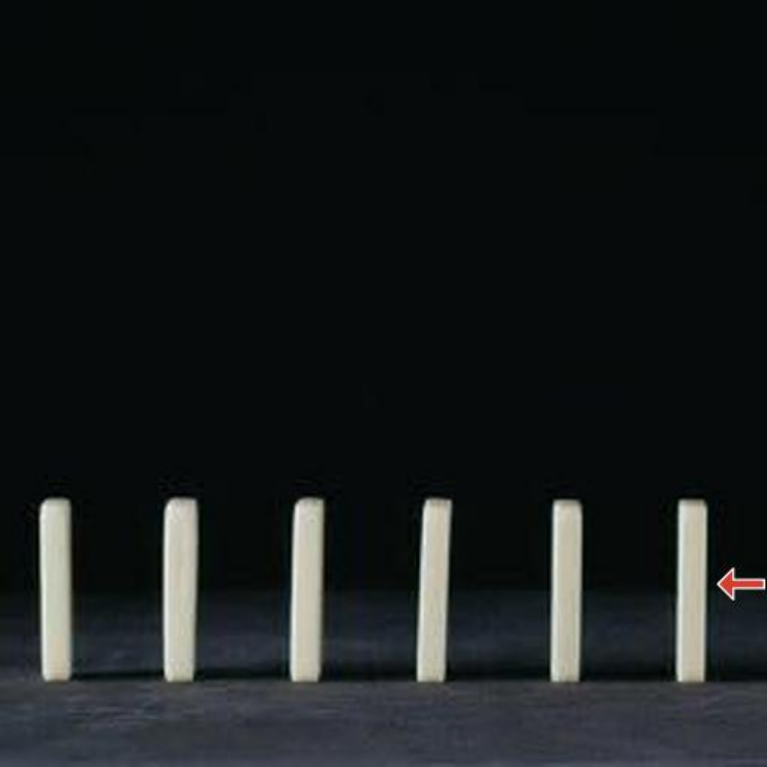} & \visvideo{seine}{domino}{3}{6}{9}{12}{15}{} & \rotatebox{270}{\hspace{-45pt}SEINE~\cite{chen2023seine}} \\

    & \visvideo{motion_guidance}{domino}{4}{6}{8}{10}{12}{} & \rotatebox{270}{\hspace{-40pt}MG~\cite{geng2024motion}} \\

     & \visvideo{ours}{domino}{4}{6}{8}{10}{12}{} & \rotatebox{270}{\hspace{-30pt}Ours} \\
    
    \end{tabular} 
    }
\captionof{figure}{\textbf{Qualitative comparison.} against DynamiCrafter(\textbf{DC})~\cite{xing2023dynamicrafter}, I2VGen-XL(\textbf{VGen})~\cite{2023I2VGen-XL}, \textbf{SEINE}~\cite{chen2023seine}, Motion Guidance(\textbf{MG})~\cite{geng2024motion}.}
\label{fig:exp_compare}
\end{table}
\begin{table}
    \centering
    \footnotesize
    \resizebox{\linewidth}{!}{
    \setlength{\tabcolsep}{0.2em} %
    \begin{tabular}{c|ccccc}
        {Input} & 
        \multicolumn{5}{c}{{Generation (left$\rightarrow$right: time steps)}} \\
    \includegraphics[width=0.16\linewidth]{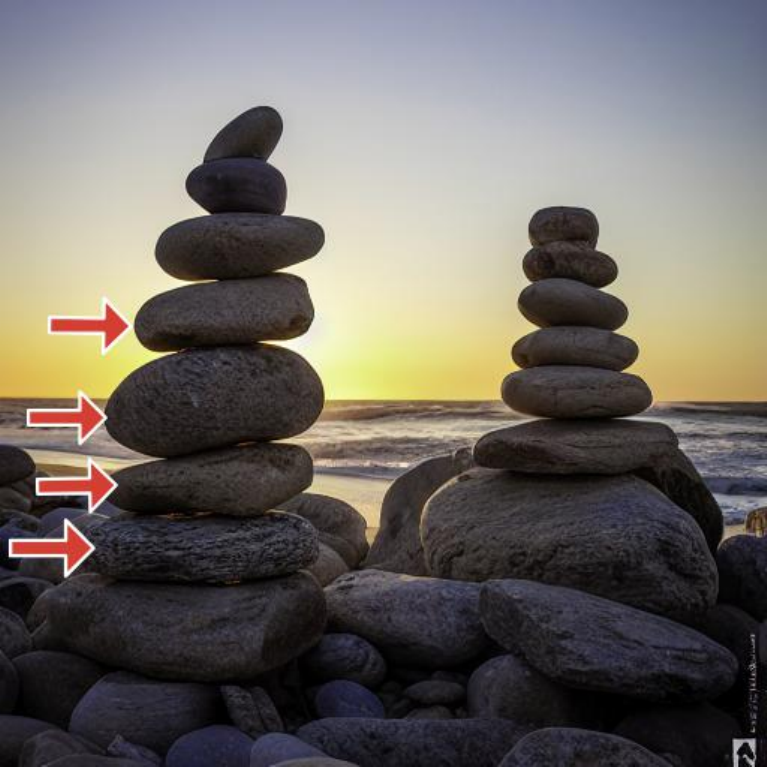} & \visvideo{ours}{stone1}{1}{4}{7}{10}{15}{} \\
    \includegraphics[width=0.16\linewidth]{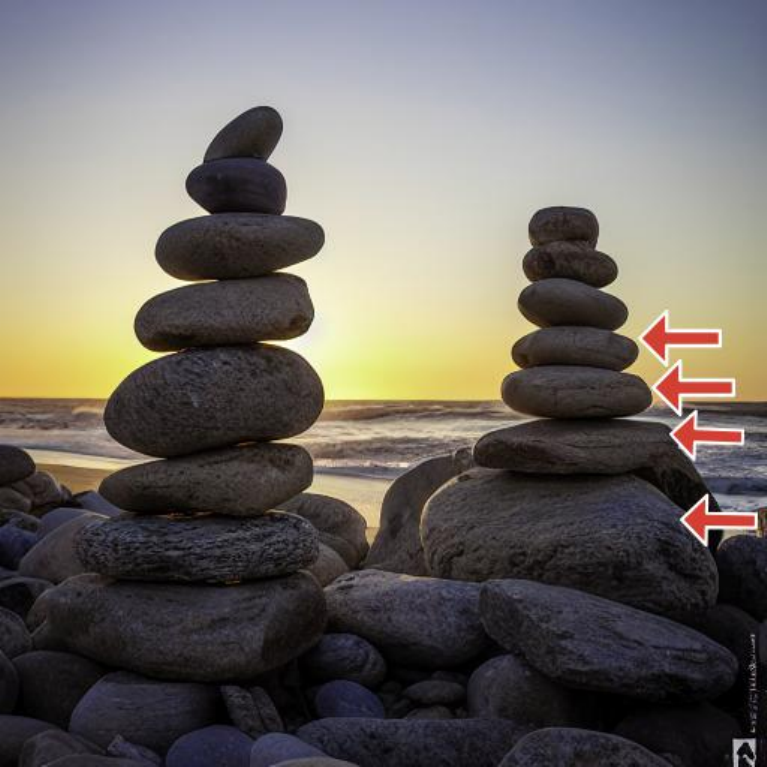} & \visvideo{ours}{stone1_v2}{2}{6}{9}{12}{15}{} \\
    \end{tabular}
    }
\captionof{figure}{\textbf{Controllability}. PhysGen generates diverse results reflecting initial forces.
}
\label{fig:exp_controllable}
\end{table}

\begin{table}
    \centering
    \footnotesize
    {
    \setlength{\tabcolsep}{0.2em} %
    \begin{tabular}{c|ccccc}
        {Input}& FG Seg & BG Seg  & Inpainting & Normal & inferred Physics\\
        \includegraphics[width=0.16\linewidth]{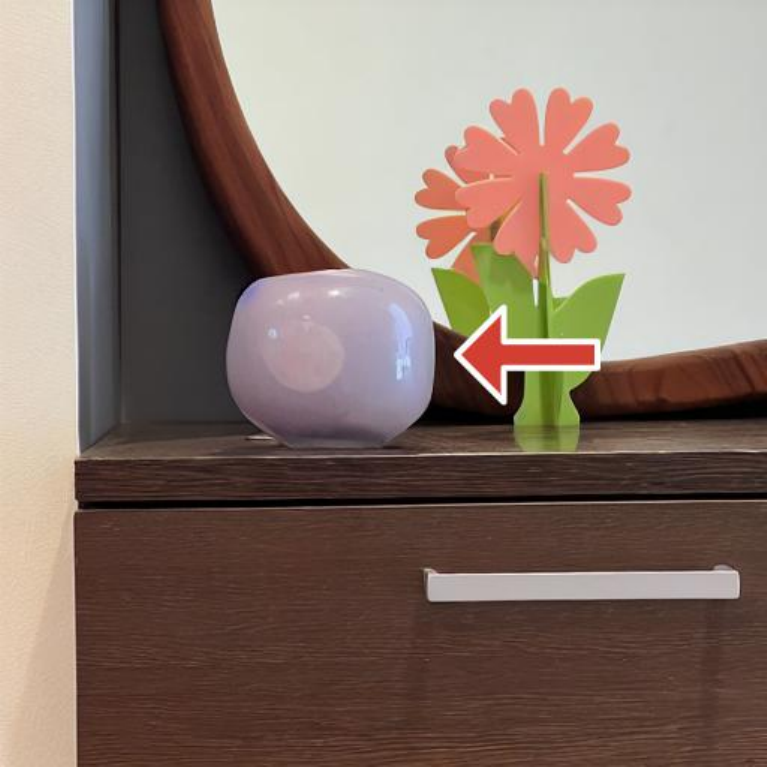} & 
        \includegraphics[width=0.16\linewidth]{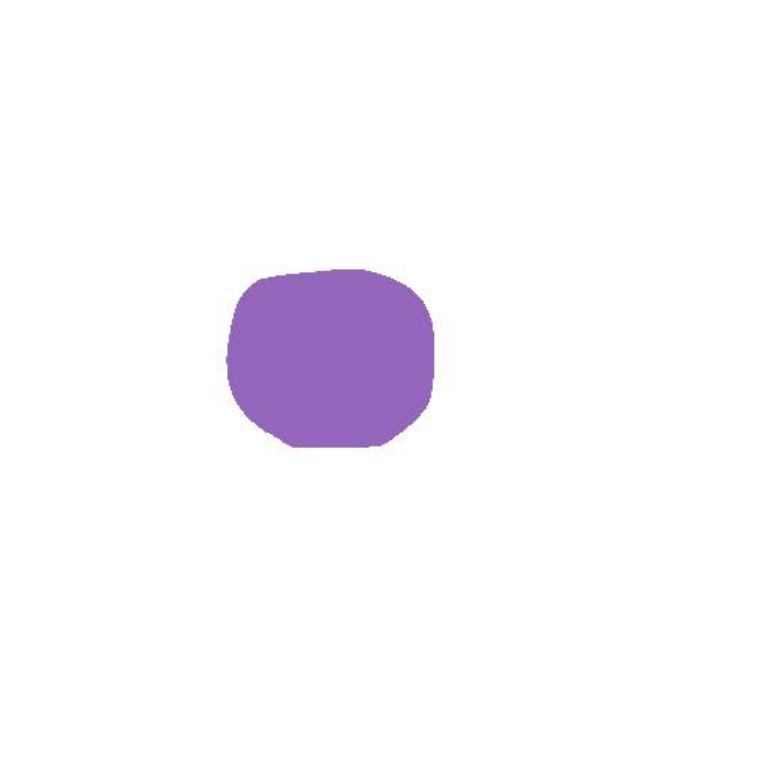} & 
        \includegraphics[width=0.16\linewidth]{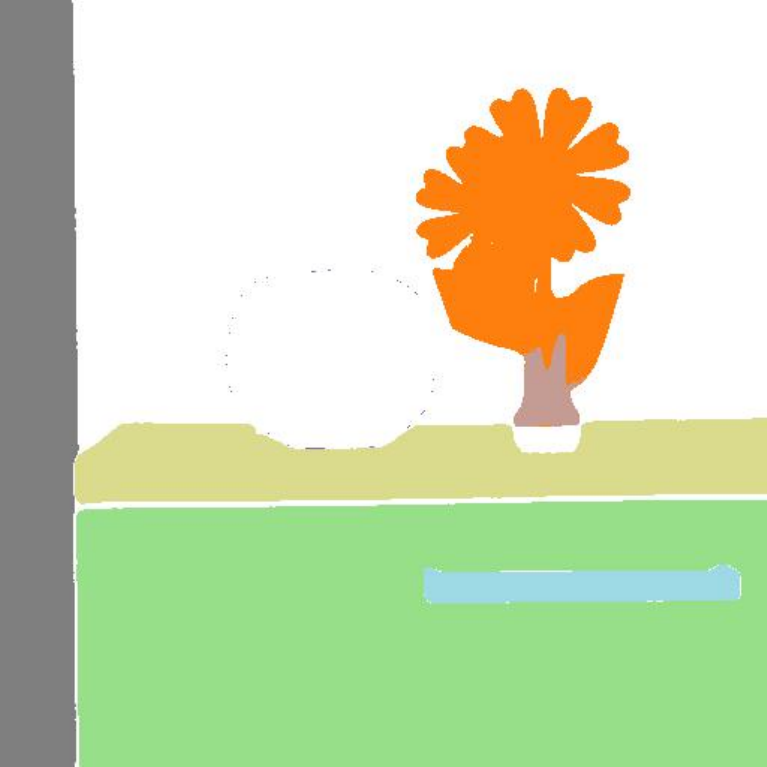} & 
        \includegraphics[width=0.16\linewidth]{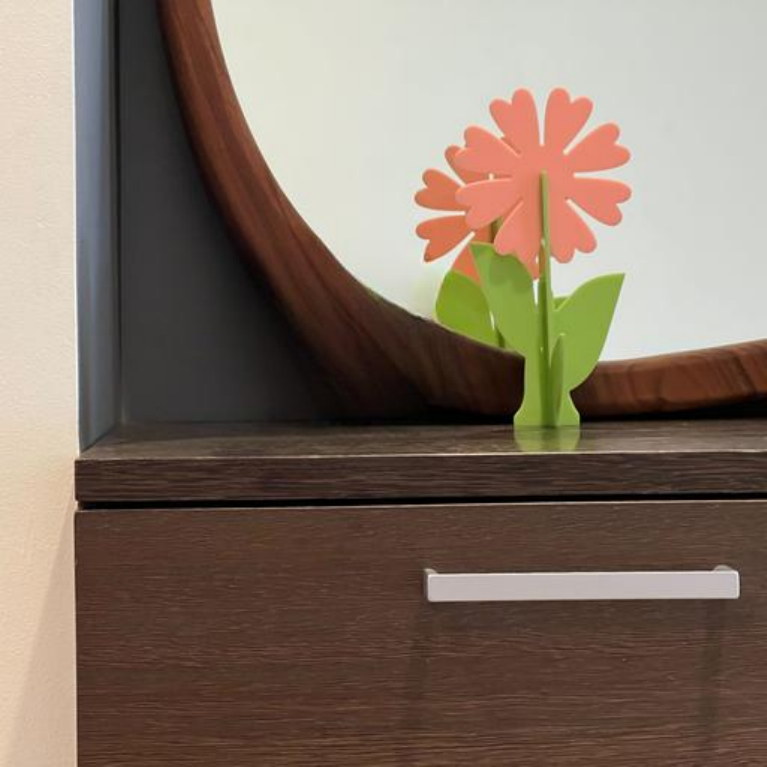} &
        \includegraphics[width=0.16\linewidth]{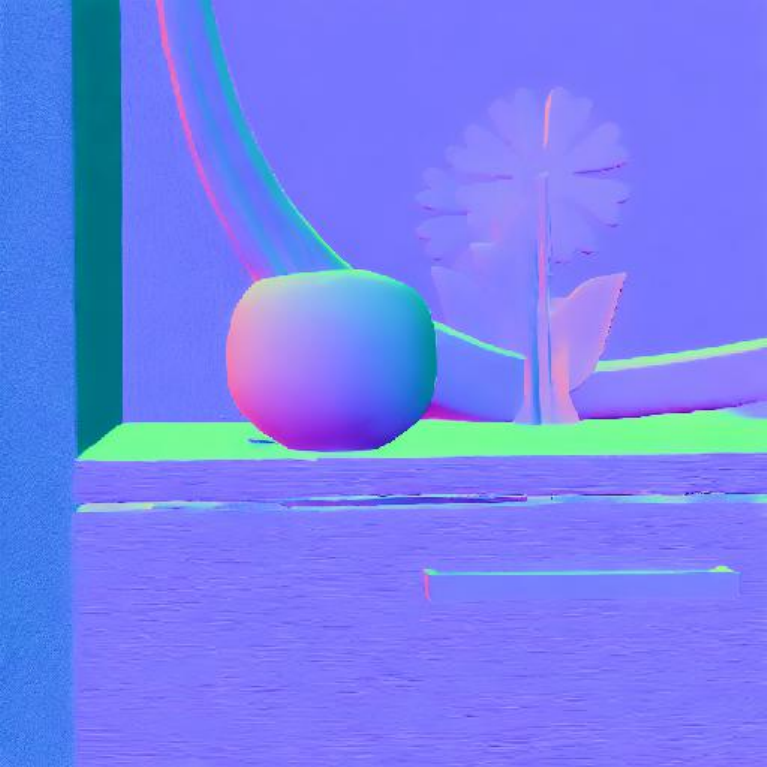} &
        \includegraphics[width=0.16\linewidth]{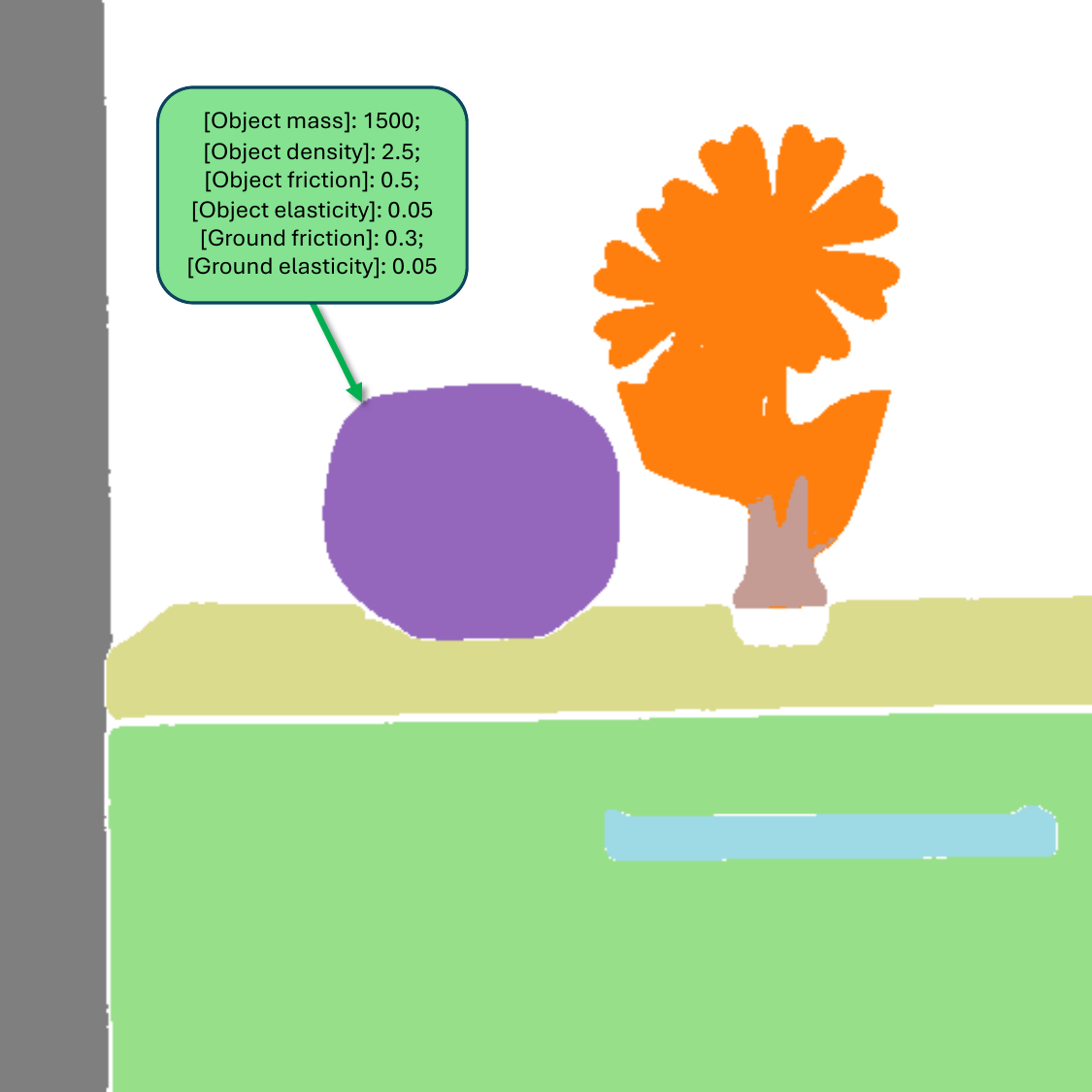} \\
        
    \end{tabular}
    }
\captionof{figure}{\textbf{Qualitative results from the perception module}. The results include foreground and background mask, inpainting output, normal map and inferred physics.}
\label{fig:exp_perception}
\end{table}

\subsection{Human evaluation}
To comprehensively assess both the \textbf{physical-realism} and \textbf{photo-realism} of the generated videos, we conducted human evaluations following a similar methodology to prior work~\cite{chen2023videocrafter1, yu2023animatezero, 2023DreamVideo, wu2023lamp}.
Here by physical-realism we mean whether the content of the generated video is realistic in its dynamics and object interactions and accurately reflects the instructed movements. 

Specifically, we include 15 videos with diverse objects and conditions for evaluation, compared against other three I2V model generations (SEINE~\cite{chen2023seine}, DynamiCrafter~\cite{xing2023dynamicrafter}, I2VGen-XL~\cite{2023I2VGen-XL}). For each compared baseline method, we meticulously adjusted the input prompts and conducted multiple runs to select the most plausible output. In contrast, our method did not undergo specific tuning for each video. 
We randomized the presentation order and asked 14 participants to rate the videos on a five-point scale, from strongly disagree (1) to strongly agree (5) for physical-realism and photo-realism. The results in ~\cref{table:user} show that our method achieves the highest ratings. Our average scores for physical-realism ($4.14$) and photo-realism ($3.86$) fall within the Agree level, while other methods perform poorly.

\subsection{Quantitative evaluation}
\label{sec:quantitative}
We quantitatively evaluate the collected GT videos of a given scene. The GT data includes 833 images. For each method, we generate 50 videos with random samples, collecting 800 images per method. Our method varies initial conditions to ensure all generated videos are different. Following~\cite{blattmann2023videoldm}, we adopt the Fréchet Inception Distance (FID)~\cite{heusel2017gans} as evaluation metric. We didn't use Fréchet Video Distance (FVD)~\cite{unterthiner2018towards} due to the small sample size and the goal to disentangle evaluation of appearance and motion. We evaluate appearance using \textbf{Image-FID} and motion using \textbf{Motion-FID}. For Motion-FID, we use RAFT~\cite{teed2020raft} to extract and colorize optical flow~\cite{baker2011database}, then compute FID from these rendered images. The results are shown in \cref{table:fid}. We notice other methods which either generate minimal motion, leading to low Image-FID and high Motion-FID (e.g., DynamiCrafter~\cite{xing2023dynamicrafter}), or produce realistic motion but fail to maintain image consistency, resulting in low Motion-FID and high Image-FID (e.g., SEINE~\cite{chen2023seine}). Our approach achieves both low Image-FID and Motion-FID compared to others.

\subsection{Analysis}

\noindent{\textbf{Perception evaluation.}}
We evaluate the perception module using 10 complex open-world images with 118 annotated movable instances (COCO~\cite{lin2014microsoft} format). Our system achieves \textbf{0.93} precision, \textbf{0.82} recall at 0.5 IoU. More details are shown in the ~\cref{sec:supp_experiments}. Fig.~\ref{fig:exp_perception} shows intermediate outputs from our perception module and simulation module.

\paragraph{\textbf{Physical reasoning.}} We compare the physical parameters estimated by GPT-4V to random values sampled from human-estimated ranges. Without GPT-4V, Image-FID increases from \textbf{105.70} to \textbf{111.01}, Motion-FID from \textbf{30.20} to \textbf{36.60}. It shows the physical reasoning are crucial for appearance and motion realism.

\paragraph{\textbf{Controllability.}}
We showcase the controllability and diversity of our method in \cref{fig:exp_controllable} by varying the initial conditions.
For different rows, we adjust the direction and magnitude of initial speed and forces. The results show diverse physical-plausible trajectories. This demonstrates the potential of applying our method for diverse physical world generations.

\paragraph{\textbf{Limitations.}}
Our method is an initial step towards physics-based video generation but has two main limitations: it focuses mainly on rigid objects, limiting its application to non-rigid ones, and it lacks a comprehensive 3D understanding, making it unable to handle out-of-plane motions. We leave leveraging deformable physics and full 3D understanding as future work.
 
\section{Conclusion}
\label{sec:conclusion}

We present PhysGen, a novel image-to-video generation method that turns a static image into a high-fidelity, physics-grounded, and temporally coherent videos. The main module is a model-based physical simulator with a data-driven video generation process.
PhysGen enables a series of controllable and interactive downstream applications. PhysGen provide a new image-to-video generation paradigm and hopefully can inspire more follow-up works.

\section*{Acknowledgements}
\label{sec:ack}
{This project is supported by NSF Awards \#2331878, \#2340254, and \#2312102, the IBM IIDAI Grant, and an Intel Research Gift. We greatly appreciate the NCSA for providing computing resources. We thank Tianhang cheng for helpful discussions. We thank Emily Chen and Gloria Wang for proofreading.}

\bibliographystyle{splncs04}
\bibliography{arxiv}

\appendix
\clearpage
\title{Supplementary Material -- PhysGen: Rigid-Body Physics-Grounded Image-to-Video Generation}

\titlerunning{PhysGen: Rigid Physics-Grounded Img2Vid}

\author{Shaowei Liu \and Zhongzheng Ren \and Saurabh Gupta* \and Shenlong Wang*}

\authorrunning{S.~Liu et al.}

\institute{University of Illinois Urbana-Champaign \\
\normalsize \url{https://stevenlsw.github.io/physgen/}
}

\maketitle

\footnotetext[0]{*\ Equal advising}

\begin{abstract}
    
The supplementary material provides implementations, additional analysis and experiments, as well as discussion of limitations in detail. In summary, we include
\begin{itemize}
    \item \cref{sec:implement}. More implementation details of each module in our pipeline. 
    \item \cref{sec:supp_analysis}. More experiment details besides the main paper.
    \item \cref{sec:supp_experiments}. More experiments on key components in our framework.
    \item \cref{sec:supp_qualitative}. Additional qualitative  comparisons and controllable generation results.
    \item \cref{sec:limitation}. Generation limitation analysis of our current approach.
    
\end{itemize}
\end{abstract}

\section{Implementation Details}
\label{sec:implement}

\subsection{Perception}

\paragraph{\textbf{Segmentation.}}
We use GPT-4V to recognize all objects in the given image and determine if they are movable. Movable objects are treated as foreground objects, while non-movable objects are treated as background objects. The query prompt is shown in \cref{fig:caption_prompt}. The outputs from GPT-4V are sent to Grounded-SAM~\cite{ren2024grounded} for instance segmentation. We also enable non-maximum suppression (NMS) to prevent overlapping segmentation. For foreground objects, we check if the segmentation is fully connected within its mask. If not, we rerun Grounded-SAM to further separate any poor segmentation from the first round. 

\paragraph{\textbf{Boundary extraction.}}
To extract boundaries from non-movable background objects, we utilize depth and normal information estimated from GeoWizard~\cite{fu2024geowizard}. We first order the background objects according to relative depth estimation and select only those whose depth range falls within that of the foreground objects. For candidate objects, we extract their segmentation boundaries within the foreground object's depth range and use corresponding normal to determine if they are planes. If so, we fit horizontal or vertical edges to the boundaries and use it as the physical boundary of the scene.

\begin{figure}[h]
\begin{tcolorbox}[colback=white]
\footnotesize
\textbf{User:} Describe all unique object categories in the given image, ensuring all pixels are included and assigned to one of the categories, do not miss any movable or static object appeared in the image, each category name is a single word and in singular noun format, do not include '-' in the name. Different categories should not be repeated or overlapped with each other in the image. For each category, judge if the instances in the image is movable, the answer is True or False. If there are multiple instances of the same category in the image, the judgement is True only if the object category satisfies the following requirements: 1. The object category is things (objects with a well-defined shape, e.g. car, person) and not stuff (amorphous background regions, e.g. grass, sky, largest segmentation component). 2. All instances in the image of this category are movable with complete shape and fully-visible. 

Format Requirement:
You must provide your answer in the following JSON format, as it will be parsed by a code script later. Your answer must look like:
{
    "category-1": False,
    "category-2": True

}
Do not include any other text in your answer. Do not include unnecessary words besides the category name and True/False values. 
\end{tcolorbox}
\caption{Prompt used for GPT-4V image recognition and movability judgment.}
\label{fig:caption_prompt}
\end{figure}

\begin{figure}[h]
\begin{tcolorbox}[colback=white]
\footnotesize
\textbf{User:} You will be given an image and a binary mask specifying an object on the image, analyze and provide your final answer of the object physical property. The query object will be enclosed in white mask. The physical property includes the mass, the friction and elasticity. The mass is in grams. The friction uses the Coulomb friction model, a value of 0.0 is frictionless. The elasticity value of 0.0 gives no bounce, while a value of 1.0 will give a perfect bounce. 

Format Requirement:
You must provide your answer in the following JSON format, as it will be parsed by a code script later. Your answer must look like:
{
    "mass": number,
    "friction": number,
    "elasticity": number
}
The answer should be one exact number for each property, do not include any other text in your answer, as it will be parsed by a code script later.
\end{tcolorbox}
\caption{Prompt used for GPT-4V image recognition and movability judgment.}
\label{fig:physics_prompt}
\end{figure}

\paragraph{\textbf{Physical properties reasoning.}}
The query prompt for reasoning the physical properties of a foreground object is shown in \cref{fig:physics_prompt}.

\paragraph{\textbf{Geometry primitives.}}
Given a object segmentation mask, we automatically choose the proper primitive that best fits the object. We first use a circle to fit the corresponding segmentation mask and compute the Intersection over Union (IoU) between the fitted mask and segmentation mask. If IoU is smaller than $0.85$, we switch to the generic polygons by extracting the contour of the segmentation. 

\paragraph{\textbf{Background Inpainting.}}
Given the foreground masks of the input image, we use off-the-shelf image inpainting model~\cite{yu2023inpaint} to recover the background scene. Considering foreground objects might have shadow beneath, we dilate the foreground segmentation mask by a kernel size of 40 pixels. To this end, we aim to get a clean background image without shadows. However, if the input image is heavy-shadowed, the inpainting model could not remove it completely. We discuss it use a detailed example in \cref{sec:limitation}. 

\begin{table}[t]
    \centering
    \footnotesize
    \resizebox{\linewidth}{!}{
    \setlength{\tabcolsep}{0.2em} 
        \begin{tabular}{c|ccccc}
        {Input} & \multicolumn{5}{c}{{Simulation (left$\rightarrow$right: time steps)}} \\
           \includegraphics[width=0.16\linewidth]{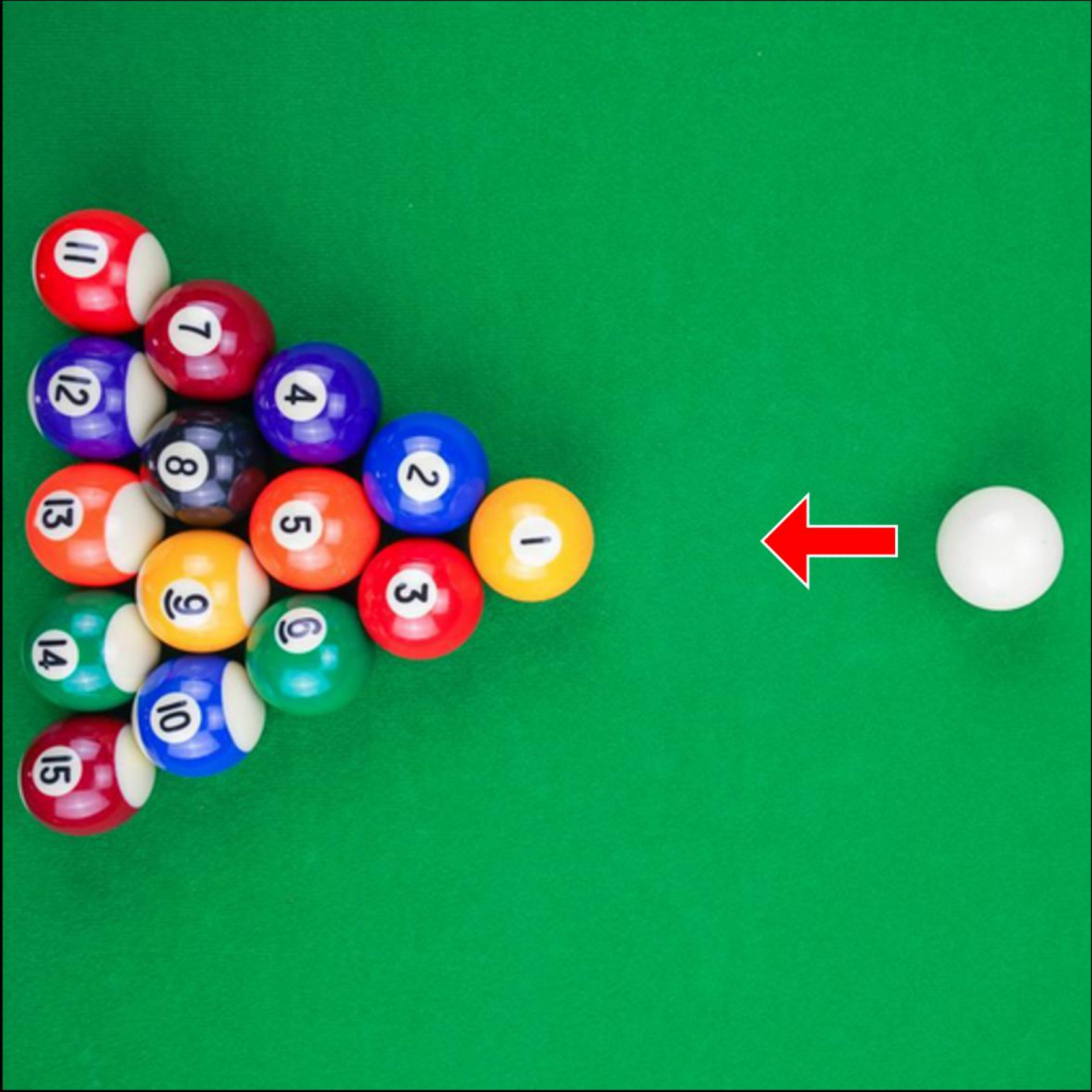} & \visvideo{ours}{pool/snapshots}{1}{4}{7}{10}{13}{} \\

           \includegraphics[width=0.16\linewidth]{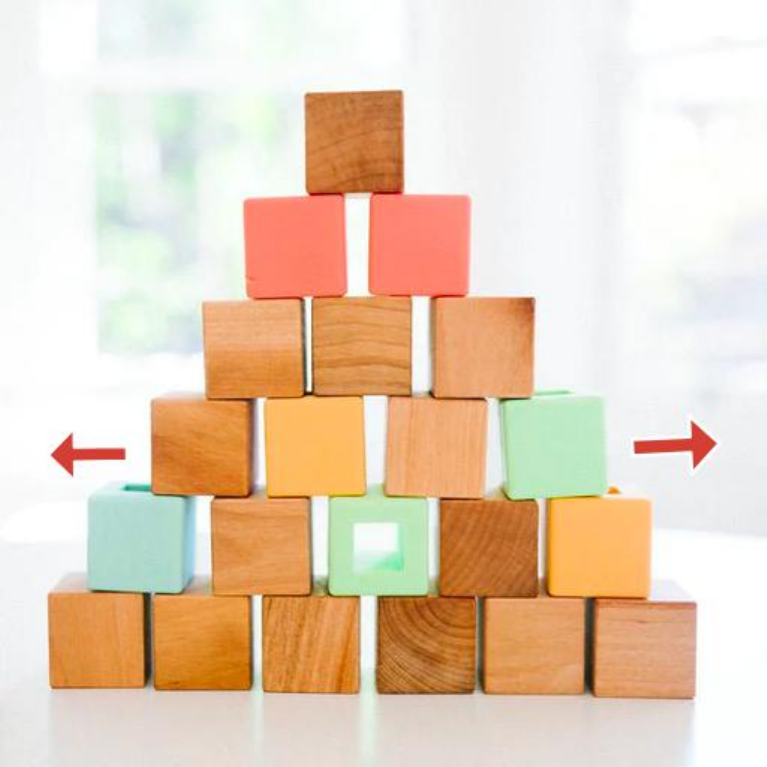} & \visvideo{ours}{boxes/snapshots}{1}{4}{8}{11}{14}{} \\
           
    \end{tabular}  
    }
    \captionof{figure}{\textbf{Image space dynamics simulation example.} We show two simulation examples with different primitives. The primitives are fitted from the segmentation mask of the input image and we perform dynamic simulation on those primitives. The \textbf{\textcolor{pink}{edges}} show the physical boundary.}
    \label{fig:supp_phy}
    \end{table}
\begin{table}[t]
    \centering
    \scriptsize
    \resizebox{.85\linewidth}{!}{
    \setlength{\tabcolsep}{0.1em} 
    \begin{tabular}{cccc}
         \includegraphics[width=0.25\linewidth]{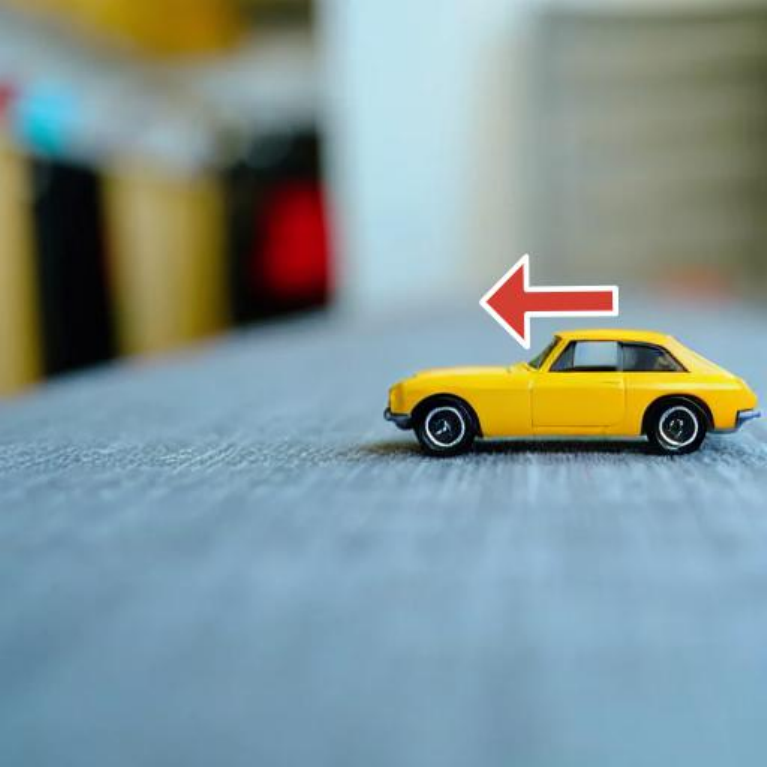} &
         \includegraphics[width=0.25\linewidth]{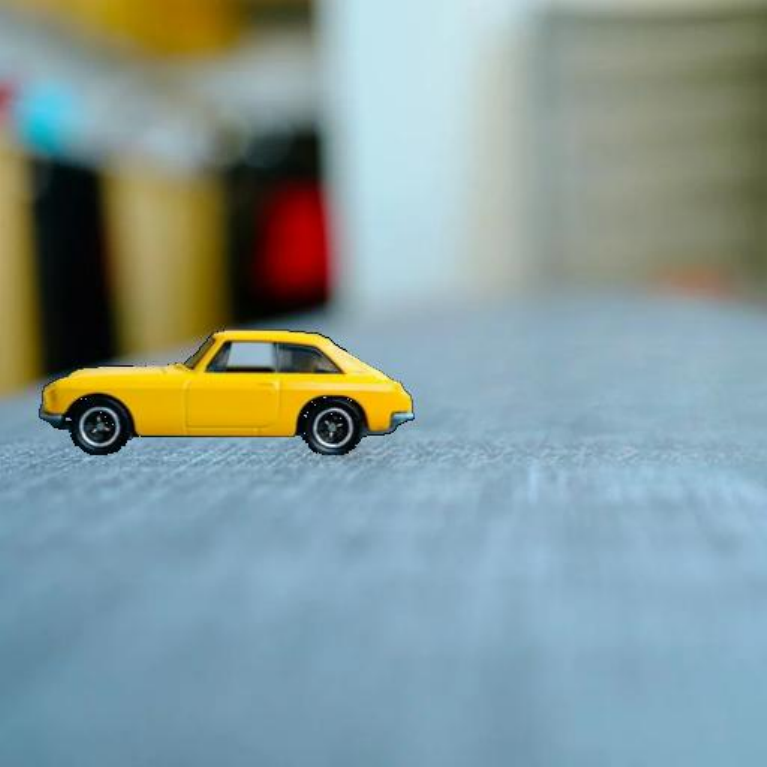} &
        \includegraphics[width=0.25\linewidth]{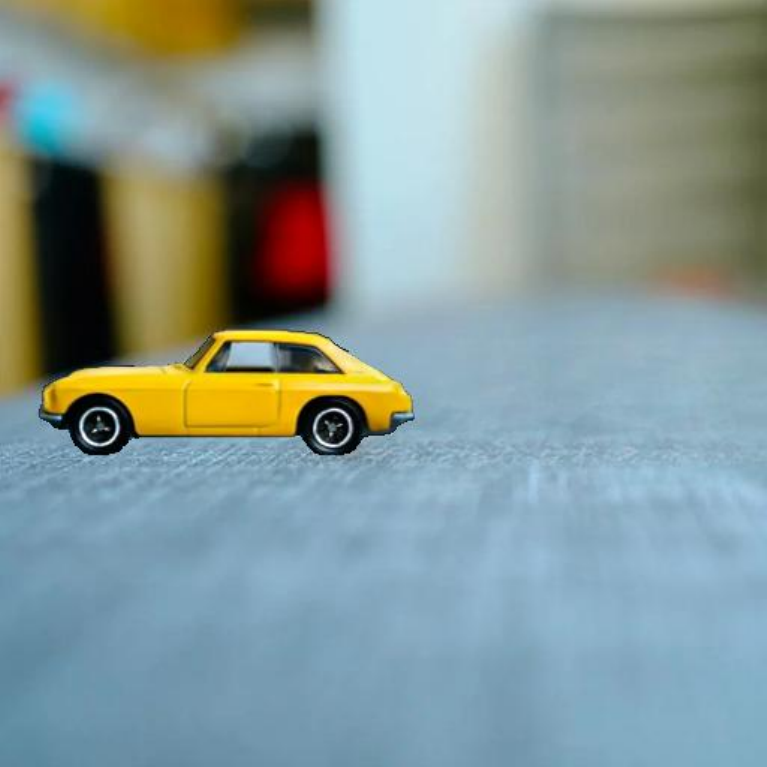} &
         \includegraphics[width=0.25\linewidth]{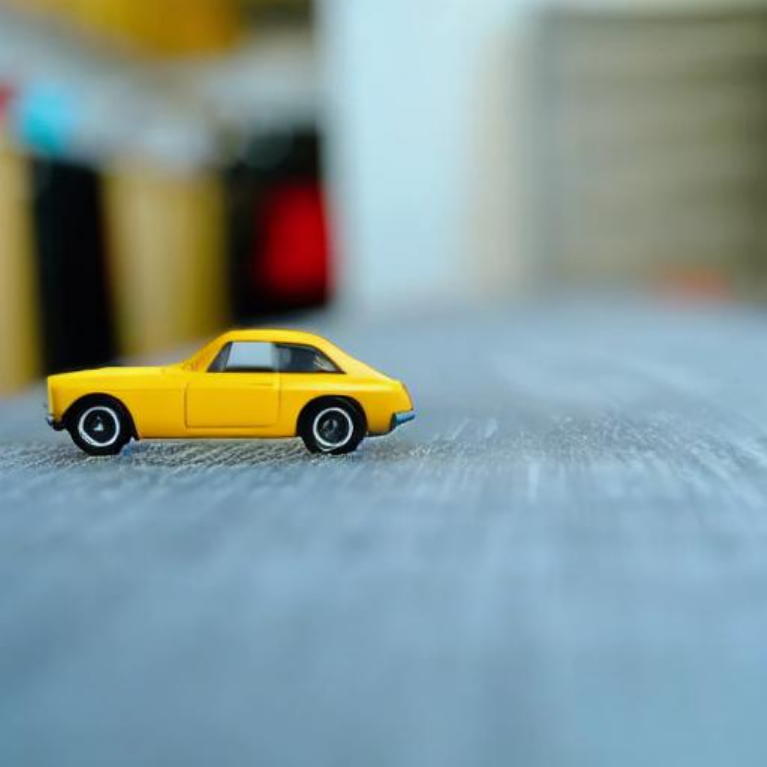} \\
         
         \includegraphics[width=0.25\linewidth]{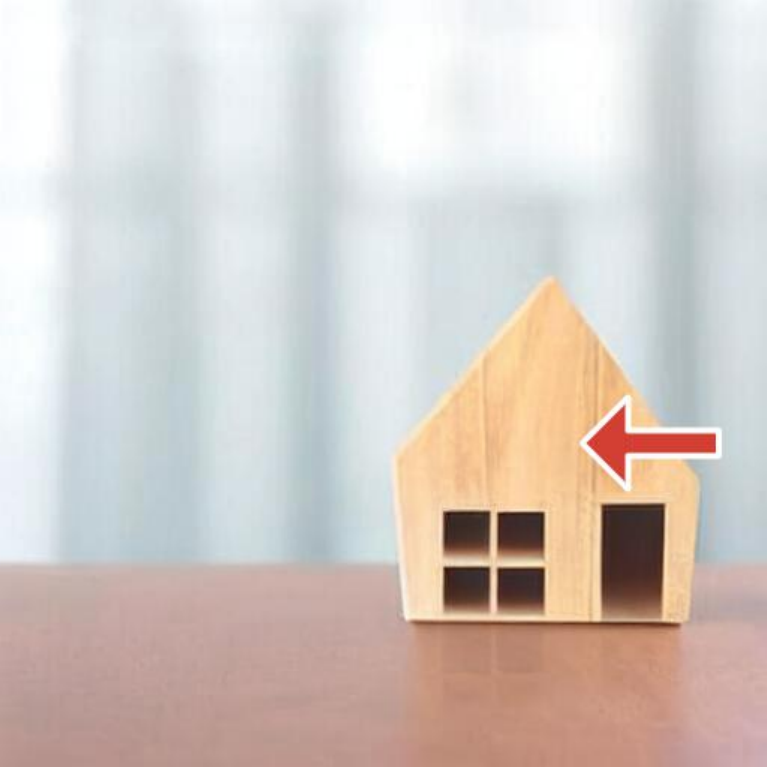} &
         \includegraphics[width=0.25\linewidth]{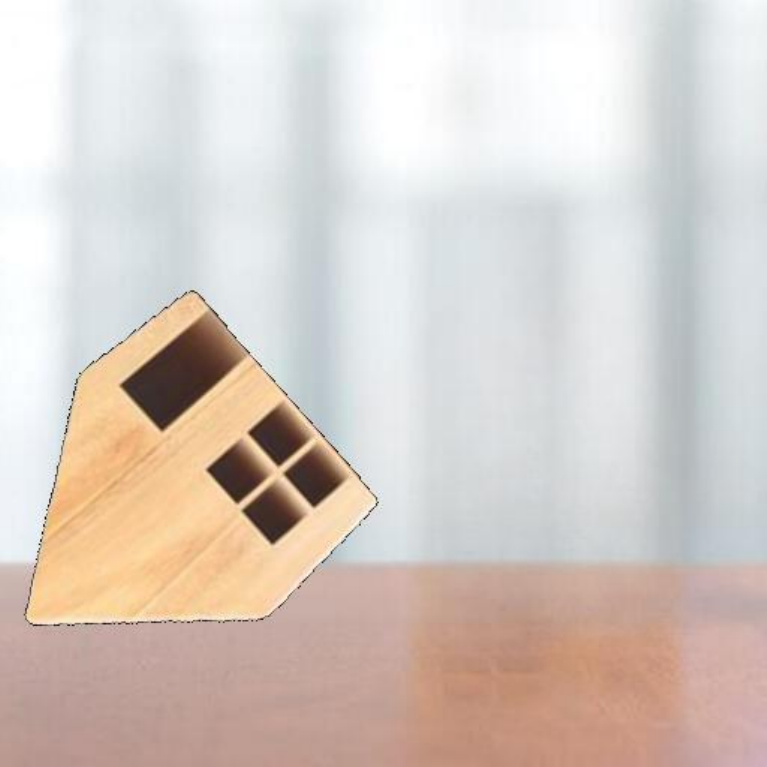} &
        \includegraphics[width=0.25\linewidth]{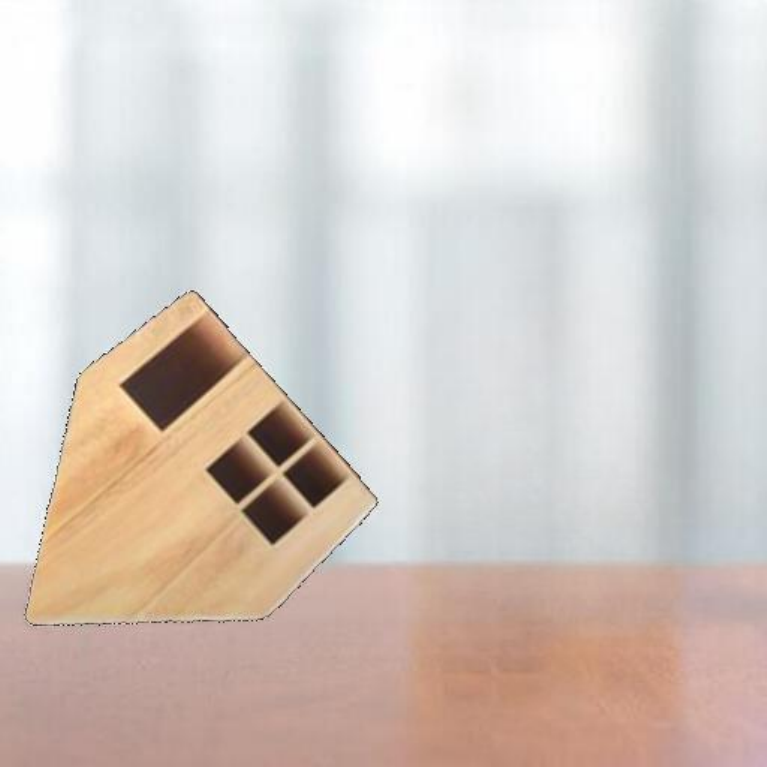} &
         \includegraphics[width=0.25\linewidth]{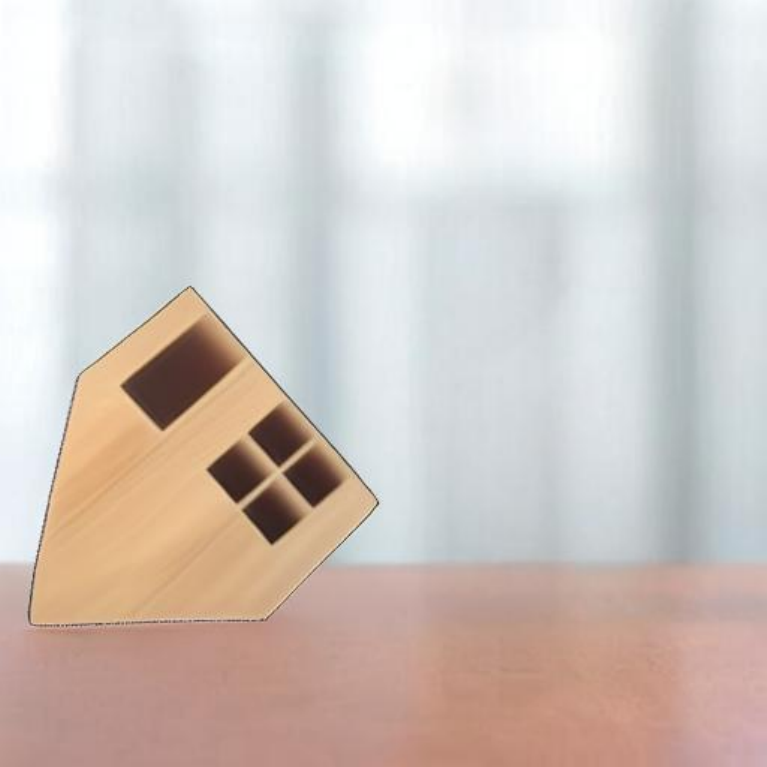} \\
         Input & Composited & Relight & Render
    \end{tabular}
    }
\captionof{figure}{\textbf{Rendered Video comparison.} The left shows the input frame, and the rest 3 are future frame generations. The composited frame hasn't been aware of the light change. The relighted output synthesized no shadows beneath. The rendered output from diffusion model is most photorealistic.}
\label{fig:supp_render}
\end{table}

\subsection{Image Space Dynamics Simulation}
In image space dynamics simulation, we set $\mathit{\Delta}_t$ as 1 second, gravity as $g=980cm/s^2$. Considering most foreground objects are captured at a similar distance, we set $1$pixel as $1$cm to map from image space to world space without reasoning metric scale. We visualize the physical simulation procedure of billiard balls and blocks with two different primitives (circle and polygon) in \cref{fig:supp_phy}.

\begin{algorithm}[tb]
 \caption{Video Generative Refinement}
  \textbf{Input}: Init $\tilde{\mathbf{z}_0}$, foreground mask $m$, video diffusion model $p_{\theta}$, noise strength $s$, fusion timestamp $\delta$, total denoising timesteps $T$ \\
  \textbf{Output}: refined latent ${\textbf{z}_0}$
    \begin{algorithmic}[1] %
        \State 
            $T \leftarrow \nint{T * s}$
            \Comment{Denoised time steps}
        \State 
            $\mathbf{z}_T, \tilde{\mathbf{z}_T} \sim q\left(
            \mathbf{z}_T \mid \tilde{\mathbf{z}_0}\right) $ \Comment{Add noise to the guidance latent code}
         \For {$t = T, T-1,...,1$}

            \If{$t \le (T - \delta)$} 
                \State 
                    $\tilde{\mathbf{z}_{t-1}} \sim q\left(\mathbf{z}_{t-1} \mid \tilde{\mathbf{z}_0}\right) $ \Comment{Add noise to guidance code at ${t-1}$}
                \State
                    $\mathbf{z}_{t-1} \sim p_\theta\left(\mathbf{z}_{t-1} \mid \mathbf{z}_t\right)$ \Comment{Denoised output from network at $t-1$}
                \State
                    $w = (T-t)/T$ \Comment{fusion weight}
                \State
                    $\mathbf{z}_{t-1} \leftarrow (1-m)\mathbf{z}_{t-1} + m\left[w\mathbf{z}_{t-1} + (1-w)\tilde{\mathbf{z}_{t-1}}\right]$  \Comment{update latent code}
            \Else
                \State
                    $\mathbf{z}_{t-1} \sim p_\theta\left(\mathbf{z}_{t-1} \mid \mathbf{z}_t\right)$ \Comment{Denoised output from network}
            \EndIf 
        \EndFor
    \State \textbf{return} ${\mathbf{z}_0}$
    \end{algorithmic}
\label{alg:denoise}
\end{algorithm}

\subsection{Generative Refinement Algorithm}
\label{sec:supp_algorithm}
We present the detailed algorithm of the proposed generative refinement with latent diffusion models in Sec. 3.4. In video diffusion model, a input video $\mathbf{V} \in \bbR^{T \times H \times W \times 3}$ is encoded to a latent vector via a encoder $\mathcal{E}$ by $\mathbf{z}_0 = \mathcal{E}(\mathbf{V}) \in \bbR^{T \times h \times w \times 3}$, where $c$ is the dimension of the latent space. The forward diffusion process~\cite{ho2020denoising} is to iteratively add
Gaussian noise to the signal, given by \cref{eq:fw}.
\begin{equation}
\label{eq:fw}
\begin{split}
      q\left(\mathbf{z}_t \mid \mathbf{z}_{t-1}\right) & =\mathcal{N}\left(\mathbf{z}_t ; \sqrt{1-\beta_{t-1}} \mathbf{z}_{t-1}, \beta_t \mathbf{I}\right), \quad t=1, \ldots, T \\
    q\left(\mathbf{z}_t \mid \mathbf{z}_0\right) &=\mathcal{N}\left(\mathbf{z}_t ;  \sqrt{\Bar{\alpha}_t} \mathbf{z}_{0}, (1 - \Bar{\alpha}_t) \mathbf{I}\right), \quad t=1, \ldots, T
\end{split}
\end{equation}
where ${\alpha}_t = 1 - \beta_t$, $\Bar{\alpha}_t = \prod_{i=1}^t \alpha_i$, 
$q\left(\mathbf{z}_t \mid \mathbf{z}_{t-1}\right)$ is the one-step forward diffusion process, and $ q\left(\mathbf{z}_t \mid \mathbf{z}_0\right)$ is $t$-step forward diffusion process. $T$ is a large integer to make the forward process completely destroys the
initial signal $\mathbf{z}_0$ resulting in $\mathbf{z}_T \sim \mathcal{N}(0, \mathbf{I})$. 
The diffusion model learns to recover $\mathbf{z}_0$ from standard Gaussian noise $\mathbf{z}_T$ by backward diffusion
process in \cref{eq:bw}.
\begin{equation}
\label{eq:bw}
    p_\theta\left(\mathbf{z}_{t-1} \mid \mathbf{z}_t\right)=\mathcal{N}\left(\mathbf{z}_{t-1} ; \mathbf{\mu}_\theta\left(\mathbf{z}_t, t\right), \mathbf{\Sigma}_\theta\left(\mathbf{z}_t, t\right)\right), \quad t=T, \ldots, 1
\end{equation}
where $\theta$ is the learned parameters of the model. The output is passed to a decoder $\mathcal{D}$ to generate the output video $\mathbf{V} = \mathcal{D}(\mathbf{z}_0)$. To this end, given the relit video $\tilde{\mathbf{V}}$, we encode to get $\tilde{\mathbf{z}_0}$, our goal is to obtain the denoised $\mathbf{z}_0$ from the pretrained latent-diffusion-based video $p_\theta$. The algorithm is shown in \cref{alg:denoise}.

\begin{table}[t]
\caption{\textbf{Runtime analysis.} We summarize the runtime of each module for each run.}
\label{table:runtime}
    \centering
    \small
    {
    \setlength{\tabcolsep}{0.4em} %
    \begin{tabular}{c|c|c|c|c|c}
        \toprule
       \multicolumn{3}{c}{\bf Perception} & \multicolumn{1}{c}{\bf Simulation} & \multicolumn{1}{c}{ \bf Render} & \multicolumn{1}{c}{\bf Refinement} \\
       \multicolumn{1}{c}{\bf Segmentation} & \multicolumn{1}{c}{\bf GPT4-V} & \multicolumn{1}{c}{\bf Inpainting} & \multicolumn{1}{c}{} & \multicolumn{1}{c}{} & \multicolumn{1}{c}{} \\
       \midrule
       50s & 10s & 20s & 5s & 60s & 35s \\
       \bottomrule
    \end{tabular}}
\end{table}

In \cref{fig:supp_render}, we compare the composed video \(\hat{\mathbf{V}}\), the relit video \(\tilde{\mathbf{V}}\), and our final output \(\mathbf{V}\) by 2 additional examples.

\subsection{Runtime Analysis}
We summarize the runtime of each module in \cref{table:runtime} for a single run. The perception module takes 1 minute, the simulation (120 steps) takes 5 seconds, the render module takes 1 minute, the generative refinement takes 35 seconds. In total it takes around 3 minute for a single generation, which is much faster than other controllable generation, \eg Motion Guidance~\cite{geng2024motion} takes 70 minutes for a single run.

\begin{figure}[t]
    \centering
     \includegraphics[width=1.0\linewidth]{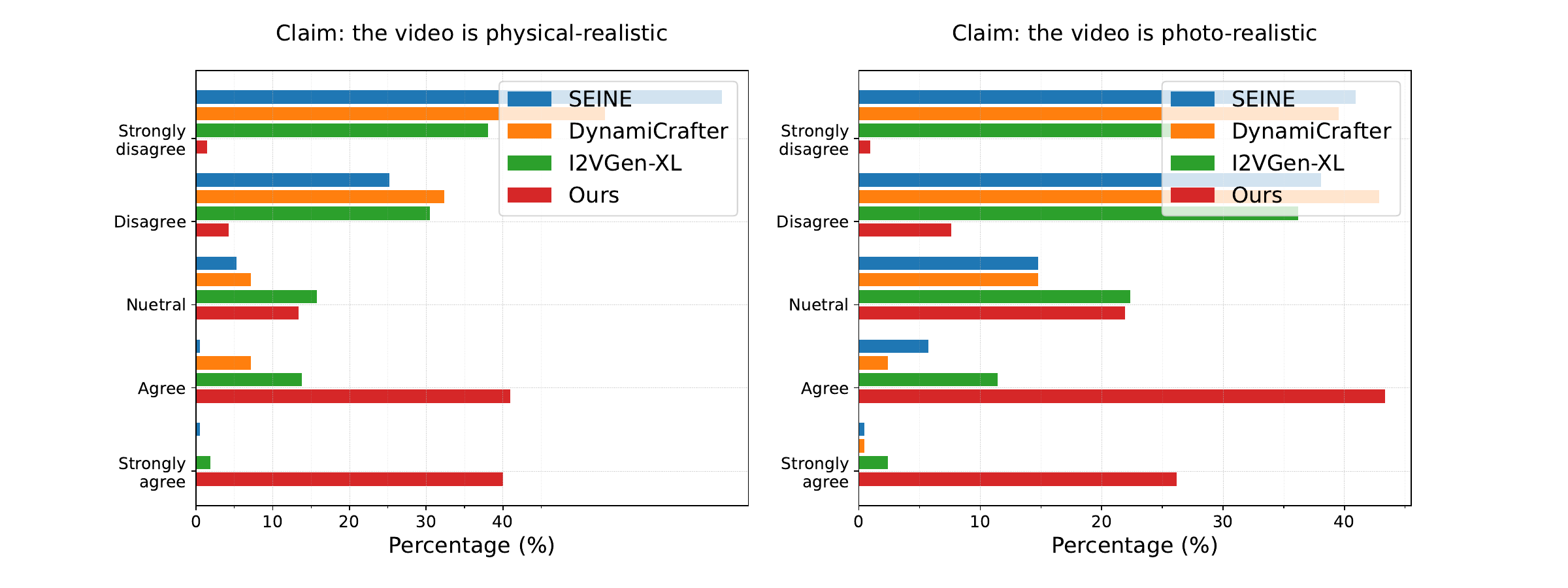} 
\caption{\textbf{Human evaluation score distribution.} The distribution of scores shows our method largely outperforms other I2V generative models in both physical-realism and photo-realism. Our average rate is close to agree for both two claims.}
\label{fig:user_study}
\end{figure}
\begin{table}[t]
    \centering
    \footnotesize
    \resizebox{\linewidth}{!}{
    \setlength{\tabcolsep}{0.2em} 
        \begin{tabular}{c|ccccc}
        \multicolumn{6}{c}{{ (left$\rightarrow$right: time steps)}} \\
        \visvideo{real}{2}{12}{13}{14}{15}{16}{17} \\
        \visvideo{real}{9}{5}{6}{10}{13}{15}{17} \\
        \visvideo{real}{12}{4}{5}{7}{9}{11}{13} \\
        \visvideo{real}{23}{4}{5}{6}{8}{10}{14} \\
        \visvideo{real}{47}{4}{5}{6}{7}{8}{10} \\
    \end{tabular}  
    }
    \captionof{figure}{\textbf{Random sampled real-captured videos.} We captured real-world videos for the same given scene 50 times to evaluate the generation fidelity. We vary initial force on the hand that applied to the blue piggy bank in each run, and record the videos. We \textbf{random select} 5 recored videos for visualization.}
    \label{fig:supp_real}
    \end{table}

\section{Experiment details}
\label{sec:supp_analysis}

\subsection{Human evaluation}
\cref{fig:user_study} shows the score distribution of the conducted human evaluation. Our method has much higher percentage in agree and strongly-agree to both claims, outperform compared I2V methods by a large margin in physical-realism and photo-realism. Our average rate falls within Agree level.

\begin{figure}[t]
    \centering
     \includegraphics[width=0.7\linewidth]{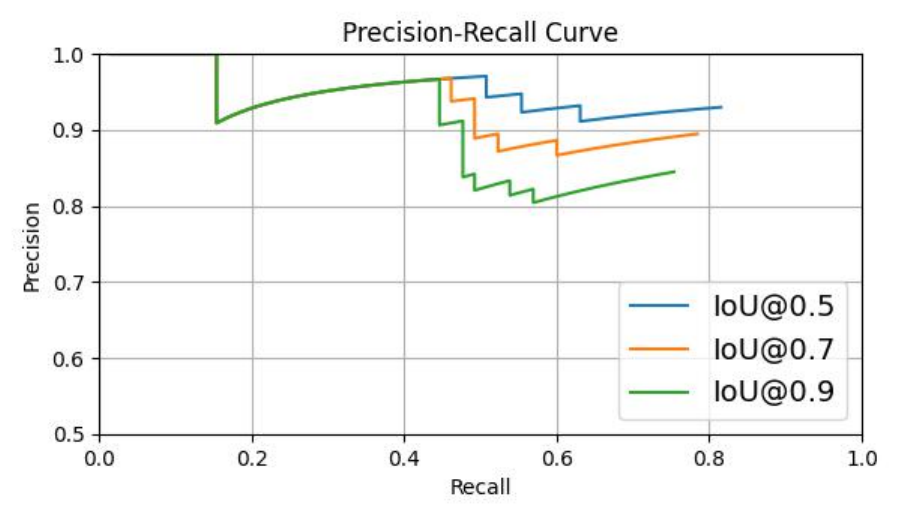} 
\caption{\textbf{Precision-recall curve of open-world movable objects segmentation.} Our proposed pipeline achieves \textbf{0.93} precision, \textbf{0.82} recall at 0.5 IoU.}
\label{fig:pr_curve}
\end{figure}
\begin{table}[!h]
    \centering
    \footnotesize
    \resizebox{0.9\linewidth}{!}{
    \setlength{\tabcolsep}{0.2em} 
    \begin{tabular}{cc}
       \includegraphics[width=0.4\linewidth]{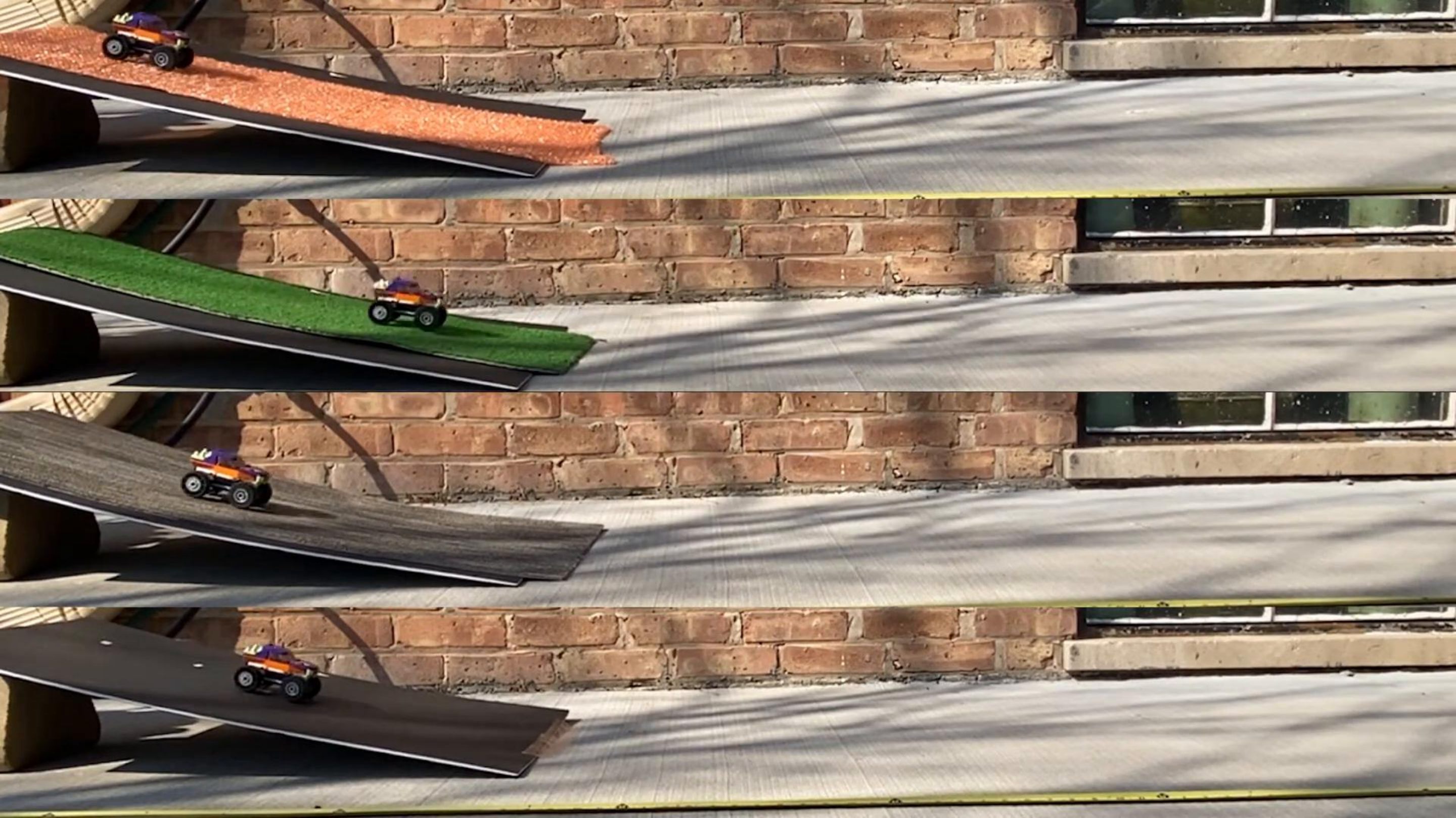} &
       \includegraphics[width=0.4\linewidth]{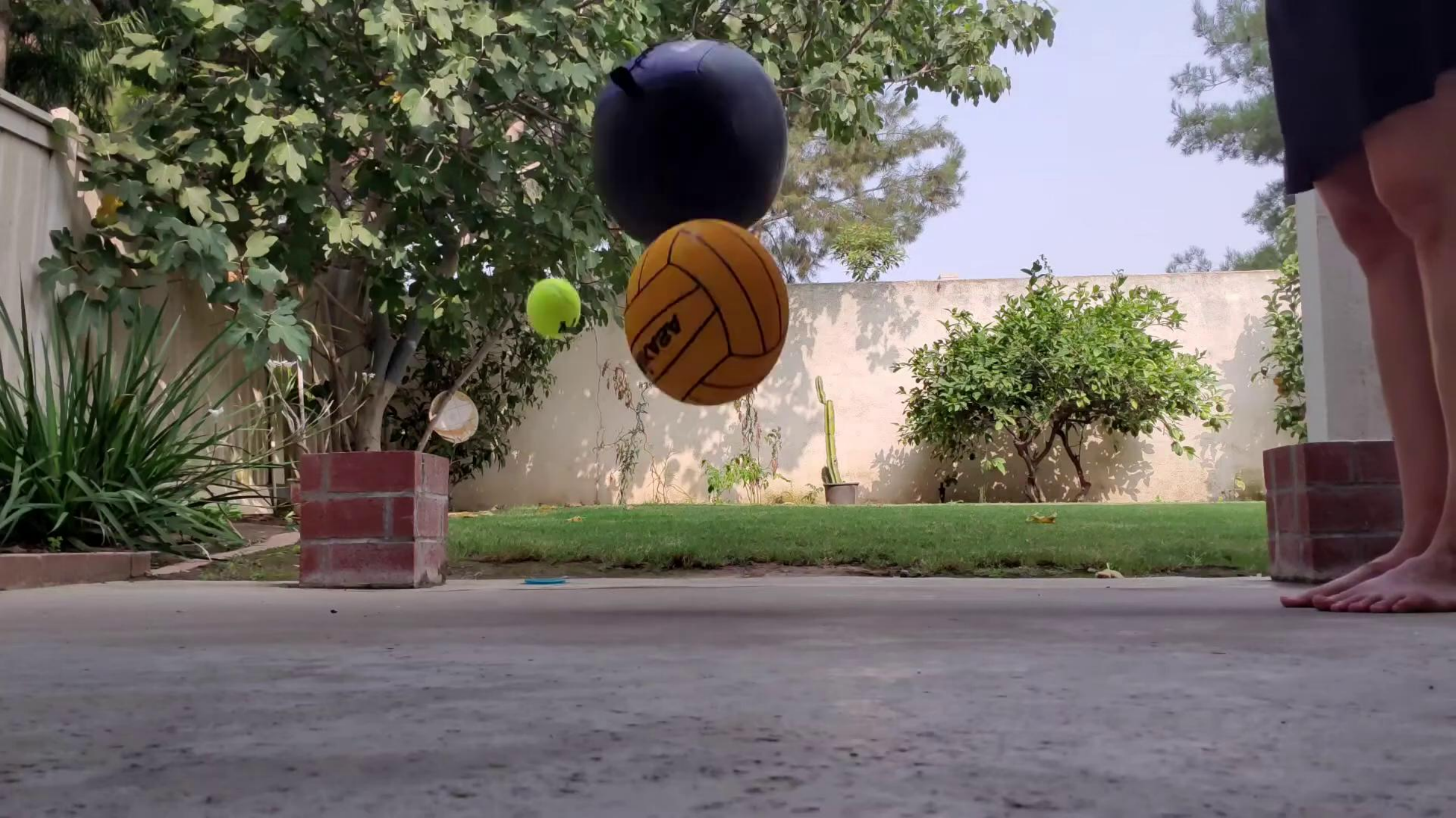} \\
    \end{tabular}  
    }
\captionof{figure}{\textbf{GPT-4V physical property estimation evaluation testing scenes.} Given directly measure friction and elasticity is hard, we use reference videos of toy cars sliding on various surfaces (left) and balls of different
materials bouncing (right) to rank materials by friction and elasticity.}
\label{fig:phy_eval}
\end{table}
\begin{table}
    \centering
    \footnotesize
    {
    \setlength{\tabcolsep}{0.2em} %
    \begin{tabular}{c|cc}
        {Input}& Pred Seg & GT Seg \\
        \includegraphics[width=0.3\linewidth]{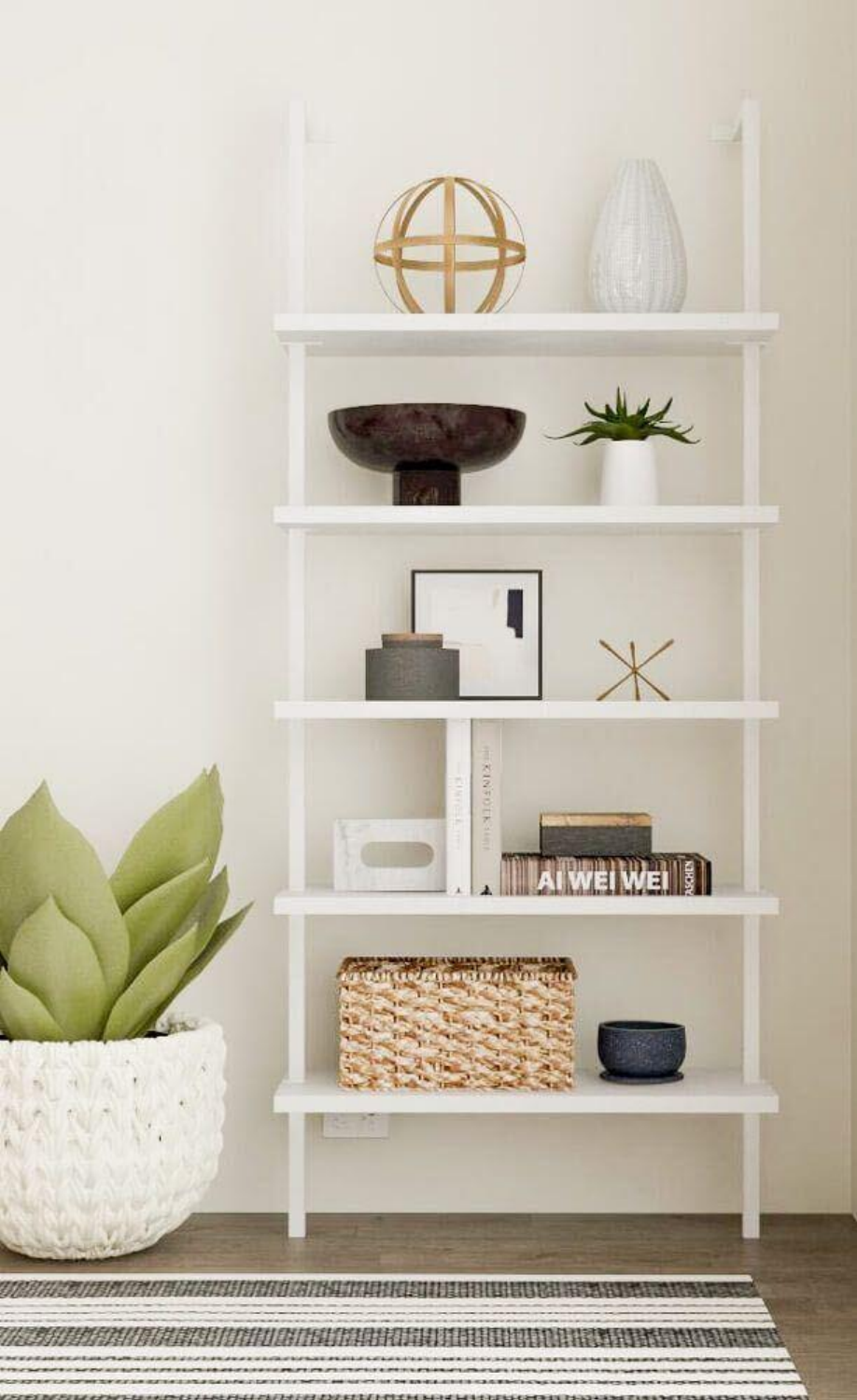} & 
        \includegraphics[width=0.3\linewidth]{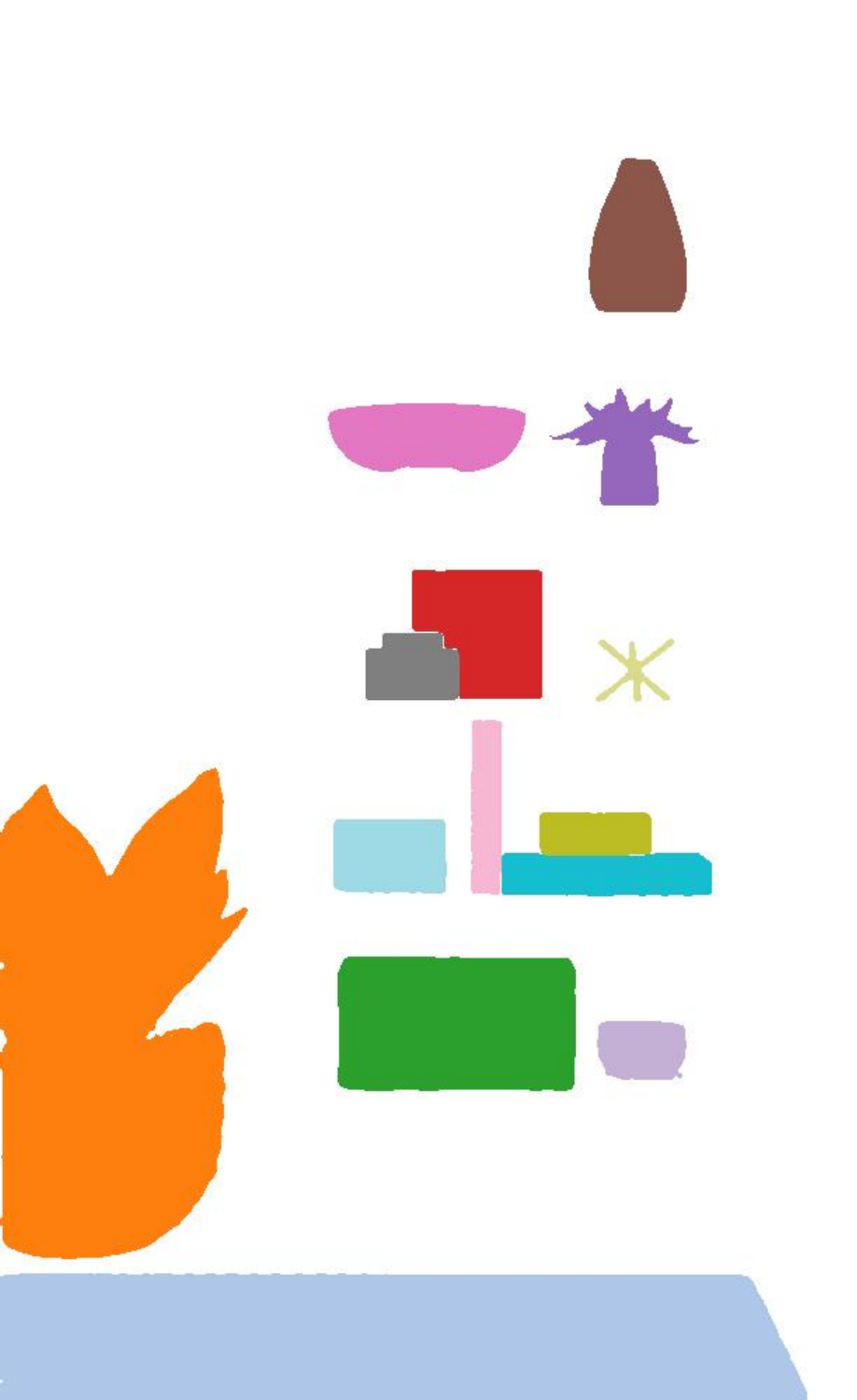} & 
        \includegraphics[width=0.3\linewidth]{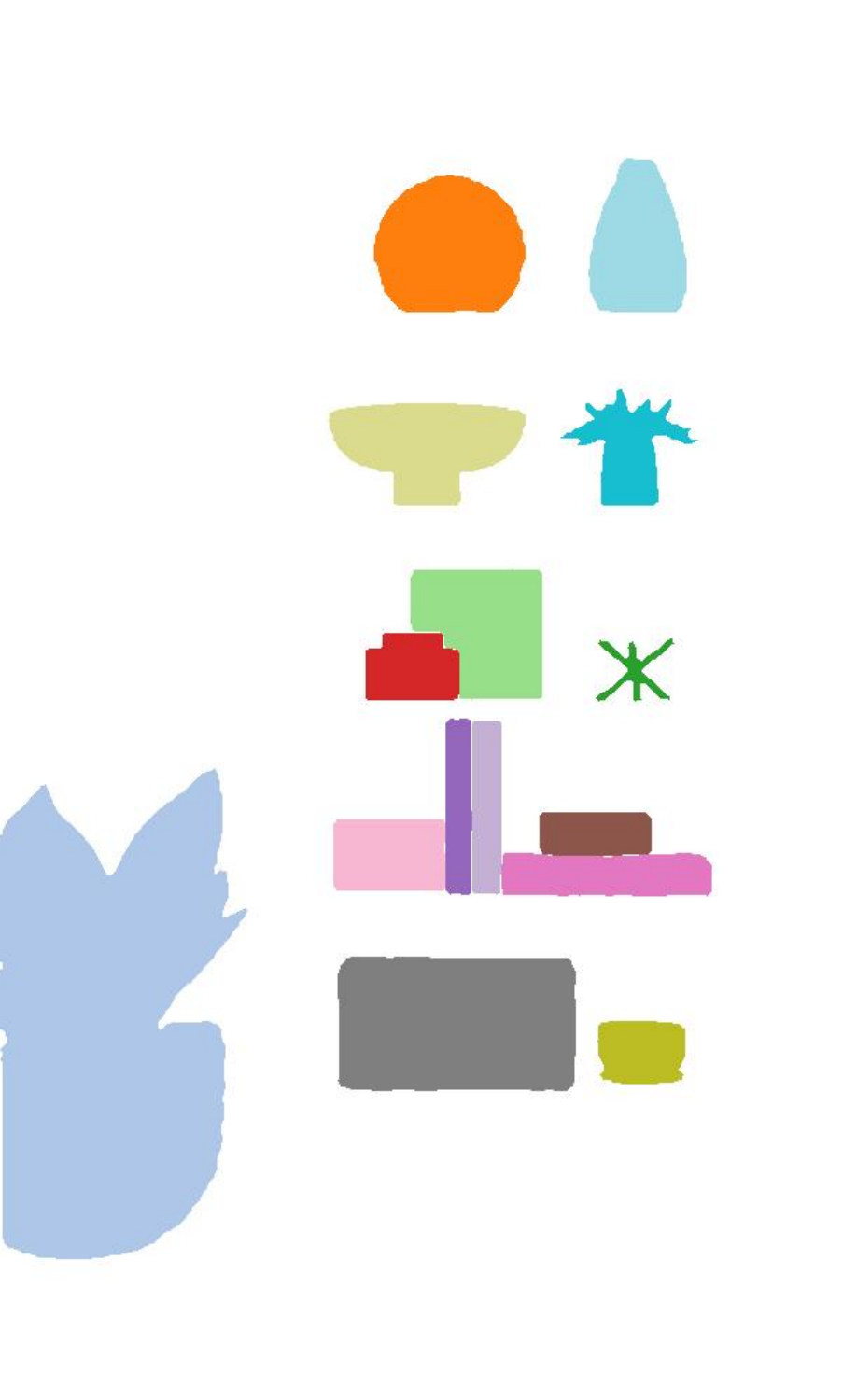} \\

        & \textbf{house}  & \\
        \includegraphics[width=0.3\linewidth]{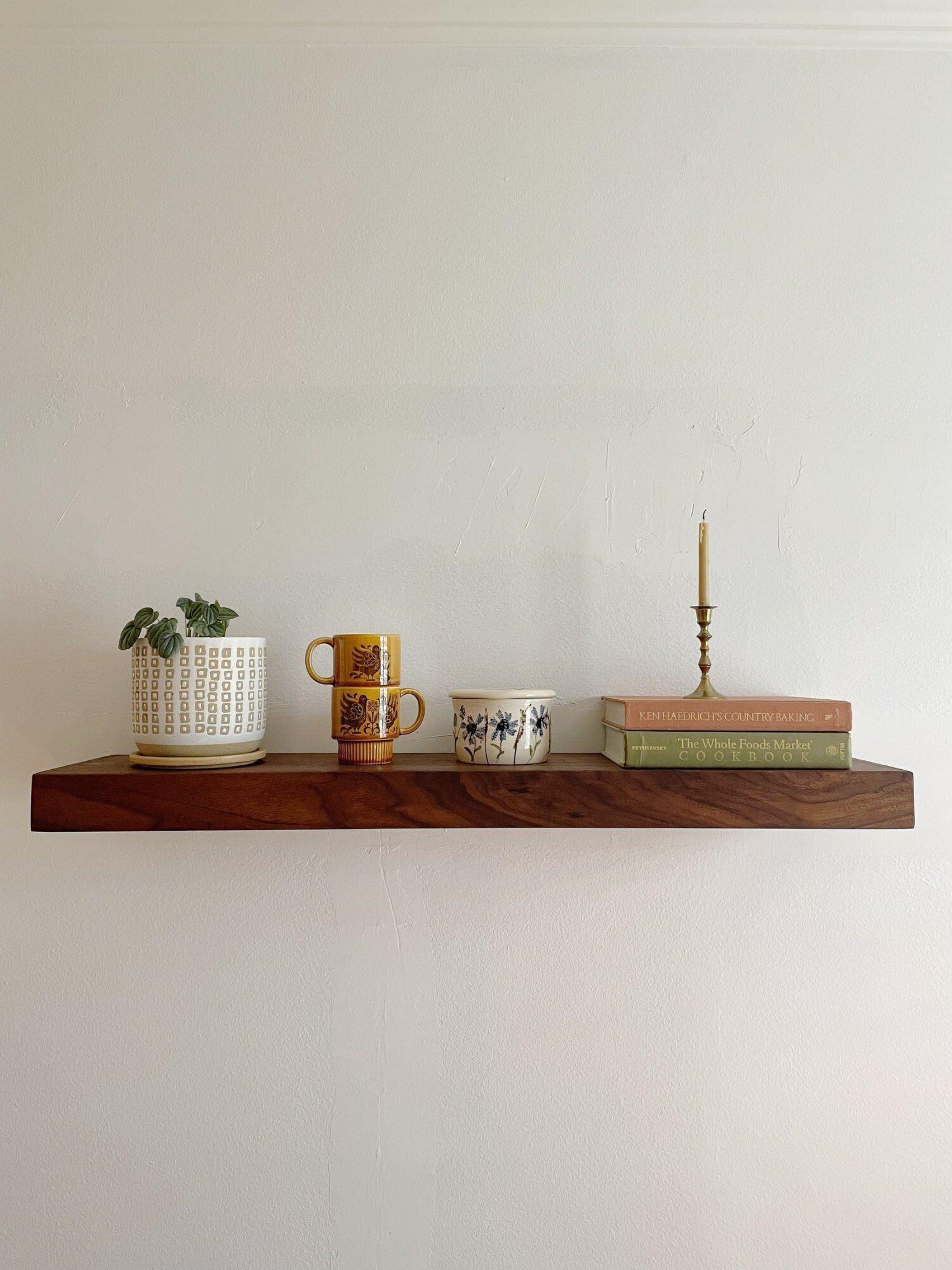} & 
        \includegraphics[width=0.3\linewidth]{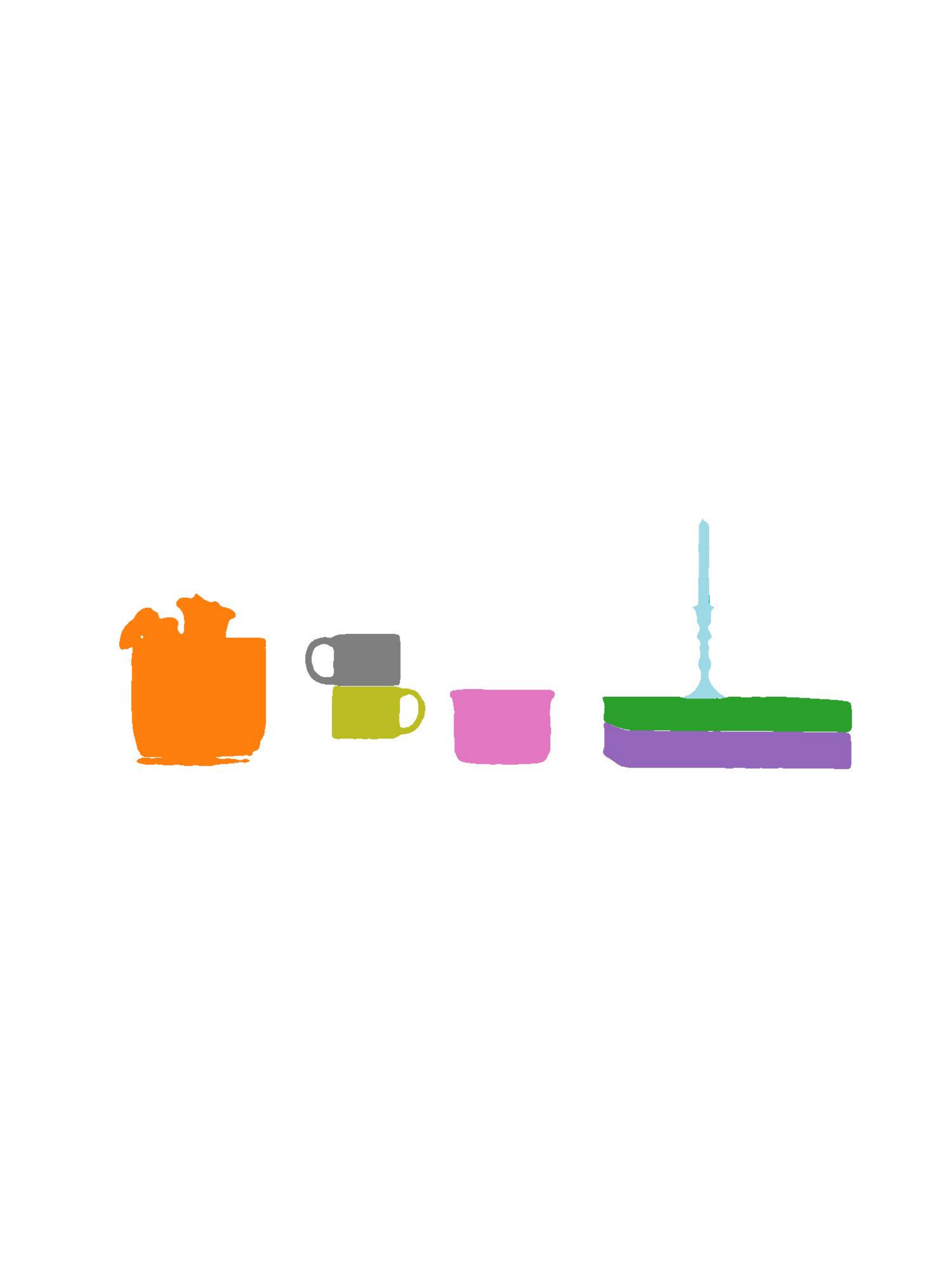} & 
        \includegraphics[width=0.3\linewidth]{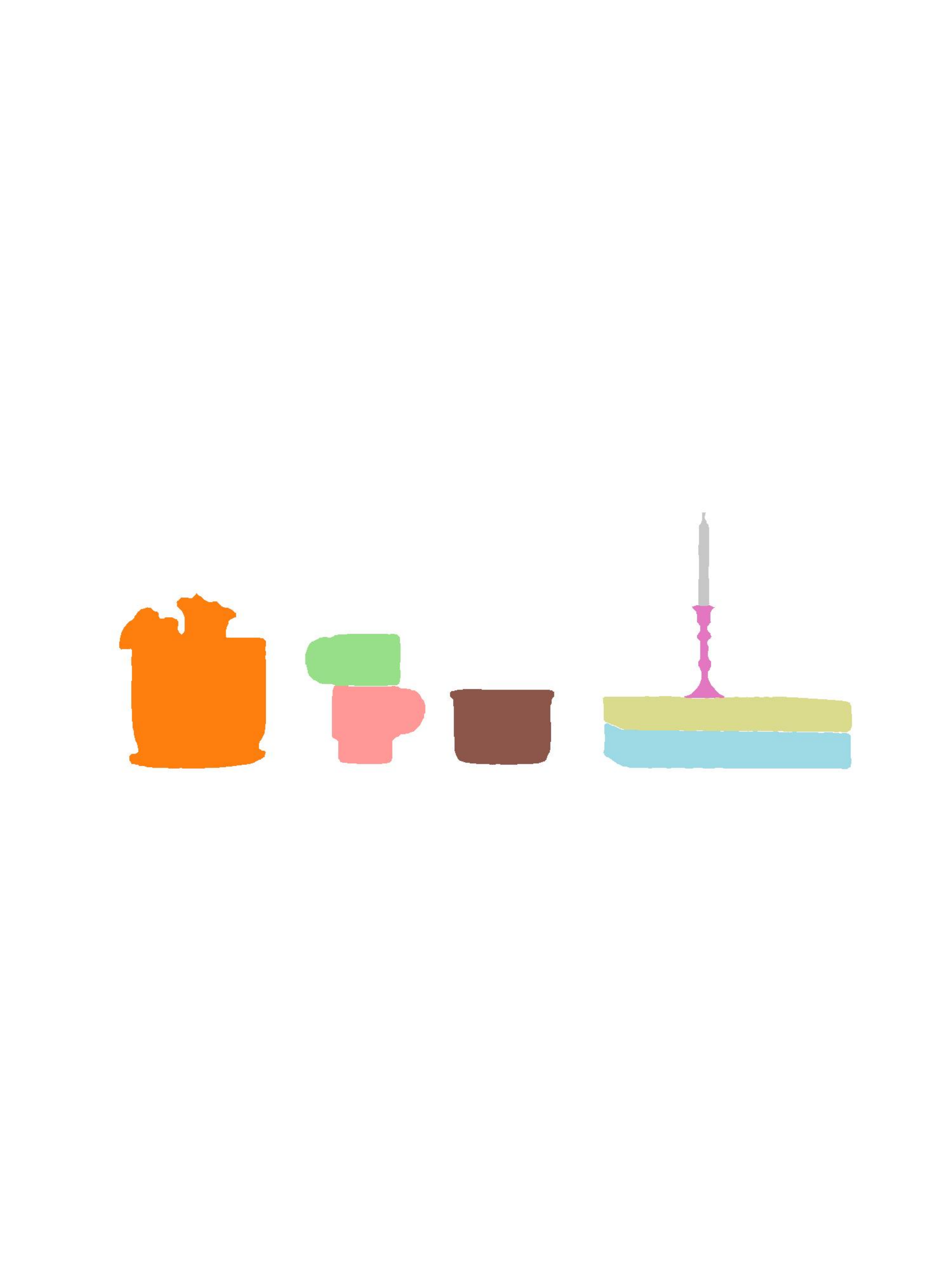} \\

        & \textbf{table} & \\
        \includegraphics[width=0.3\linewidth]{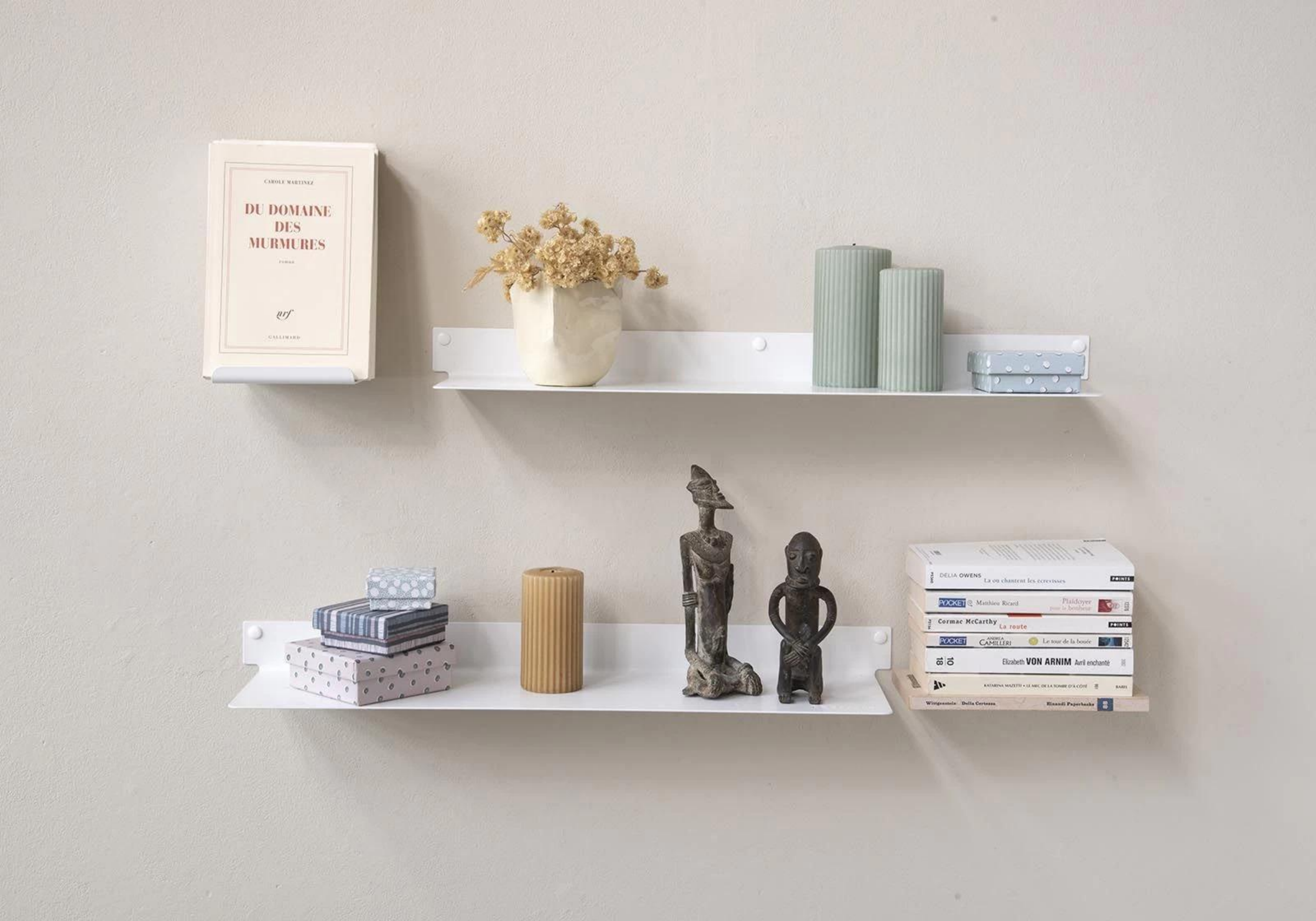} & 
        \includegraphics[width=0.3\linewidth]{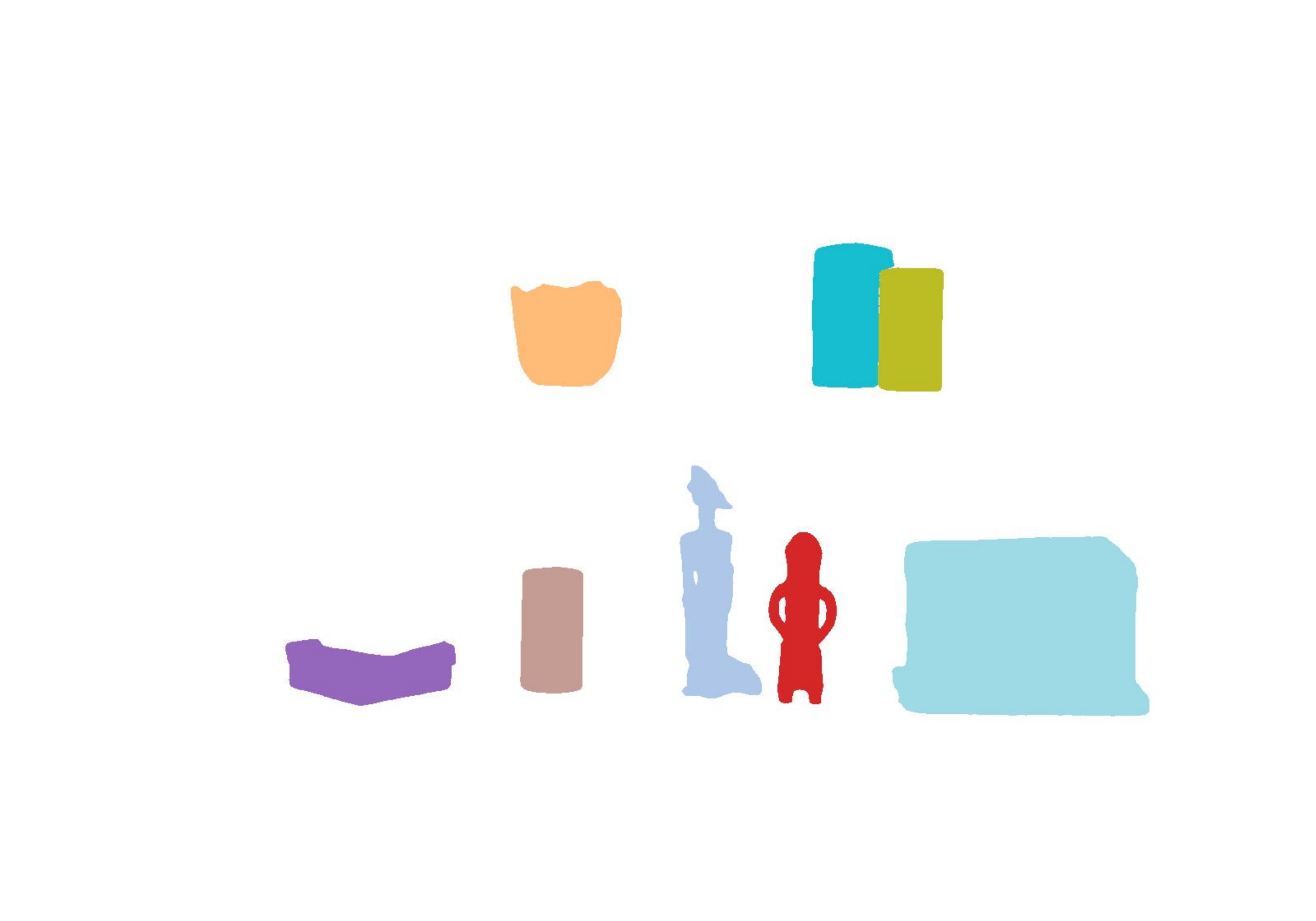} & 
        \includegraphics[width=0.3\linewidth]{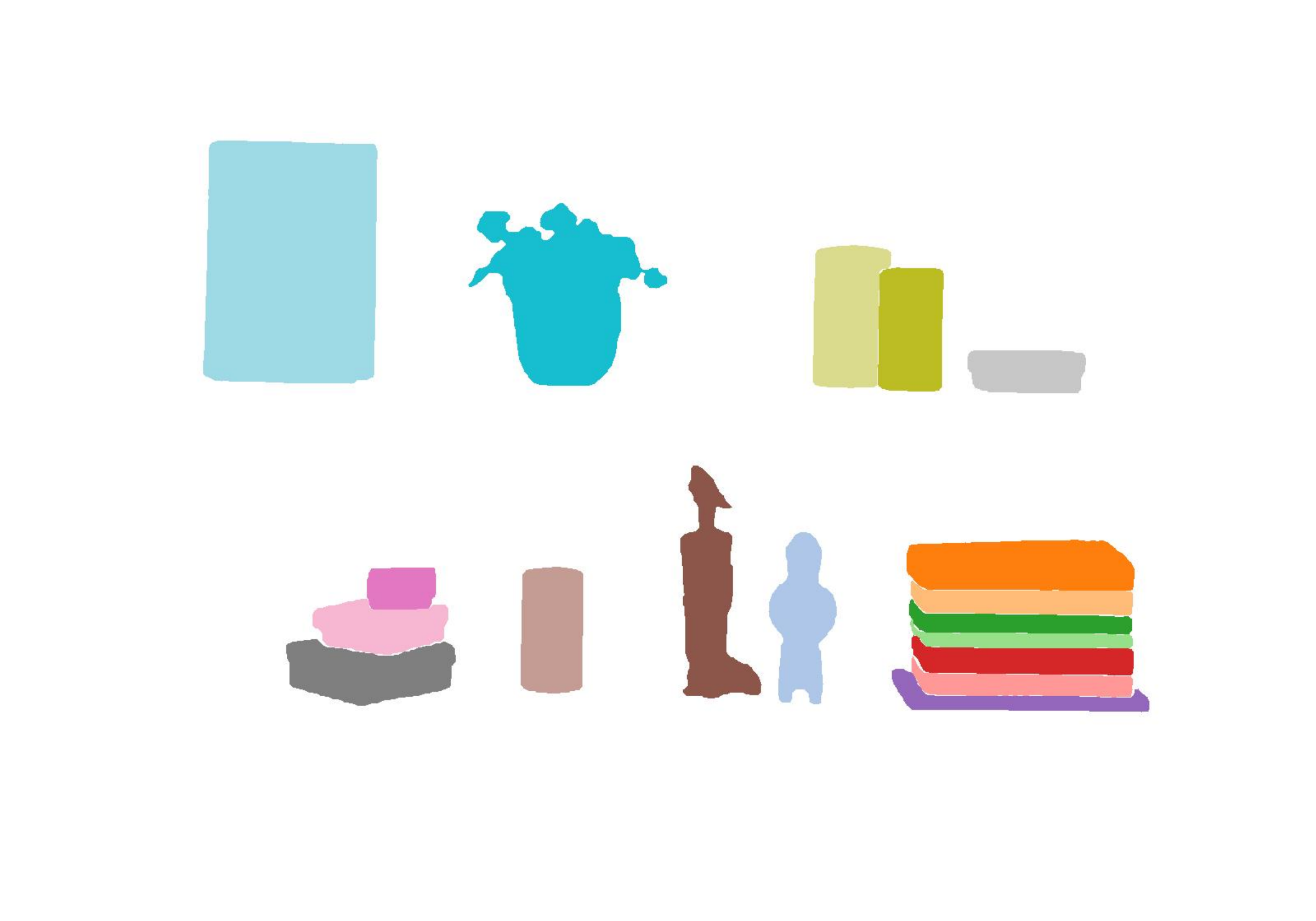} \\
        & \textbf{shelf} & \\
        \includegraphics[width=0.3\linewidth]{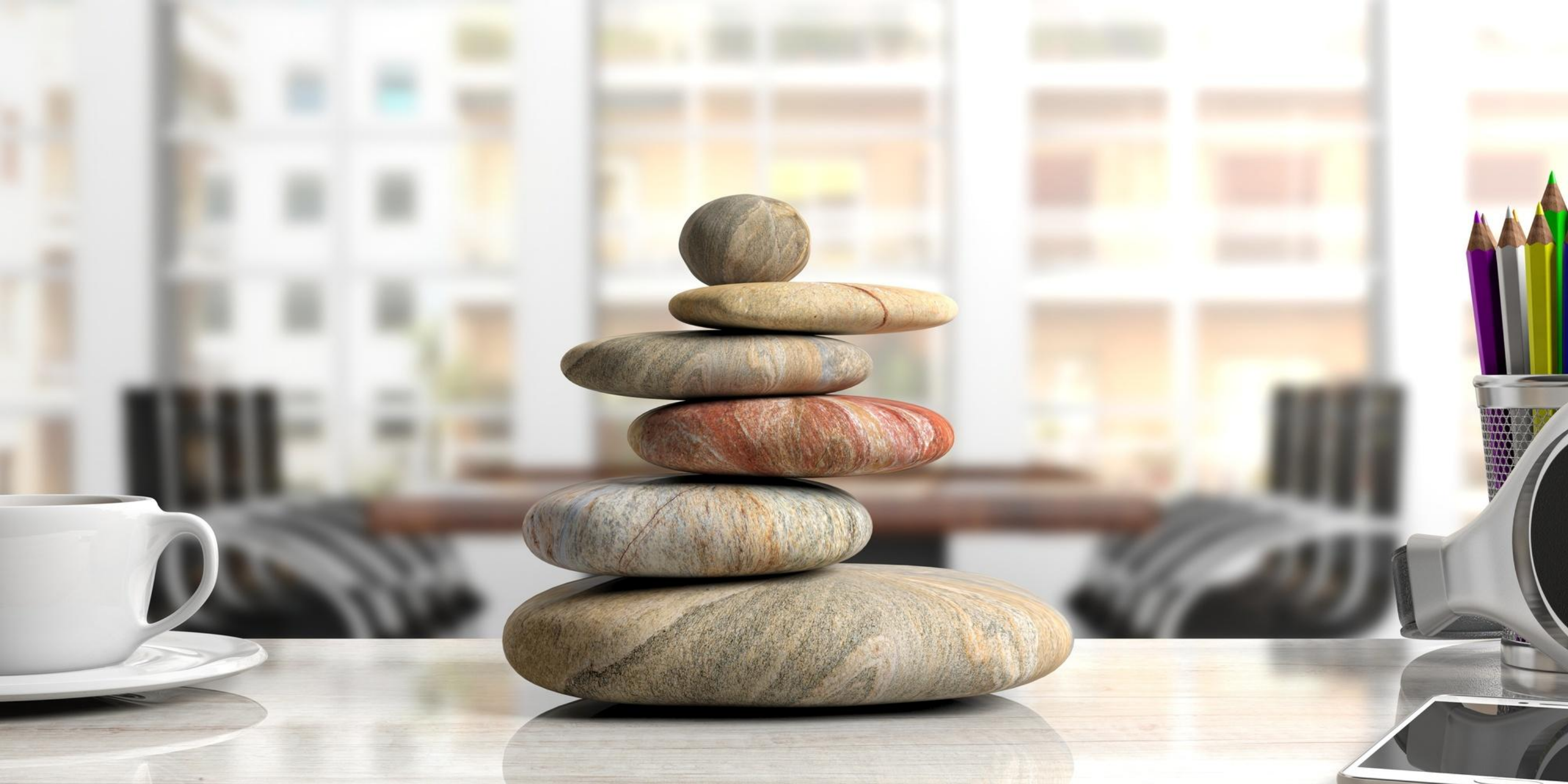} & 
        \includegraphics[width=0.3\linewidth]{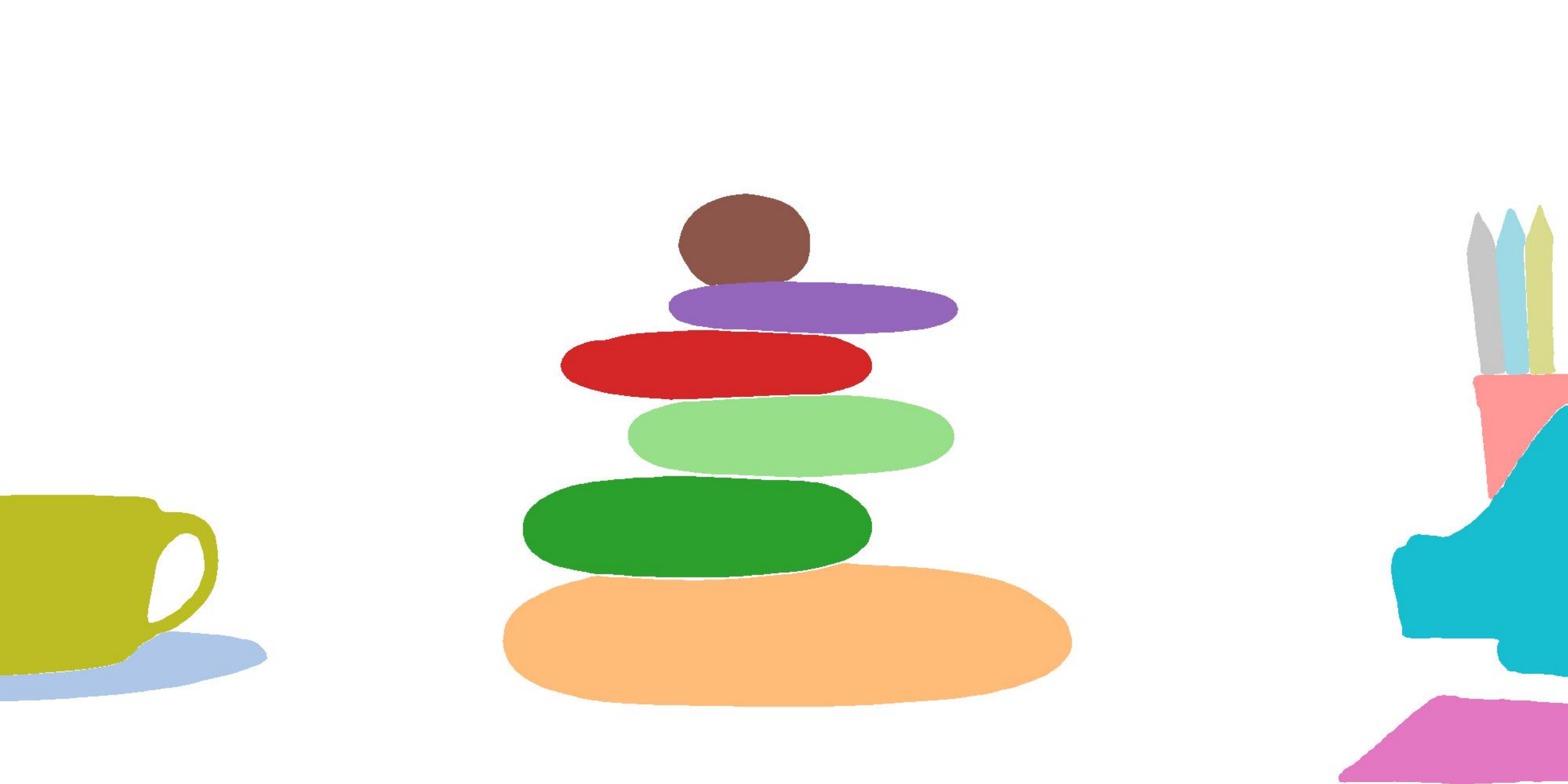} & 
        \includegraphics[width=0.3\linewidth]{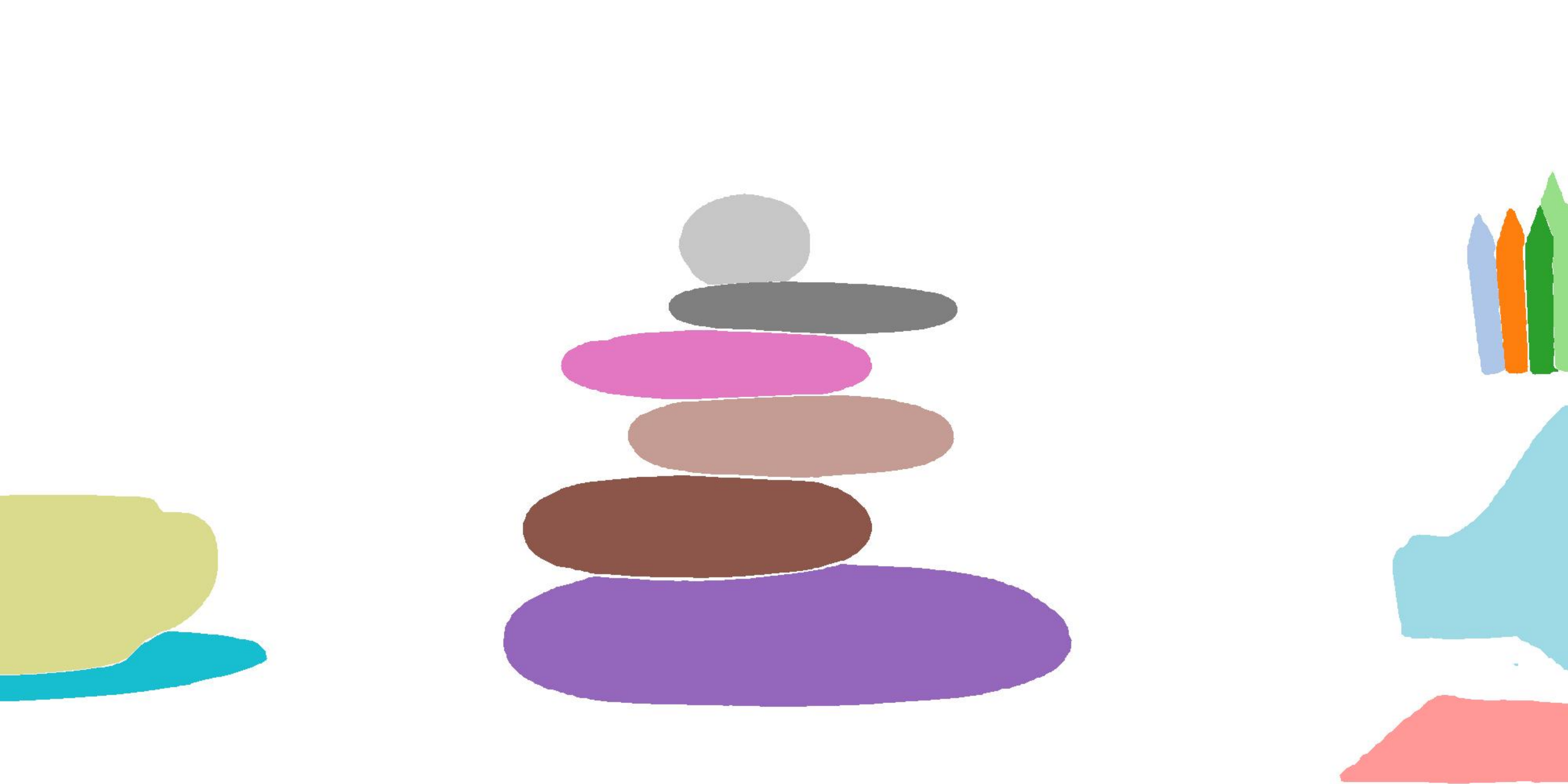} \\
       & \textbf{stone} & \\
    \end{tabular}
    }
\captionof{figure}{\textbf{Qualitative visualization of open-world movable objects segmentation results}. From left to right, we show the selected input image, inferred segmentation from our perception module and GT movable object segmentations.}
\label{fig:seg}
\end{table}
\begin{table}[t]
    \centering
    \footnotesize
    \resizebox{\linewidth}{!}{
    \setlength{\tabcolsep}{0.2em} 
    \begin{tabular}{c|cccccc}
        {Input} & 
        \multicolumn{5}{c}{{Generation (left$\rightarrow$right: time steps)}} \\
    
    \includegraphics[width=0.16\linewidth]{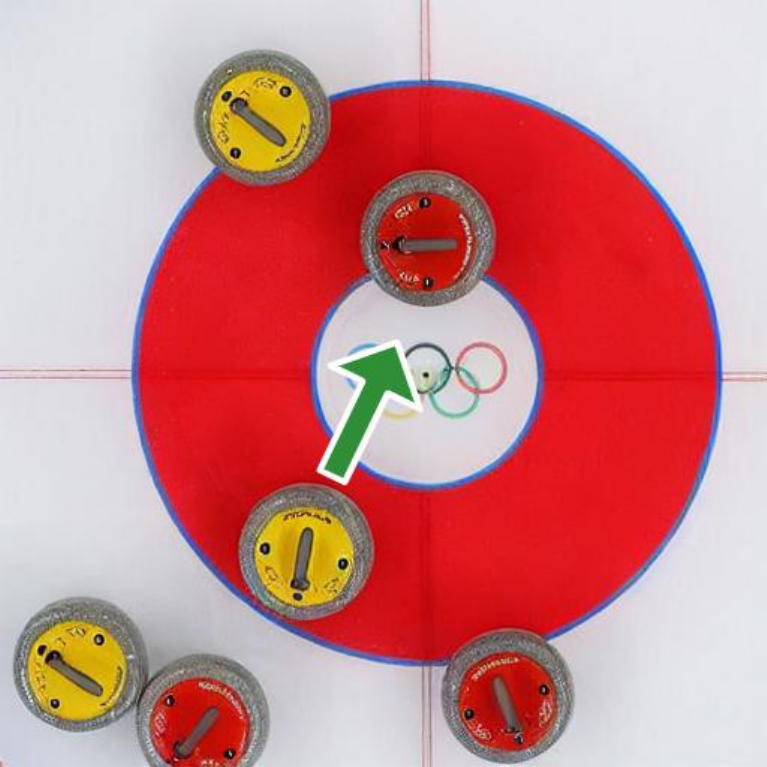} & \visvideo{dynamcrafter}{balls}{0}{8}{10}{13}{15}{} & \rotatebox{270}{\hspace{-40pt}DC~\cite{xing2023dynamicrafter}} \\
     & \visvideo{i2vgen-xl}{balls}{4}{5}{8}{11}{14}{} & \rotatebox{270}{\hspace{-40pt}VGen~\cite{2023I2VGen-XL}} \\

    & \visvideo{seine}{balls}{4}{6}{8}{10}{12}{}
    & \rotatebox{270}{\hspace{-40pt}SEINE~\cite{chen2023seine}} \\
    
     & \visvideo{ours}{balls}{2}{5}{8}{11}{14}{} & \rotatebox{270}{\hspace{-30pt}Ours} \\

     \includegraphics[width=0.16\linewidth]{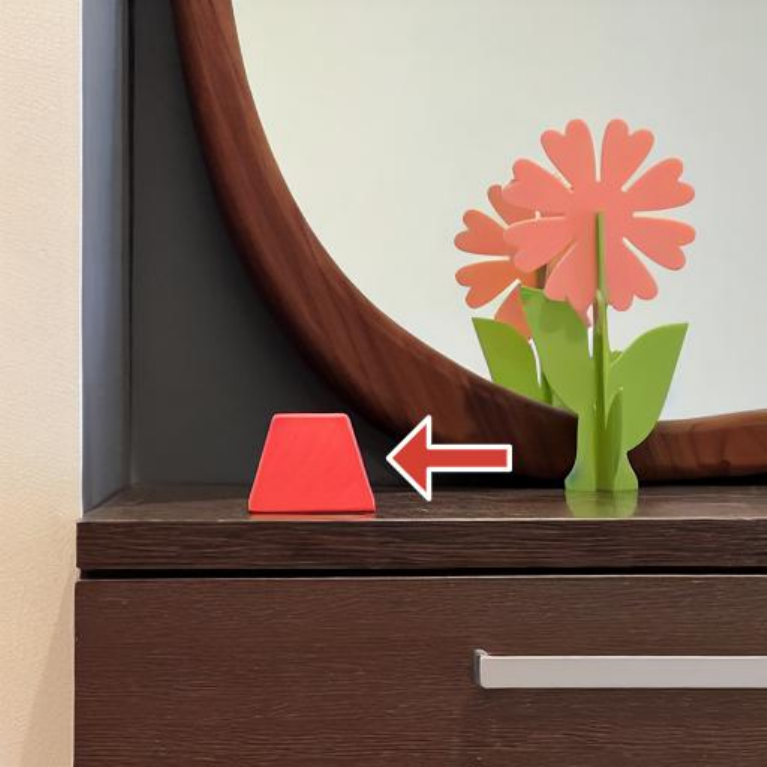} & \visvideo{dynamcrafter}{wall_toy}{2}{5}{8}{12}{15}{} & \rotatebox{270}{\hspace{-40pt}DC~\cite{xing2023dynamicrafter}} \\
     & \visvideo{i2vgen-xl}{wall_toy}{5}{8}{10}{12}{14}{} & \rotatebox{270}{\hspace{-40pt}VGen~\cite{2023I2VGen-XL}} \\

    & \visvideo{seine}{wall_toy}{6}{8}{10}{12}{15}{}
    & \rotatebox{270}{\hspace{-40pt}SEINE~\cite{chen2023seine}} \\
    
     & \visvideo{ours}{wall_toy}{2}{3}{4}{6}{8}{} & \rotatebox{270}{\hspace{-30pt}Ours} \\

    \end{tabular} 
    }
\captionof{figure}{\textbf{Additional qualitative comparison} against I2V generative models: DynamiCrafter(\textbf{DC})~\cite{xing2023dynamicrafter}, I2VGen-XL(\textbf{VGen})~\cite{2023I2VGen-XL}, \textbf{SEINE}~\cite{chen2023seine}.}
\label{fig:supp_compare1}
\end{table}

\begin{table}
    \centering
    \footnotesize
    \resizebox{\linewidth}{!}{
    \setlength{\tabcolsep}{0.2em} 
    \begin{tabular}{c|cccccc}
        {Input} & 
        \multicolumn{5}{c}{{Generation (left$\rightarrow$right: time steps)}} \\
    
    \includegraphics[width=0.16\linewidth]{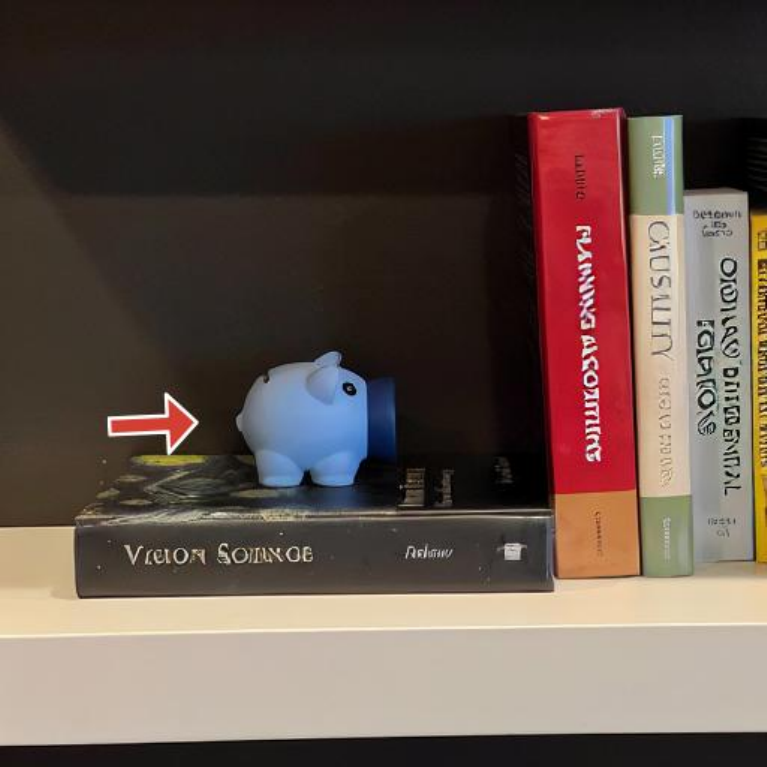} & \visvideo{dynamcrafter}{book_pig}{3}{5}{7}{9}{12}{} & \rotatebox{270}{\hspace{-40pt}DC~\cite{xing2023dynamicrafter}} \\
     & \visvideo{i2vgen-xl}{book_pig}{1}{3}{7}{11}{14}{} & \rotatebox{270}{\hspace{-40pt}VGen~\cite{2023I2VGen-XL}} \\

    & \visvideo{seine}{book_pig}{4}{6}{10}{12}{14}{}
    & \rotatebox{270}{\hspace{-40pt}SEINE~\cite{chen2023seine}} \\
    
     & \visvideo{ours}{book_pig}{2}{4}{6}{8}{11}{} & \rotatebox{270}{\hspace{-30pt}Ours} \\

     \includegraphics[width=0.16\linewidth]{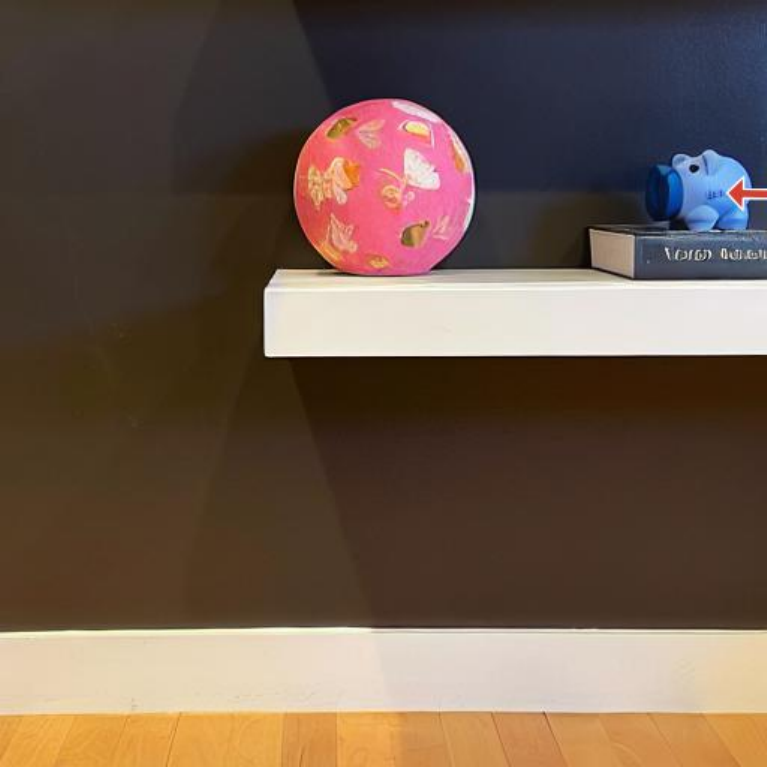} & \visvideo{dynamcrafter}{pig_ball}{2}{4}{6}{8}{10}{} & \rotatebox{270}{\hspace{-40pt}DC~\cite{xing2023dynamicrafter}} \\
     & \visvideo{i2vgen-xl}{pig_ball}{2}{4}{8}{10}{13}{} & \rotatebox{270}{\hspace{-40pt}VGen~\cite{2023I2VGen-XL}} \\

    & \visvideo{seine}{pig_ball}{6}{7}{8}{9}{10}{}
    & \rotatebox{270}{\hspace{-40pt}SEINE~\cite{chen2023seine}} \\
    
     & \visvideo{ours}{pig_ball}{1}{3}{5}{7}{9}{} & \rotatebox{270}{\hspace{-30pt}Ours} \\

    \end{tabular} 
    }
\captionof{figure}{\textbf{Additional qualitative comparison} against I2V generative models: DynamiCrafter(\textbf{DC})~\cite{xing2023dynamicrafter}, I2VGen-XL(\textbf{VGen})~\cite{2023I2VGen-XL}, \textbf{SEINE}~\cite{chen2023seine}.}
\label{fig:supp_compare2}
\end{table}
\begin{table}
    \centering
    \footnotesize
    \resizebox{\linewidth}{!}{
    \setlength{\tabcolsep}{0.2em} 
    \begin{tabular}{c|ccccc}
        {Input} & 
        \multicolumn{5}{c}{{Generation (left$\rightarrow$right: time steps)}} \\
    \includegraphics[width=0.16\linewidth]{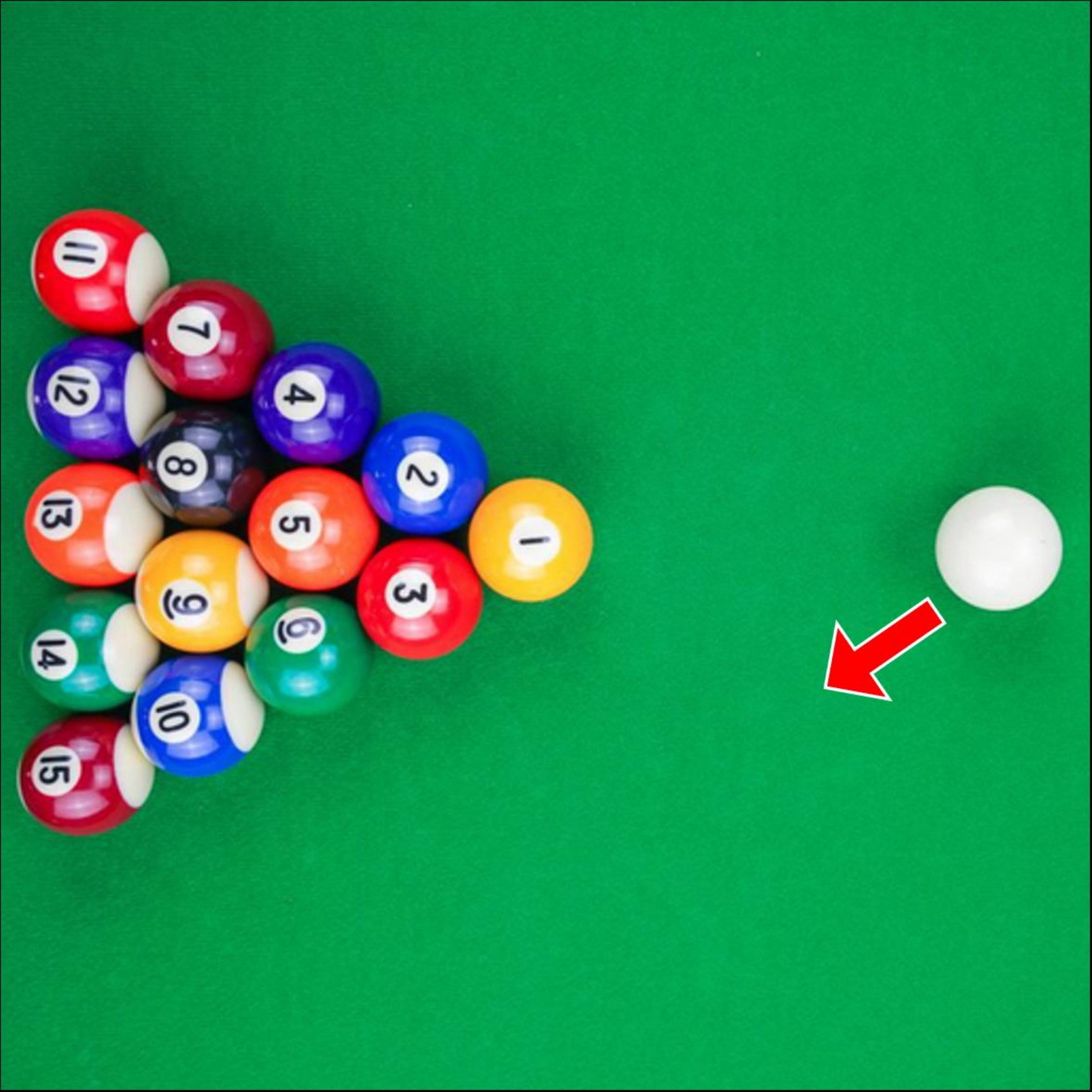} & \visvideo{ours}{pool_v2}{1}{2}{5}{8}{10}{} \\
    \includegraphics[width=0.16\linewidth]{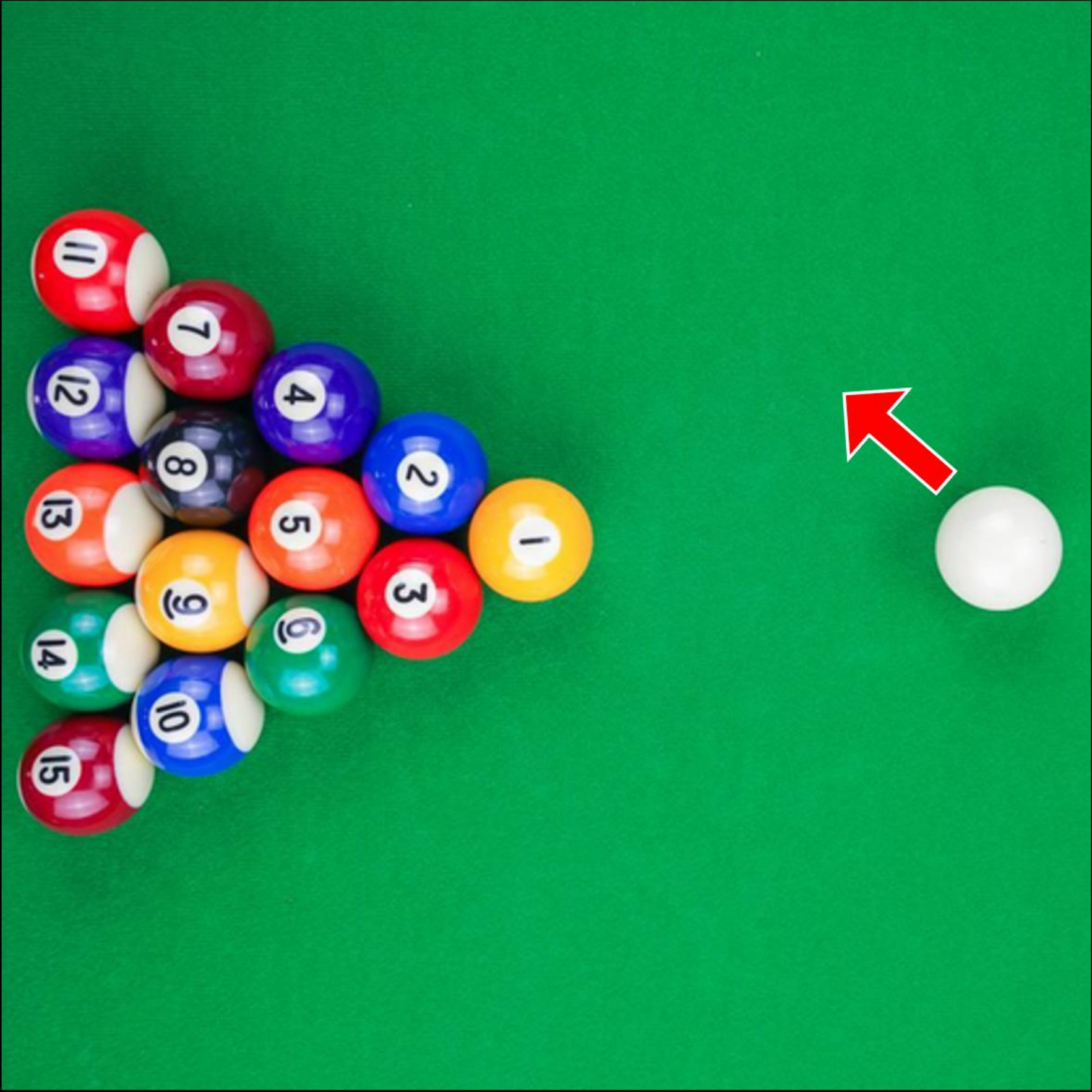} & \visvideo{ours}{pool_v3}{1}{2}{4}{8}{12}{} \\
    \includegraphics[width=0.16\linewidth]{src_figs/ours/pool/pool_input3.pdf} & \visvideo{ours}{pool_v4}{1}{3}{5}{7}{10}{} \\
    \includegraphics[width=0.16\linewidth]{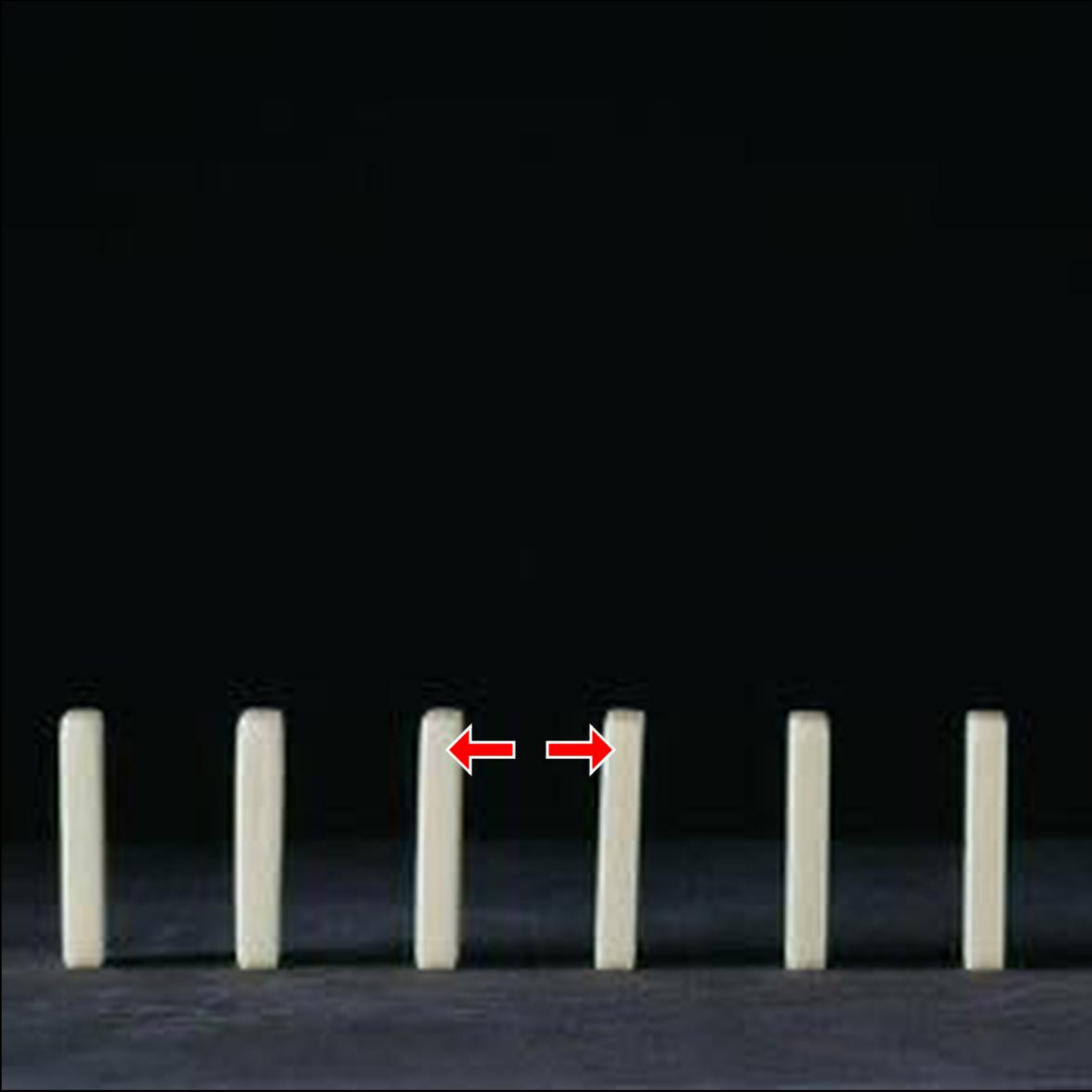} & \visvideo{ours}{domino_v2}{2}{5}{7}{9}{11}{} \\
    \includegraphics[width=0.16\linewidth]{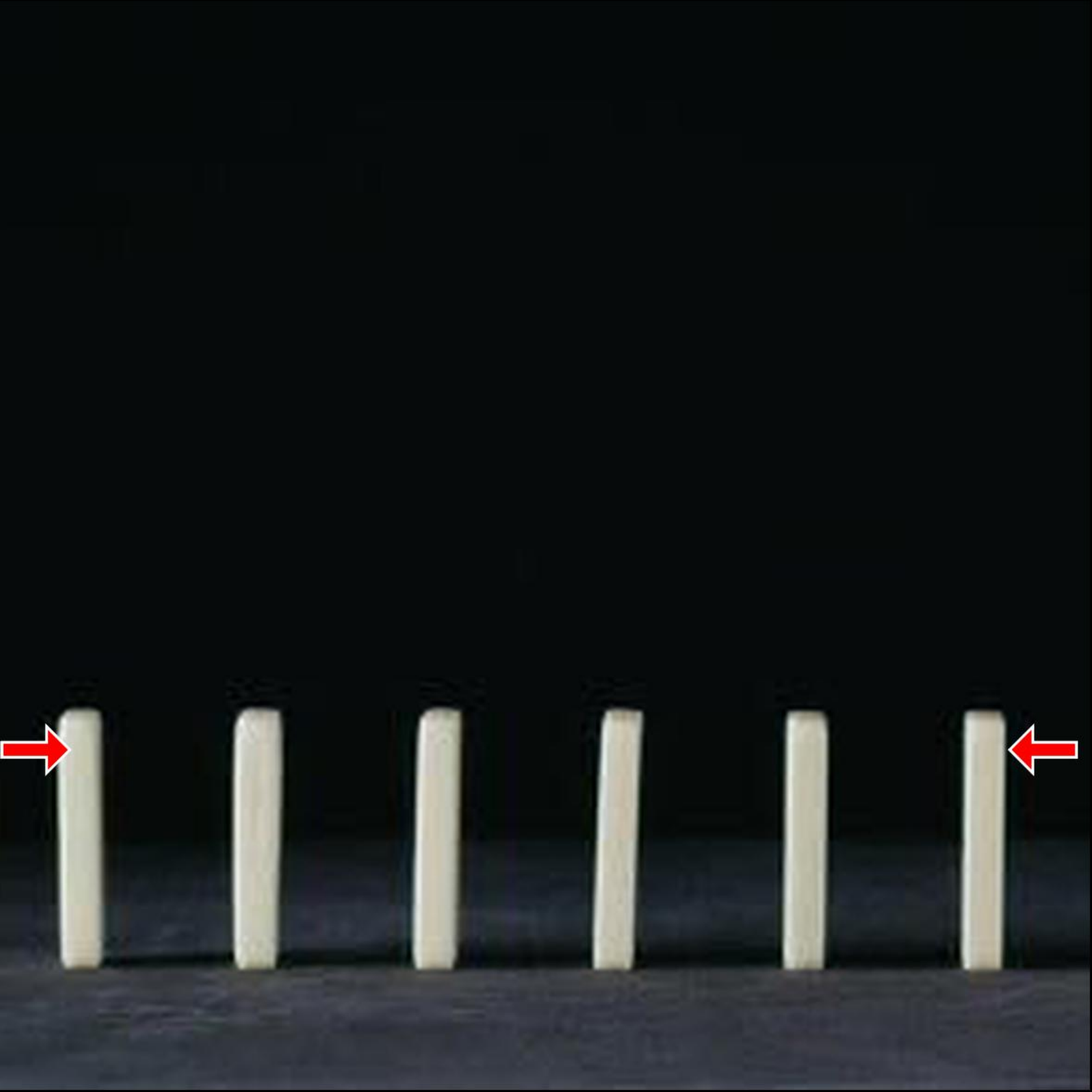} & \visvideo{ours}{domino_v3}{5}{7}{9}{12}{15}{} \\
    \includegraphics[width=0.16\linewidth]{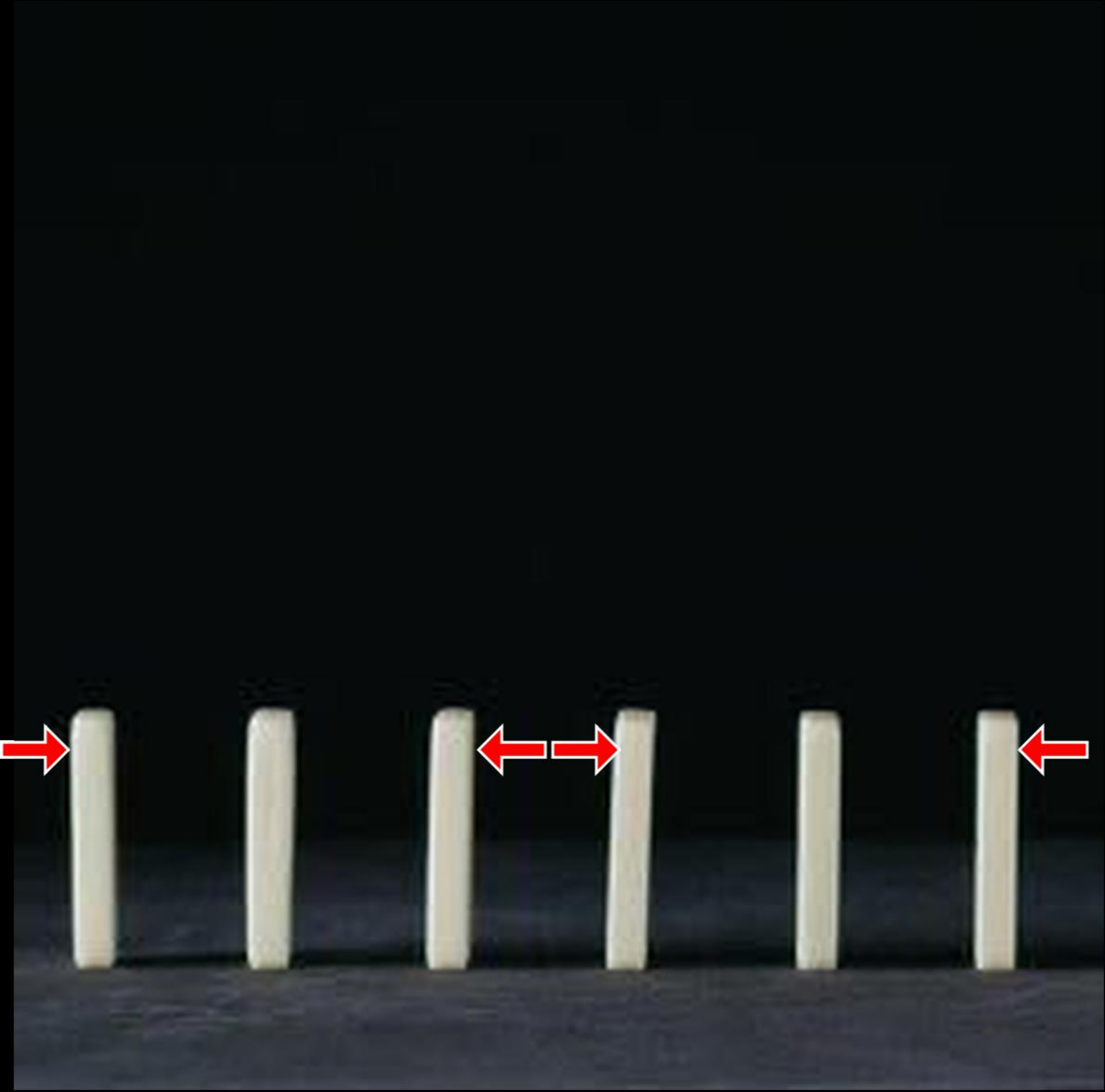} & \visvideo{ours}{domino_v4}{4}{7}{9}{12}{15}{} \\
    \end{tabular}
    }
\captionof{figure}{\textbf{More controllable video generation}.}
\label{fig:controllable}
\end{table}
\begin{table}[t]
    \centering
    \scriptsize
    \resizebox{.75\linewidth}{!}{
    \setlength{\tabcolsep}{0.1em} 
    \begin{tabular}{ccc}
     \includegraphics[width=0.33\textwidth]{src_figs/ours/domino/fig6_input1.pdf} 
     & \includegraphics[width=0.33\textwidth]{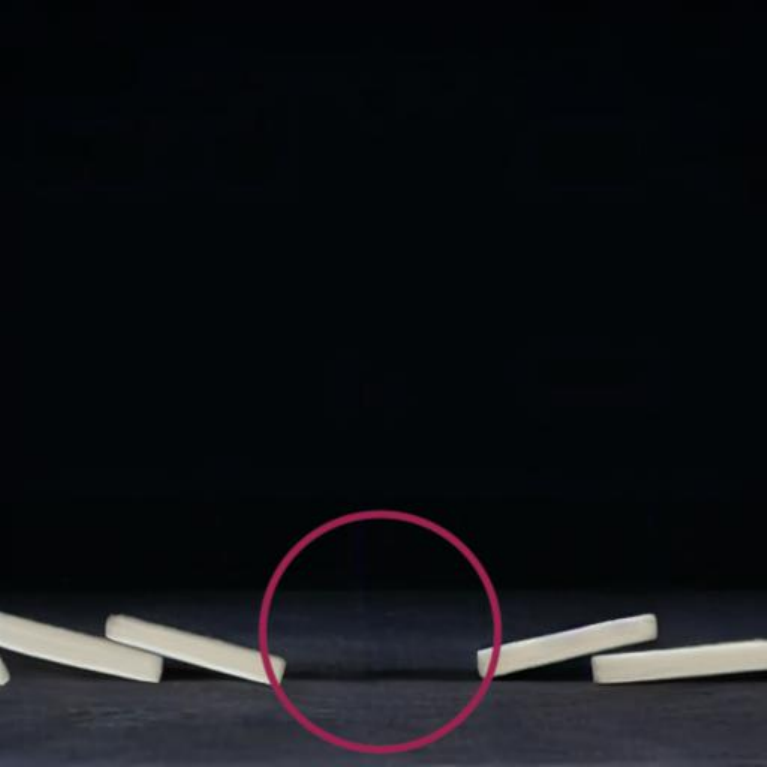} & 
     \includegraphics[width=0.33\textwidth]{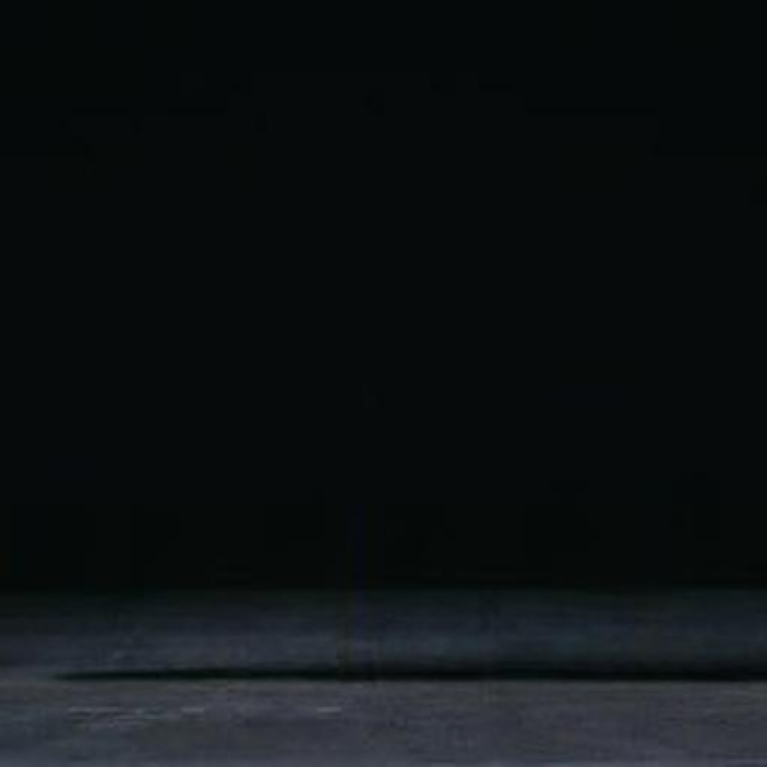} \\
     {Input} & {Artifact in generation}  & {Bg Inpainting} \\

    \includegraphics[width=0.33\textwidth]{src_figs/ours/boxes/frame_0.pdf} 
     & \includegraphics[width=0.33\textwidth]{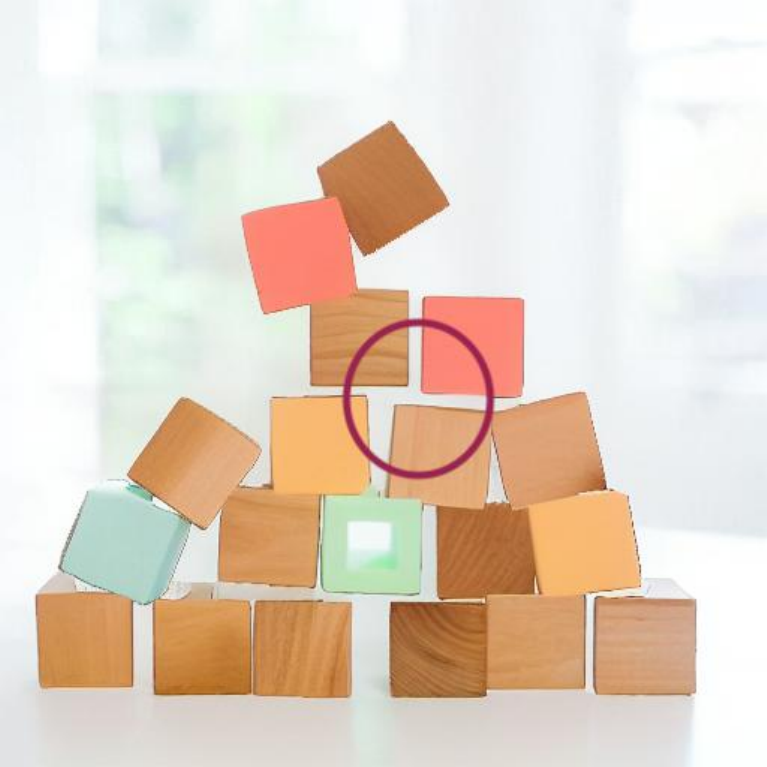} & 
     \includegraphics[width=0.33\textwidth]{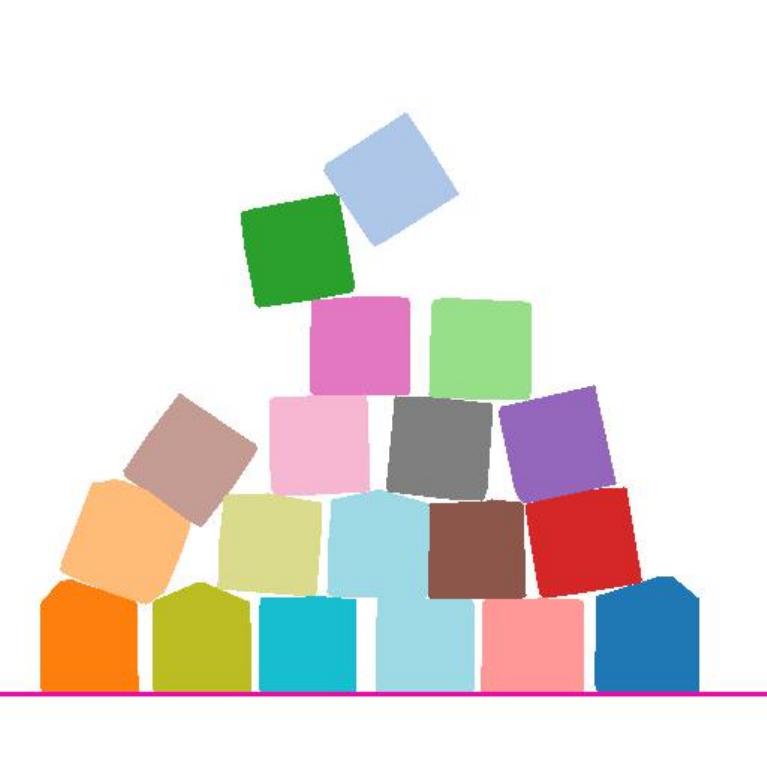} \\
     {Input} & {Artifact in generation}  & {Simulation} \\

    \includegraphics[width=0.33\textwidth]{src_figs/ours/pool/frame_0.pdf} 
     & \includegraphics[width=0.33\textwidth]{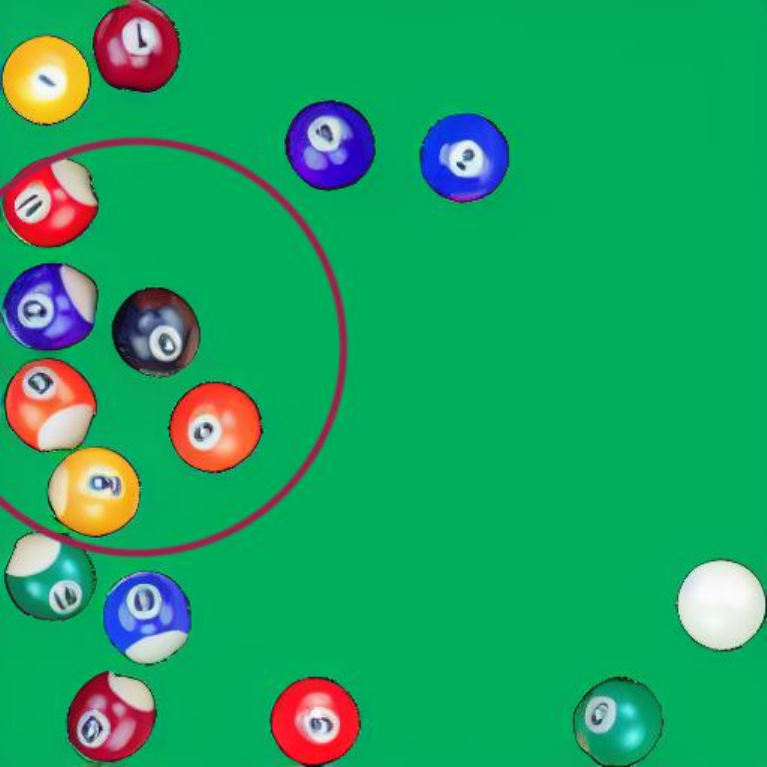} & 
     \includegraphics[width=0.33\textwidth]{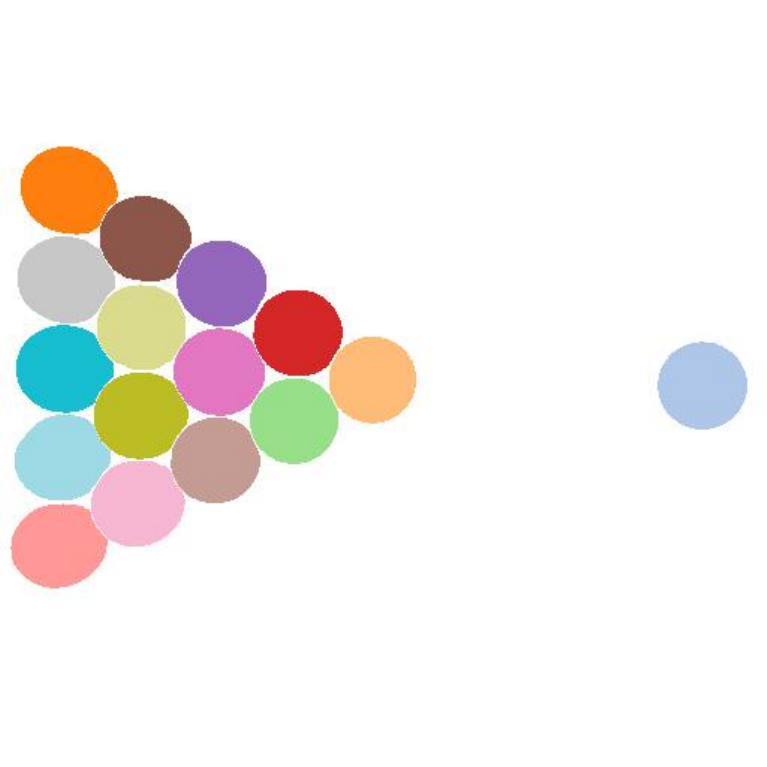} \\
     {Input} & {Artifact in generation}  & {Segmentation} \\

    \includegraphics[width=0.33\textwidth]{src_figs/ours/boxes/frame_0.pdf} 
     & \includegraphics[width=0.33\textwidth]{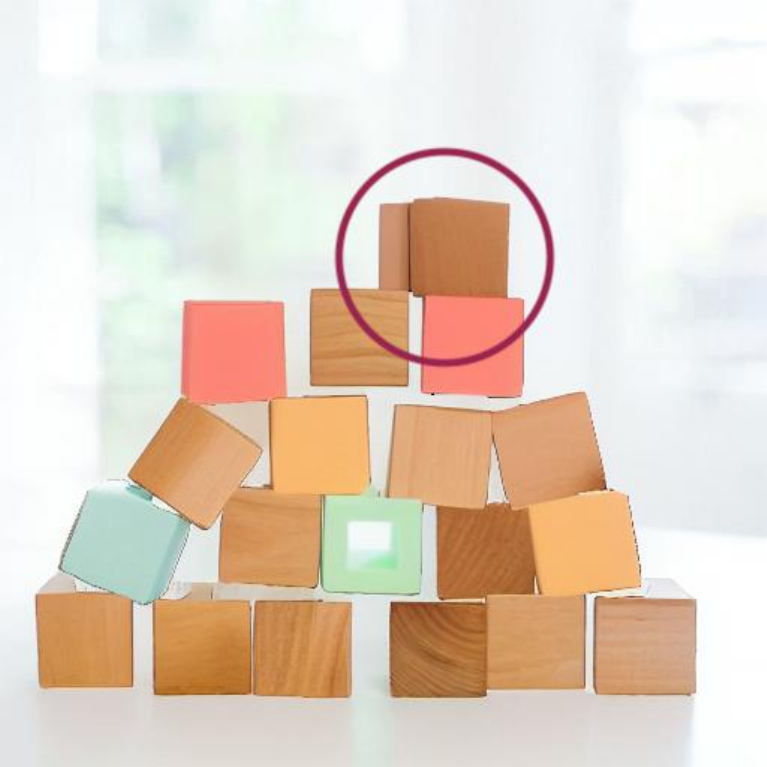} & 
     \includegraphics[width=0.33\textwidth]{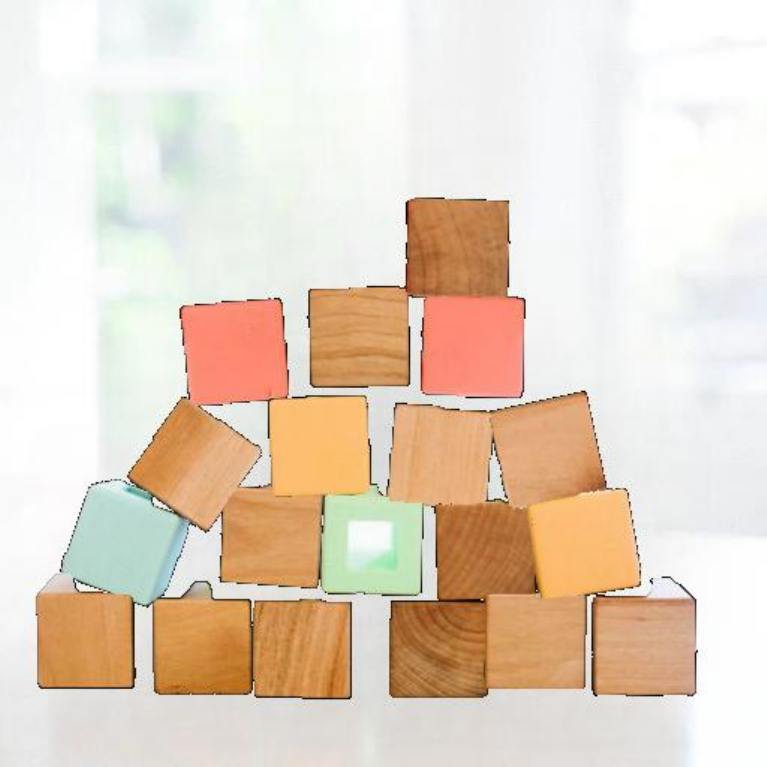} \\
     {Input} & {Artifact in generation}  & {Rendered image before refinement} \\
     
    \end{tabular}
    }
\captionof{figure}{\textbf{Limitation analysis.} We showcase 3 different examples of our current method's limitation. The left column shows the input image, the middle column shows the sampled frame from the generation video with artifacts, the right column shows the underlying reason for the artifact.}
\label{fig:supp_limitation}
\end{table}

\subsection{Quantitative evaluation}
\label{sec:supp_quantitative}
To quantitative evaluate the generation performance and compare with other approaches, we record 50 videos for a given scene as GT by varying initial forces applied on the object. The random selected sequences are shown in \cref{fig:supp_real}.

\section{More experiments}
\label{sec:supp_experiments}

\subsection{Open-world movable object segmentation} 

To measure the performance of the open-world movable object segmentation, we collect 10 images of very complicated open-world scenes where 118 movable instances are annotated following COCO~\cite{lin2014microsoft} format. The perception system works well, achieves \textbf{0.93} precision \textbf{0.82} recall under $0.5$ IoU. The precision-recall curve is shown in \cref{fig:pr_curve}. Qualitative results are shown in \cref{fig:seg}.

\subsection{GPT-4V physical property estimation evaluation}

We evaluate the physical property estimator GPT-4V used in the paper. For \emph{1) mass}: follow~\cite{albert2024phynerf}, we select 20 different portable objects from ABO dataset and find that GPT4-V has an average absolute error of $\textbf{0.39 kg}$ and achieves $\textbf{75\%}$ accuracy within $30\%$ of the GT.
\emph{2) friction and elasticity}: Since GT is unavailable, we use reference videos of toy cars sliding on various surfaces and balls of different materials bouncing to rank materials by friction and elasticity. Comparing the models' rankings to the ones in the videos, GPT-4V gives reasonable estimations, getting $\textbf{12 out of 13}$ for friction and $\textbf{6 out of 7}$ for elasticity across different comparisons. Two testing scenarios of friction and elasticity are shown in \cref{fig:phy_eval}.

\section{More qualitative results}
\label{sec:supp_qualitative}

\subsection{Qualitative comparison}
We visualize more qualitative comparisons in \cref{fig:supp_compare1} and \cref{fig:supp_compare2}. As can be seen, image-video generation methods could not generate physically plausible outputs. We notice DynamiCrafter~\cite{xing2023dynamicrafter} model tends to generate static scene with lighting or viewpoint changes, without output reasonable physical motion. SEINE~\cite{chen2023seine} and I2VGen-XL~\cite{2023I2VGen-XL} outputs reasonably apparent motions, but the motions are not physical realistic, and the foreground objects could not guarantee consistency across different time steps. Please check the supplementary video for details.

\subsection{More controllable generation}
We show more controllable generation results in \cref{fig:controllable}.

\section{Limitation Analysis}
\label{sec:limitation}
We summarize 4 different generation artifacts in \cref{fig:supp_limitation}. 

\paragraph{\textbf{Incorrect Inpainting.}} The first row shows the domino example. The generation results contains incorrect shadow in the middle of the frame. The reason is that the shadow of the boxes could not be fully removed in the inpainted image as shown on the right.

\paragraph{\textbf{Gap between segmentations and primitives.}}
The second row shows the blocks example where there is a gap between different blocks in the generated videos, whereas no such phenomenon appeared in simulation. The reason is the primitives used in the simulation is slight different from the real segments, thus could cause some gap in the simulated results and rendered outputs.

\paragraph{\textbf{Inaccurate Segmentation.}} The third row shows the billiard example where the balls' shape is not perfect circle in the generated frames given the input segmentation mask is not accurate on the boundaries. Thus the synthesized ball has artifacts near its boundary. The generated video quality is affected by the segmentation mask of the input image.

\paragraph{\textbf{Hallucinations introduced by diffusion refinement.}} The fourth rows shows the blocks example where the diffusion refinement brings hallucination of the input object. In our proposed generative refinement algorithm, the generated latent injects into both the foreground and background. Thus in some scenarios there could be hallucinations of the foreground object and slightly modify its appearance, \eg the block on the top.

\end{document}